\documentclass[pdflatex,sn-basic]{sn-jnl}%

\usepackage[T1]{fontenc}
\usepackage{lmodern}
\usepackage{fix-cm}
\usepackage{anyfontsize}
\usepackage{graphicx}%
\usepackage{multirow}%
\usepackage{amsmath,amssymb,amsfonts}%
\usepackage{amsthm}%
\usepackage{mathrsfs}%
\usepackage[title]{appendix}%
\usepackage{xcolor}%
\usepackage{textcomp}%
\usepackage{manyfoot}%
\usepackage{booktabs}%
\usepackage{algorithm}%
\usepackage{algorithmicx}%
\usepackage{algpseudocode}%
\usepackage{listings}%
\usepackage{longtable}
\usepackage{array}
\usepackage{colortbl}
\usepackage{xcolor}

\usepackage{xspace}
\makeatletter
\DeclareRobustCommand\onedot{\futurelet\@let@token\@onedot}
\def\@onedot{\ifx\@let@token.\else.\null\fi\xspace}

\def\eg{\emph{e.g}\onedot} 
\def\ie{\emph{i.e}\onedot} 
 
 \def\vs{\emph{vs}\onedot}

\makeatother

\usepackage{eso-pic}
\usepackage{url}
\AddToShipoutPicture*{%
     \AtTextUpperLeft{%
         \put(-3.5,10){
           \begin{minipage}{\textwidth}
              \scriptsize
              \MakeUppercase{Preprint} 
           \end{minipage}}%
     }%
}

\begin{document}

\title[Reducing Label Dependency for Underwater Scene Understanding: A Survey of Datasets, Techniques and Applications]{Reducing Label Dependency for Underwater Scene Understanding: A Survey of Datasets, Techniques and Applications}

\author*[]{\fnm{Scarlett} \sur{Raine*}}\email{sg.raine@qut.edu.au}

\author[]{\fnm{Frederic} \sur{Maire}}\email{f.maire@qut.edu.au}

\author[]{\fnm{Niko} \sur{Suenderhauf}}\email{niko.suenderhauf@qut.edu.au}

\author[]{\fnm{Tobias} \sur{Fischer}}\email{tobias.fischer@qut.edu.au}

\affil[]{\orgdiv{QUT Centre for Robotics}, \orgname{Queensland University of Technology (QUT)}, \orgaddress{\street{George St}, \city{Brisbane}, \postcode{4000}, \state{Queensland}, \country{Australia}}}

\abstract{Underwater surveys provide long-term data for informing management strategies, monitoring coral reef health, and estimating blue carbon stocks. Advances in broad-scale survey methods, such as robotic underwater vehicles, have accelerated and increased the range of marine surveys but generate large volumes of imagery requiring analysis. Computer vision methods such as semantic segmentation aid automated image analysis, but typically rely on fully supervised training with extensive labelled data. While ground truth label masks for tasks like street scene segmentation can be quickly and affordably generated by non-experts through crowdsourcing services like Amazon Mechanical Turk, marine ecology presents greater challenges. The complexity of underwater images, coupled with the specialist expertise needed to accurately identify species at the pixel level, makes this process costly, time-consuming, and heavily dependent on domain experts. In recent years, numerous works have performed automated analysis of underwater imagery, and a smaller number of studies have focused on weakly supervised approaches which aim to reduce the amount of expert-provided labelled data required. This survey focuses on approaches which reduce dependency on human expert input, while comprehensively reviewing the prior and related approaches to position these works in the wider field of underwater perception. Further, we offer a comprehensive overview of coastal ecosystems and the challenges of underwater imagery. We provide background on weakly and self-supervised deep learning, and integrate these elements into a taxonomy that centres on the intersection of underwater monitoring, computer vision, and deep learning, while motivating approaches for weakly supervised deep learning with reduced dependency on domain experts for data annotations. Lastly, the survey examines available datasets and platforms, and identifies the gaps, barriers, and opportunities for automating underwater surveys.}


\maketitle

\section{Introduction}\label{sec:intro}

Marine surveys are critical for scientists to monitor long term changes in the health of reef ecosystems and to enable evidence-based decision-making on reef management strategies~\citep{ditria2022artificial}. Traditionally, surveys are performed manually by divers, snorkellers or using helicopters. However, the range and accuracy of manual surveys is limited and species identification requires visual observation by marine ecologists. Therefore, broad-scale survey methods are becoming increasingly prevalent and utilise remote sensing, drones, towed underwater vehicles, and autonomous underwater vehicles~\citep{gonzalez2020monitoring, murphy2010observational}.  The significant increase in the amount of data generated by these methods requires automated methods, such as machine learning and deep learning, to process the information and provide analysis to ecologists in a timely and accurate manner~\citep{gonzalez2020monitoring}.

One method for automated image analysis is semantic segmentation, a computer vision task in which the class of every pixel in a query image is predicted~\citep{guo2018review}. In recent years, it has become common to train deep learning models to perform semantic segmentation (refer to the glossary in Table~\ref{table:glossary} for definitions of key domain-specific and technical terms used in this survey).  These models are often trained using full supervision, \ie on pairs of images and dense ground truth masks, where the mask is comprised of a semantic label for every corresponding pixel in the paired image~\citep{zhang2019survey}. These dense masks are typically created by human experts who manually label each pixel.  However, underwater imagery has visually unique characteristics which complicate the process of labelling images.

\begin{figure*}[t]
\centering
\begin{tabular}{ccc}
    \includegraphics[width=40mm]{fig1_a.jpg} &
    \includegraphics[width=40mm]{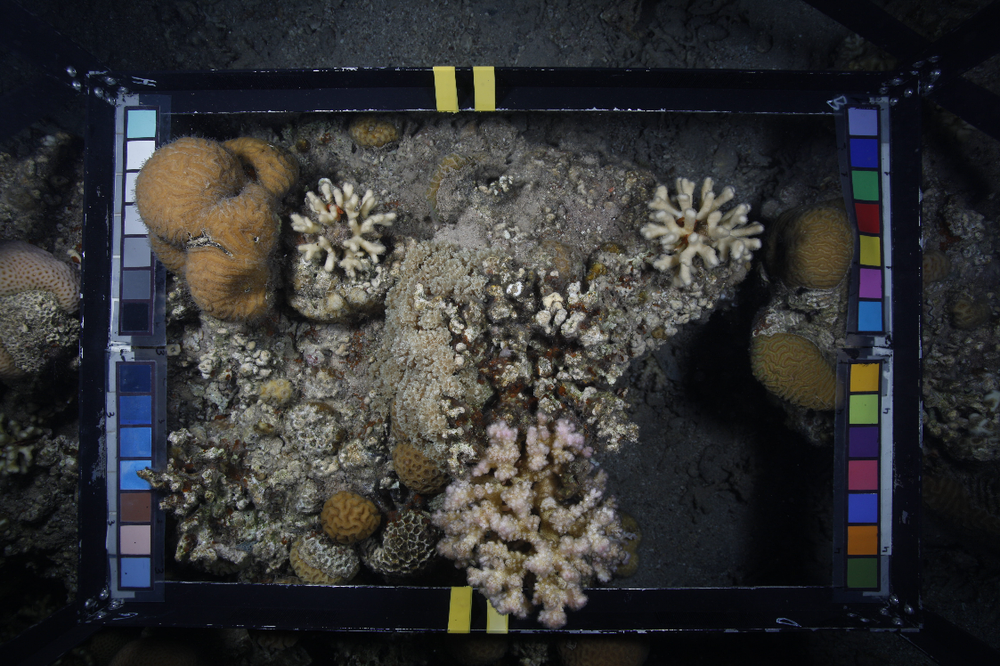} &
    \includegraphics[width=40mm]{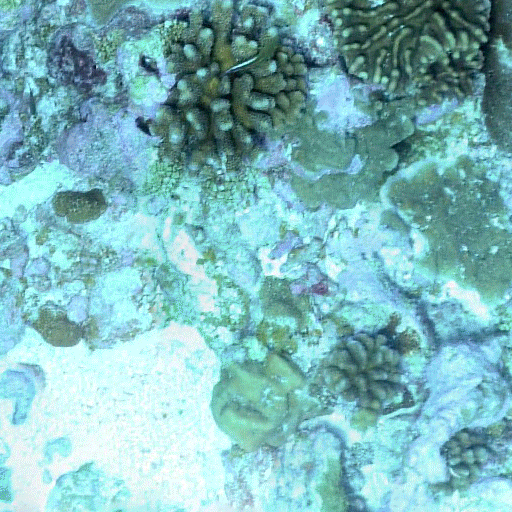} \\
    a) & b) & c)\\
    \includegraphics[width=40mm]{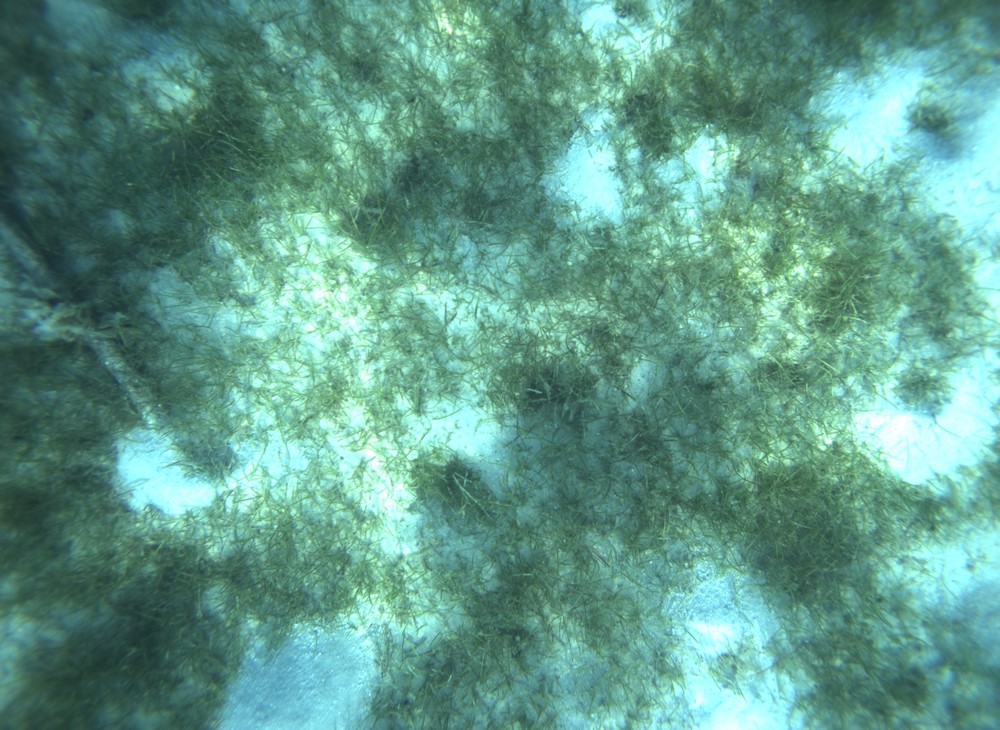} &     \includegraphics[width=40mm]{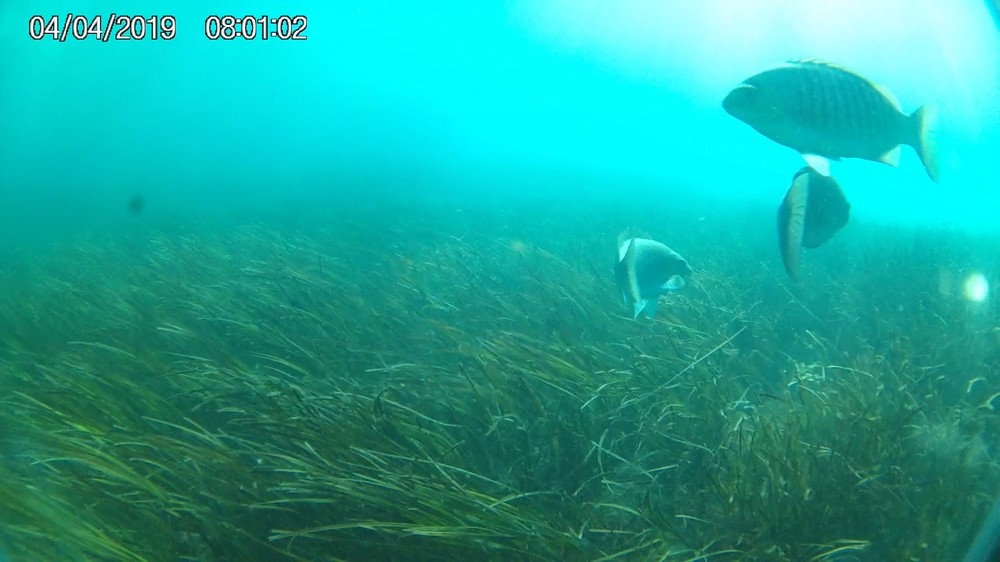} &
    \includegraphics[width=40mm]{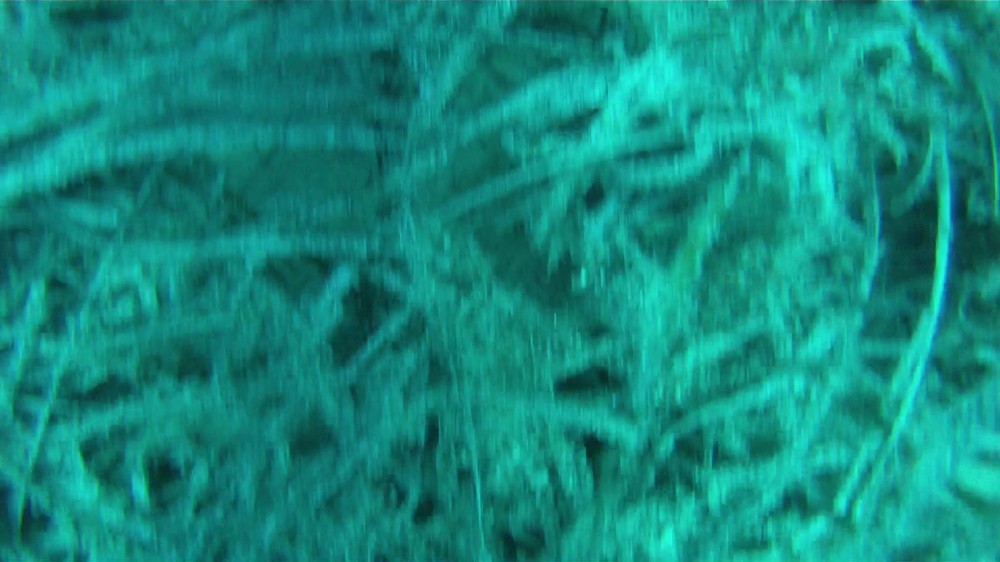} \\
    d) & e) & f) \\
\end{tabular}
\caption[Variation in Underwater Imagery]{Examples of variation in underwater imagery. Images\footnotemark~show differences in camera quality, lighting, turbidity, camera viewpoint and resolution.}
\label{fig:example-variation}
\end{figure*}

Analysing seagrass images for the different species present is necessary for accurately estimating the amount of carbon stored, and for monitoring the composition and health of meadows over time~\citep{lavery2013variability}. In the case of coral imagery, differentiating between coral species is needed for determining ecosystem health indicators and tracking responses to environmental changes~\citep{murphy2010observational}. However, underwater imagery of seagrass meadows or coral reefs is visually complex and characterised by overlapping instances and poorly defined boundaries~\citep{noman2023improving}. It is also highly variable and often affected by turbidity, discolouration due to the water column, depth and lighting, and impacted by shadows and camera blur (Fig.~\ref{fig:example-variation})~\citep{sun2018transferring, jin2017deep}.  
Accurately determining the marine species present is complicated as instances of the same species can exhibit different visual characteristics based on stage of growth, health and location. In the case of seagrass meadows, images can contain hundreds or thousands of individual seagrass leaves, which often belong to different species of seagrasses and overlap with one another.  \footnotetext{Image a) is publicly released in the \textit{XL CATLIN Seaview Survey} dataset~\citep{gonzalez2014catlin, gonzalez2019seaview} under a \href{https://creativecommons.org/licenses/by/3.0/deed.en}{CC BY 3.0 licence}; image b) is publicly released in the \textit{EILAT Fluorescence} dataset~\citep{beijbom2016improving} under a \href{https://creativecommons.org/publicdomain/zero/1.0/}{Public Domain Licence}; image c) is publicly released as part of the \textit{UCSD Mosaics} dataset~\citep{edwards2017large, alonso2019coralseg} and reproduced with permission from Springer Nature; image d) was taken by Serena Mou using the \textit{FloatyBoat} Autonomous Surface Vehicle (ASV)~\citep{mou2022reconfigurable} and used with permission; image e) is publicly released in the \textit{Global Wetlands} dataset~\citep{ditria2021annotated} under a \href{https://creativecommons.org/licenses/by/4.0/}{CC BY 4.0 licence}; and f) is from the \textit{Looking for Seagrass} dataset~\citep{reus2018looking}, released under a \href{https://github.com/EnviewFulda/LookingForSeagrass?tab=BSD-2-Clause-1-ov-file}{BSD 2-Clause Licence}.}This image complexity and level of expertise required to accurately determine the species of each pixel makes it infeasible to use crowd-sourcing annotation services such as Amazon Mechanical Turk\footnote{\url{https://www.mturk.com/}.}, and prohibitively costly, time-consuming and difficult for domain experts to provide this level of annotation. Computer vision approaches for underwater imagery must therefore balance the time and cost for domain experts to label training images, the desired resolution and accuracy of the algorithm, and the computation time per image \citep{li2024survey}.

\begin{figure}[t]
\centering
\includegraphics[width=0.9\columnwidth, clip, trim=0cm 0cm 0cm 0cm]{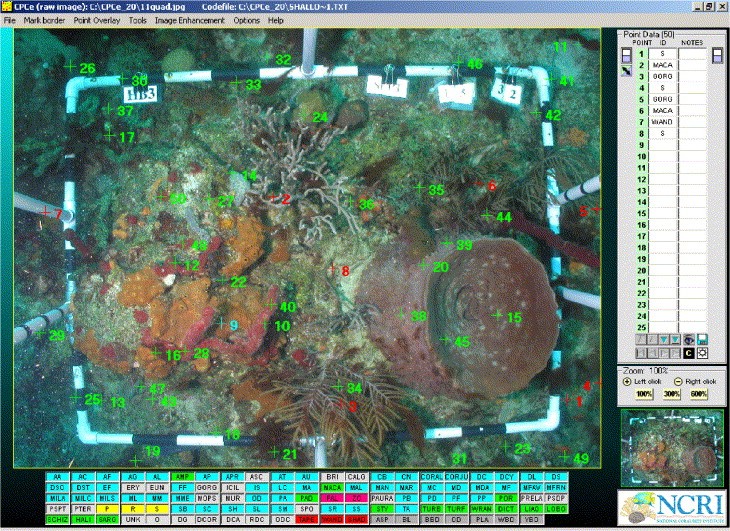} 
\caption[Coral Point Count Annotation]{An example image labelled with sparse random points in the Coral Point Count with Excel extensions (CPCe) software\footnotemark~\citep{kohler2006coral}.}
\label{fig:example-cpc}
\end{figure}

In this survey, we provide an overview of weakly supervised analysis of underwater imagery by considering the literature for each of the distinct components: coastal ecosystems, challenges of working with underwater imagery, computer vision, deep learning, and approaches for automated analysis of seagrass and coral imagery.  We then review the weakly and self-supervised approaches both generally and in the underwater domain.  

This survey also performs a comprehensive review of the available underwater datasets.  Historically, marine scientists have been interested in estimating the percentage cover of marine species. To obtain these estimates, scientists have collected large quantities of photo-quadrat images collected at intervals along a transect of the reef (Fig.~\ref{fig:example-cpc}).  Photo-quadrats are top-down images taken by a stationary diver, where each image is scaled or cropped to represent a certain area, \eg 1m by 1m of the seafloor~\citep{gonzalez2019seaview}.  The photo-quadrats are labelled by domain experts using the Coral Point Count (CPC) method: a set number of sparse points either arranged in a grid or randomly distributed~\citep{kohler2006coral}. Further, this survey analyses the availability of data collected for the purposes of computer vision or deep learning. These datasets are labelled using dense pixels, polygons, bounding boxes, patches and image labels.

\footnotetext{Reprinted from Computers \& Geosciences, Vol 32, Kevin E. Kohler \& Shaun M. Gill, Coral Point Count with Excel extensions (CPCe): A Visual Basic program for the determination of coral and substrate coverage using random point count methodology, pages 1259-1269, Copyright (2006), with permission from Elsevier.} 

The use of broad-scale survey methods such as autonomous underwater or surface vehicles often necessitates light-weight, fast processing of images.  In addition, automated processing of imagery can be run in real-time to enable dynamic adjustment of the survey path towards targets of interest~\citep{li2022real}. Even if images are not analysed live during deployment, there is a need for fast, often overnight, analysis of data collected to inform the next day of monitoring, or to enable training or re-training of models. Furthermore, field deployments can be remote, with limited access to compute and cloud infrastructure. Deep learning models for analysing underwater imagery should therefore prioritise fast computation times and lightweight architectures. We discuss applications which require reduced dependency on domain expert labels and highlight the limitations and opportunities for future work. 

\begin{figure}[t]
\centering
\includegraphics[width=\columnwidth, clip, trim=4.6cm 11cm 4.6cm 4.5cm]{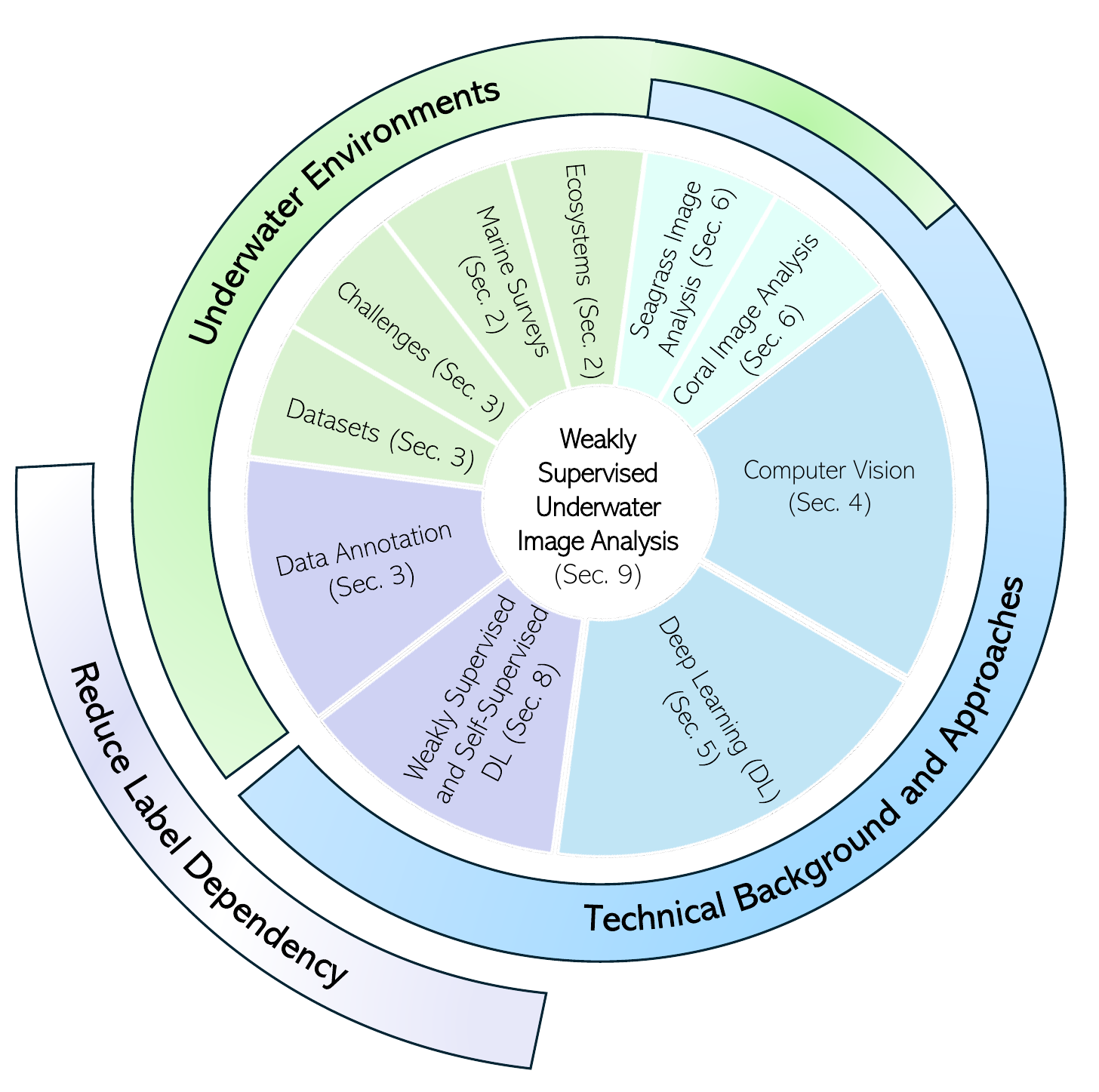} 
\caption{Weakly Supervised Underwater Image Analysis aims to reduce dependency on domain expert labelling, and operates at the intersection of underwater environmental monitoring, computer vision and deep learning. This survey reviews relevant prior approaches in other fields as well as specialised implementations for analysis of underwater imagery.  This figure depicts the main sections of this survey and how the different facets are connected.}
\label{fig:front-page}
\end{figure}

Although there have been recent survey articles on underwater computer vision \citep{gonzalez2023survey, mittal2022survey}, emerging technologies for underwater surveys \citep{igbinenikaro2024emerging}, and underwater photogrammetry \citep{remmers2023close}, there is a significant gap in addressing the traditional dependency on full supervision and expert-annotated labels, which are often costly and time-consuming to obtain.  In this survey, we propose a taxonomy of literature which positions weak supervision for underwater image analysis at the centre (Fig.~\ref{fig:front-page}). This taxonomy also visualises the structure of the survey and highlights the relations between the different sections.

This survey first explains the importance of marine surveys and describes the challenges associated with species identification in marine environments (Section~\ref{sec:lit-marine}).  We identify key challenges of analysing underwater imagery, review the available datasets, and describe typical data annotation methods in Section~\ref{sec:lit-underwater-imagery}). We then review relevant background literature for computer vision tasks (Section~\ref{sec:lit-back-CV}) and deep learning concepts (Section~\ref{sec:lit-types-dl}).  As seen in Fig.~\ref{fig:front-page}, at the intersection of literature on underwater environments and the technical background lies approaches for analysis of seagrass meadow (Section~\ref{sec:lit-seagrass}) and coral reef imagery (Section~\ref{sec:lit-coral}). The focus on reducing label dependency is highlighted in Section~\ref{sec:lit-weakly}, and then applied to underwater images in Section~\ref{sec:weakly-underwater}, which brings together the three key concepts of weakly supervised deep learning for analysis of underwater imagery.  Finally, we discuss the challenges and gaps for weakly supervised image analysis and the opportunities for future work in Section~\ref{sec:discussion}, before summarising the key findings and concluding the survey in Section~\ref{sec:conc}. We provide a glossary for domain-specific and technical terms commonly used in this review in Table~\ref{table:glossary}.  Appendix~\ref{appendix:A} provides a comprehensive overview of the publicly available datasets of underwater imagery. 

The survey primarily concentrates on the task of semantic segmentation, which is the task of predicting the class of every pixel in an image. Approaches which perform object detection, which aims to find distinct objects of interest in an image and identifies targets with a bounding box around the predicted region; image-level classification, in which the class of an entire image is predicted; and instance segmentation, which predicts the pixels belonging to each object in an image, without any consideration of the semantic class of the segments proposed, are not covered exhaustively but provide context and background for the state of the field. 

We define the term `underwater imagery' to encompass any imagery collected by a variety of methods: human divers and snorkellers, surface vehicles, autonomous and semi-autonomous underwater vehicles.  This underwater imagery can have a variety of visual characteristics, camera viewpoints and varying qualities and resolutions, as shown in Fig.~\ref{fig:example-variation}.

This survey explicitly considers underwater imagery for two ecological tasks: mapping and monitoring seagrass meadows for estimation of carbon sequestration or for seagrass meadow management and conservation; and analysis of coral imagery for monitoring the abundance and densities of coral species in the face of natural and anthropogenic threats such as tropical cyclones or increased ocean temperatures. We do not comprehensively review approaches for other tasks \eg object detection of the Crown of Thorns Starfish, or detection and tracking of fish. 

This survey focuses on RGB imagery, and does not leverage fluorescence, depth, multi- or hyperspectral imaging.  Many survey platforms are designed for citizen scientist usage and are therefore simple and cost-effective, and use RGB underwater cameras such as GoPros\footnote{\url{https://gopro.com/en/au/}.} to collect imagery.  This survey therefore focuses specifically on RGB imagery only because fluorescence information, depth maps and other spectral bands are not always available.

This survey also focuses specifically on imagery collected underwater because it enables high resolution, multi-species analysis of the species present.  We do not review in detail approaches which use drone or satellite imagery.

\begin{longtable}{|>{\raggedright\arraybackslash}p{0.25\textwidth}|>{\raggedright\arraybackslash}p{0.7\textwidth}|}
\caption{Glossary of Domain-Specific and Technical Terms} \label{table:glossary} \\
\hline
\rowcolor{gray!20}
\multicolumn{2}{|c|}{\textbf{Domain-Specific Terms}} \\
\hline
\textbf{Term} & \textbf{Explanation} \\
\hline
\endfirsthead

\hline
\textbf{Term} & \textbf{Explanation} \\
\hline
\endhead

\hline
\multicolumn{2}{|r|}{{\textit{Continued on next page}}} \\
\hline
\endfoot

\hline
\endlastfoot

\arrayrulecolor{gray!30}
Turbidity & A measure of the presence of suspended particles in a liquid, influencing its clarity. \\
\hline
Water column & The vertical expanse of water between the surface and the ocean seafloor. \\
\hline
Percentage cover & Percentage cover refers to the proportion of an area covered by a species of interest. \\
\hline
Inter-tidal zone & The inter-tidal zone is under water at high tide and above water at low tide. \\
\hline
Sub-tidal zone & The sub-tidal zone remains under water except during extremely low tides. \\
\hline
Benthos & The community of organisms that live on, in, or near the seafloor. \\
\hline
Benthic habitat mapping & Creating maps of the seafloor and the habitats present. A benthic habitat is a group of organisms typically found together in a community, \eg coral reefs, seagrass meadows, kelp forests, etc. \\
\hline
Bathymetric data & Measurements of the depth and topography of the seafloor, providing detailed information on underwater terrain and elevation. \\
\hline
Transect & A linear, predefined survey path along which observations, measurements, or photos are systematically collected to assess the species, habitat type, or environmental conditions in a specific area. \\
\hline
Coral Point Count (CPC) & A commonly used data annotation scheme in which randomly placed or grid spaced sparse points are labelled by a domain expert. The labelled points are used to extrapolate estimates of the percentage coverage of the species present. \\
\hline
Random points & A set quantity (typically 50, 100, or 300) of pixels are randomly selected in an image for annotation by a domain expert. \\
\hline
Grid points & A set quantity (typically 50, 100, or 300) of evenly spaced pixels are selected for annotation by a domain expert. \\
\hline
Grid cells & An image is divided into a grid of cells, where each cell is assigned a label for the dominant class in that region by a domain expert. \\
\hline
Photo-quadrat & A image taken of the seafloor using a highly controlled camera setup \eg a frame is used to ensure the camera height and area of seafloor captured is consistent. These images are collected at defined intervals along a transect, and typically labelled using Coral Point Count. \\
\arrayrulecolor{black}
\hline
\rowcolor{gray!20}
\multicolumn{2}{|c|}{\textbf{Technical Terms}} \\
\hline
\textbf{Term} & \textbf{Explanation} \\
\hline
\arrayrulecolor{gray!30}
Image classification & A computer vision task for assigning a single label or category to an entire image based on its content. \\
\hline
Patch classification & The task of assigning a single label or category to a small patch taken from a larger image.\\
\hline
Coarse segmentation & The task of assigning a single label or category to every grid cell of a larger image, such that the relative density of each class in the image could be calculated.  \\
\hline
Semantic segmentation & The task of classifying each pixel in an image into a predefined category, enabling the model to understand and delineate different objects or regions within a scene. \\
\hline
Instance segmentation & The task of identifying each distinct object instance in an image, such that each individual segment is assigned a unique ID. \\
\hline
Panoptic segmentation & This task combines both semantic and instance segmentation, such that each segment is assigned both a class label and a unique ID to distinguish individual objects within each class. \\
\hline
Supervision & In the context of deep learning, \textit{supervision} refers to the extent that labelled data is used for training models. \textit{Fully supervised} indicates that pixel-level annotations are provided for each input, while \textit{weakly supervised} involves coarser or less complete labels, such as image-level classes or sparse point labels. \\
\hline
Superpixel & A group of adjacent pixels in an image that share similar characteristics, \eg colour or texture. \\
\hline
Point label propagation & An algorithm which takes sparse point labels, and expands or \textit{propagates} these points to other pixels in the given image, to create an augmented pseudo ground truth mask for the image. \\
\arrayrulecolor{black}
\hline
\end{longtable}

\section{Coastal Ecosystems}
\label{sec:lit-marine}

Positioned at the intersection of marine ecology, computer vision, artificial intelligence and robotics, this survey examines approaches for automated image analysis in marine conservation. The purpose of this section is to provide the necessary background information about seagrass meadows (Section~\ref{subsec:lit-back-seagrass}), coral reefs (Section~\ref{subsec:lit-back-coral}) and marine surveys (their importance is discussed in Section~\ref{subsec:lit-marine-survey} and survey methods are described in Section~\ref{subsec:lit-marine-methods}).

\subsection{Seagrass Meadows}
\label{subsec:lit-back-seagrass}
\begin{figure}[t]
\centering
\setlength{\tabcolsep}{2pt}
\centerline{\begin{tabular}{cc}
    \includegraphics[height=50mm]{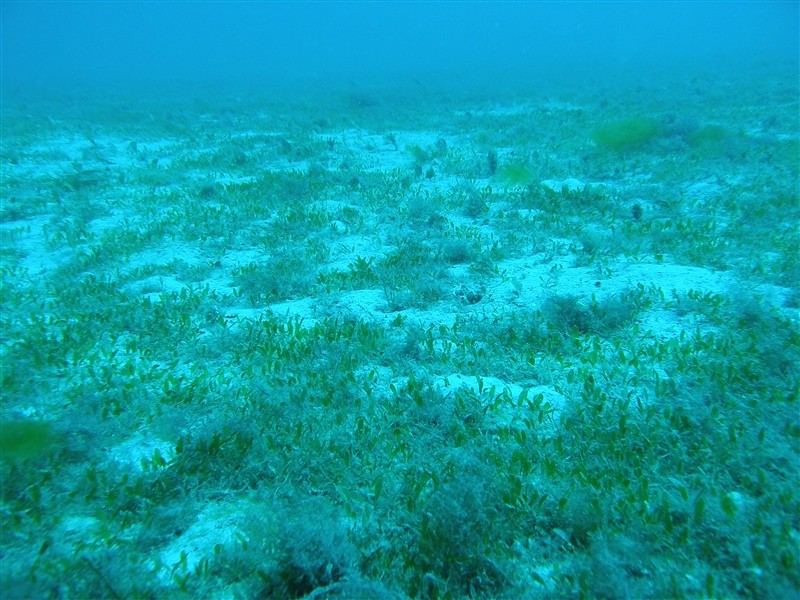} &
    \includegraphics[height=50mm]{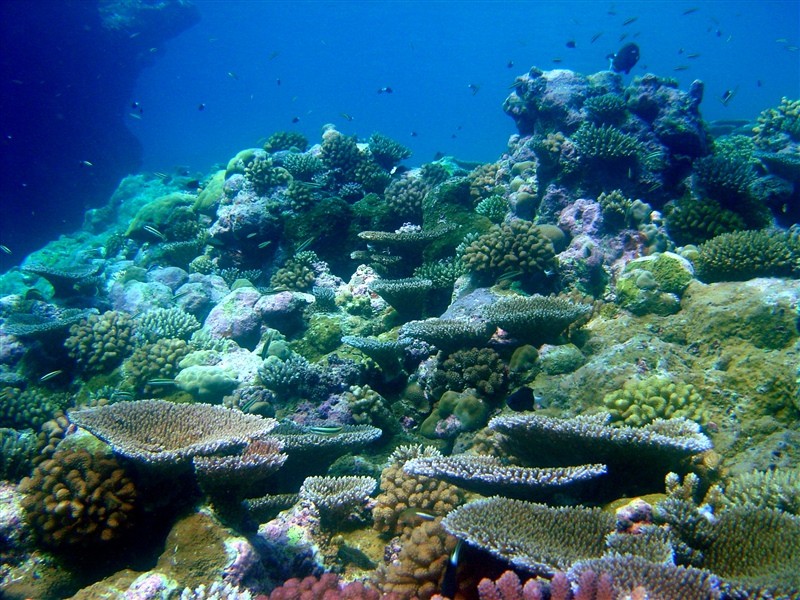} \\
    a) Seagrass meadows & b) Coral reefs \\
\end{tabular}}
\caption[Examples of Underwater Ecosystems]{Examples\footnotemark~of two important coastal ecosystems: seagrass meadows. Seagrass meadows (a) are communities of marine flowering plants which play a role in stabilising sediment, supporting marine life and sequestering blue carbon. Coral reefs (b) are living structures that support a rich diversity of marine species and act as natural protective barriers for coastlines.}
\label{fig:ecosystems}
\end{figure}

In this survey, we first consider seagrass meadows and describe their importance and ecosystem services. Seagrass meadows are comprised of marine flowering plants which grow on the seafloor of the inter-tidal to shallow sub-tidal zone~ (Fig.~\ref{fig:ecosystems})~\citep{mtwana2016seagrass}.  Seagrass meadows are important ecosystems which play a multi-functional role -- they support fisheries, act as nurseries for marine species and provide a source of food for dugongs~\citep{mtwana2016seagrass, campagne2015seagrass}.

Beyond their role in supporting marine biodiversity, seagrass meadows act as a natural barrier, protecting shorelines from erosion caused by waves and currents~\citep{ondiviela2014role}. These habitats play a pivotal role in maintaining water quality by filtering pollutants and excess nutrients, thereby enhancing overall water clarity and promoting the health of adjacent marine ecosystems~\citep{campagne2015seagrass}. Furthermore, seagrass beds are significant contributors to global carbon sequestration efforts, as they capture and store carbon from the atmosphere in their biomass and sediment~\citep{macreadie2014quantifying}. While marine vegetated habitats, \ie seagrass meadows, mangroves, macroalgae and salt marches, comprise only 0.2\% of the ocean, they contribute an estimated 50\% of the carbon in marine sediments~\citep{duarte2013role}.

All seagrass meadows are important ecosystems, however certain types of seagrass provide different ecosystem services~\citep{mtwana2016seagrass}. Larger seagrass species, \eg \textit{Posidonia} and \textit{Zostera}, generally provide a wider range of ecosystem services, while smaller sized seagrass species such as \textit{Halophila} are the preferred food source for dugongs~\citep{mtwana2016seagrass}. In addition, the amount of carbon sequestered by different seagrass species can vary significantly~\citep{lavery2013variability}. Availability of information on which species are present and the distribution of the species is critical to ensure sound decision-making and management strategies for seagrass meadows, and accurate estimation of carbon stocks~\citep{mazarrasa2018habitat, lavery2013variability}.  An analysis of binary and species-level approaches is presented in Section~\ref{sec:lit-seagrass}. 

\subsection{Coral Reefs}
\label{subsec:lit-back-coral}
In this survey, we also consider coral reefs as a critical underwater ecosystem, and use this section to describe their characteristics, importance and the threats to reefs. Coral reefs are diverse and complex underwater ecosystems comprised of thousands of coral polyps which are fixed in place, forming the structure of the reef (Fig.~\ref{fig:ecosystems})~\citep{goreau1979corals}.  Reefs are characterized by their wealth of marine life and biodiversity, including coral, fish, molluscs, marine mammals and other organisms~\citep{sorokin2013coral}. Coral reefs protect coastlines from erosion by acting as natural barriers against waves and storms~\citep{reguero2018coral}. They support economies through tourism and fishing, and their worth is estimated as hundreds of billions of dollars annually~\citep{anthony2020interventions}.  However, coral reefs are increasingly threatened by human activities including climate change-induced ocean warming, increased ocean acidification, over-fishing, pollution, damage caused by tropical cyclones and predation by the Crown of Thorns Starfish (COTS)~\citep{anthony2020interventions, cornwall2021global}. There is a clear need for urgent and broad-scale interventions to prevent loss of coral reef ecosystems~\citep{anthony2020interventions}.

\footnotetext{These images are in the public domain and have been re-used without alteration from the \href{https://photolib.noaa.gov/About}{NOAA Photo Library}.}

The most critical intervention is climate change mitigation through reduction of greenhouse gas emissions~\citep{shaver2022roadmap}.  However there are many other policy and management decisions that can contribute to coral recovery including reducing nutrient run-off pollution, preventing over-fishing of certain species, management of predators~\citep{fabricius2010three}, improving coral biological resilience against increased temperatures~\citep{anthony2020interventions}, and coral re-seeding and gardening~\citep{doropoulos2019optimizing, mcleod2022coral}. All of these interventions require detailed data on coral species abundance to inform these strategies and monitor outcomes~\citep{ditria2022artificial, gonzalez2020monitoring}. Image analysis approaches for coral imagery are reviewed in Section~\ref{sec:lit-coral}.

\begin{figure}[t]
\centering
\setlength{\tabcolsep}{4pt}
\centerline{\begin{tabular}{cc}
    \includegraphics[height=35mm, clip, trim=0cm 0cm 0cm 0cm]{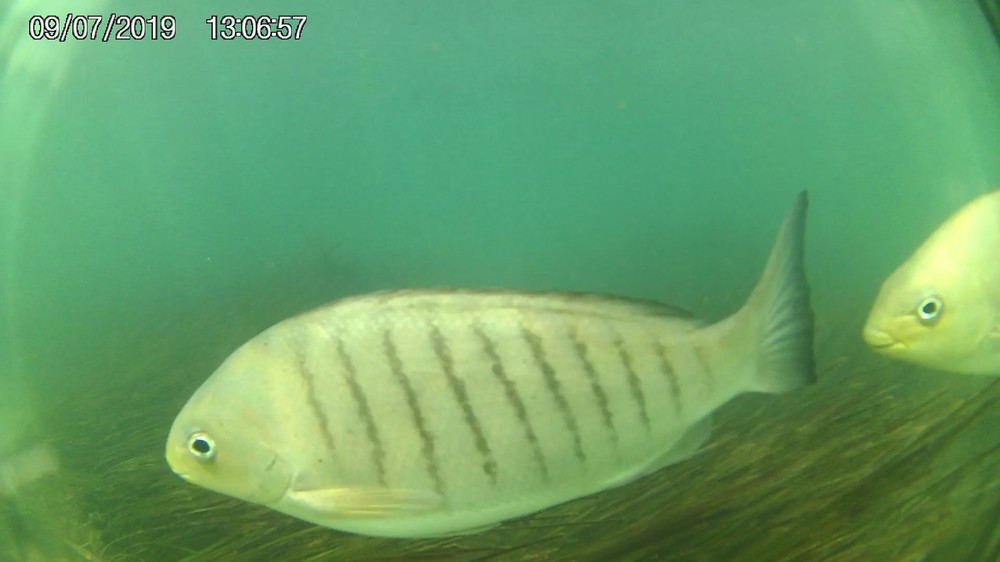} &
    \includegraphics[height=35mm]{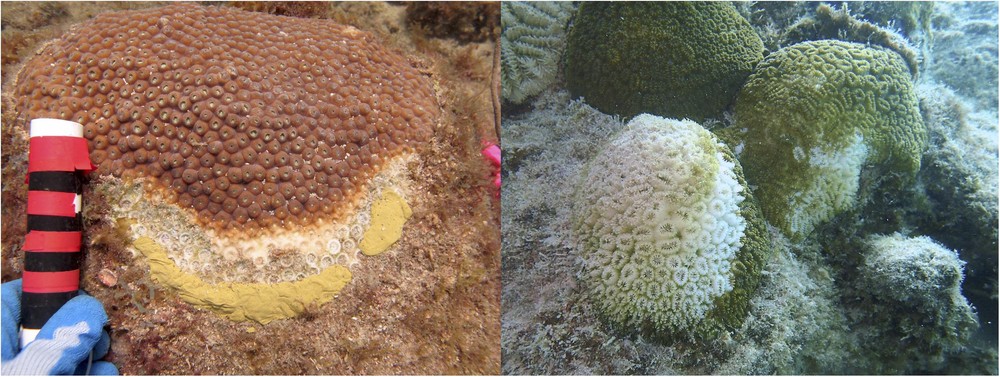} \\
    a) & b) \\
    \\
    \includegraphics[height=37mm, clip, trim=1cm 3cm 1cm 3cm]{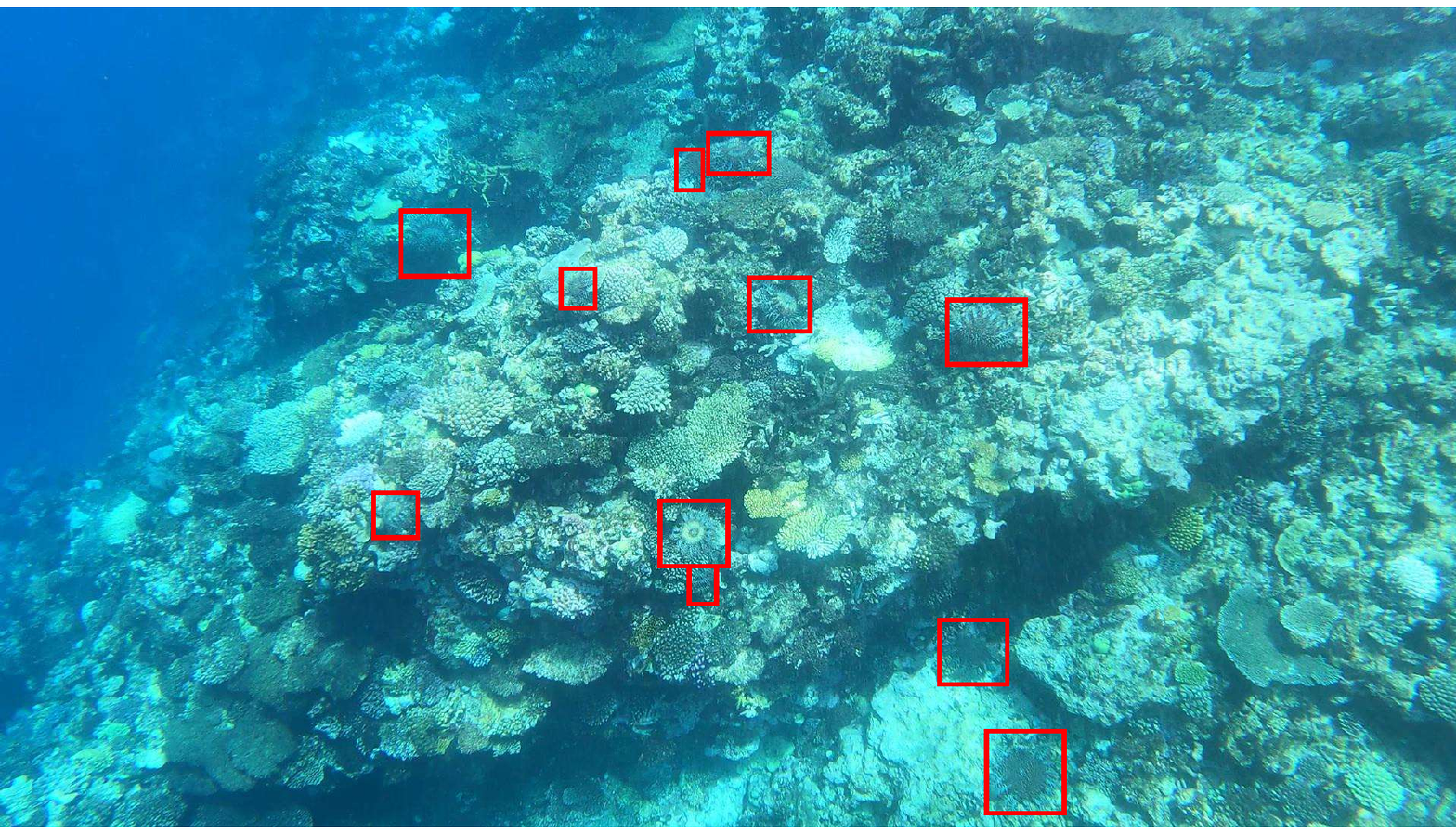} & 
    \includegraphics[height=37mm, clip, trim=0.5cm 0cm 0.5cm 0cm]{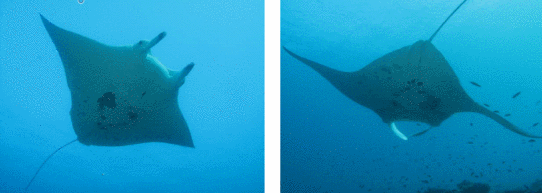} \\
    c) & d)\\
\end{tabular}}
\caption[Underwater Monitoring Tasks]{Examples\footnotemark~of underwater monitoring tasks, including: a) quantifying fish populations, b) detecting the Stone Coral Tissue Loss Disease (SCTLD), c) mapping the Crown-of-Thorns Starfish (COTS), and d) performing re-identification of manta rays.}
\label{fig:example-monitoring-tasks}
\end{figure}

\subsection{Importance of Marine Surveys}
\label{subsec:lit-marine-survey}
Marine surveys are a critical tool in the conservation of seagrass and reef ecosystems~\citep{murphy2010observational}.  This section motivates the importance of surveys and describes the variety of purposes and associated tasks.

\footnotetext{Image a) was released publicly as part of the `Global Wetlands' dataset~\citep{ditria2021annotated} under a \href{https://creativecommons.org/licenses/by/4.0/}{CC BY 4.0 licence}; image b) is reused without changes from~\cite{aeby2019pathogenesis}, Copyright © 2019 Aeby, Ushijima, Campbell, Jones, Williams, Meyer, H\"{a}se and Paul, released under the \href{https://creativecommons.org/licenses/by/4.0/}{CC BY licence}; image c) was publicly released in the `Underwater Crown-of-Thorn Starfish (COTS) imagery sample dataset'~\citep{liu2022underwater} released under a \href{https://creativecommons.org/licenses/by/4.0/}{CC BY 4.0 licence}; and image d) is reused without changes from~\cite{moskvyak2021robust} © 2021 IEEE.}

Management responses and policy decisions must be grounded in robust, accurate and up-to-date ecological information~\citep{gonzalez2020monitoring}.  The impact of natural and anthropogenic effects such as pollution, climate change, sea temperature increase and over-fishing must be monitored and evaluated over the long-term~\citep{xu2019deep, beijbom2015towards}. Surveys are used to determine ecosystem health indicators such as the biomass, density, size, habitat type and the distribution of fish and other marine species~\citep{murphy2010observational}. They can also monitor invasive algae~\citep{muntaner2023deep, gao2022algaenet}. Effective management of marine ecosystems requires available data at varying spatio-temporal scales~\citep{ditria2022artificial, lindenmayer2009adaptive}. Availability of long-term data enables quantification and tracking of responses to environmental changes, greater transparency of reef health and impacts of conservation approaches, and evidence-based policy and decision making~\citep{ditria2022artificial}.

Marine surveys also assist in providing the necessary information for active response strategies to a range of reef threats (Fig.~\ref{fig:example-monitoring-tasks}). The Stone Coral Tissue Loss Disease (SCTLD) was first detected in Miami, Florida in 2014 and is a devastating disease which kills entire colonies of stony corals within months~\citep{papke2024stony}.  It has since been detected in another 28 countries, and the spread of the disease is not yet fully understood~\citep{papke2024stony}.  Underwater monitoring over varying spatial and temporal scales is critical for early detection and management of the disease~\citep{combs2021quantifying, papke2024stony}.  In addition, detection and mapping of COTS enables teams of divers to be quickly sent to the most affected areas~\citep{li2022real}.  

Static remote cameras, sometimes baited to attract fish, collect images to be used for monitoring fish (Fig.~\ref{fig:example-monitoring-tasks}) or are equipped with machine or deep learning methods for fish detection~\citep{ditria2021annotated, saleh2023novel, boom2012supporting, boom2014research, sun2018transferring, wang2019underwater, spampinato2010automatic}. Surface vehicles automated with computer vision can also be used to collect coral spawn and then perform targeted distribution of corals to the right substrates, thus improving survival rates~\citep{mou2022reconfigurable, dunbabin2020uncrewed}. 

There has also been considerable research towards re-identification of whale~\citep{brooks2010seeing, wang2020method}, manta ray~\citep{moskvyak2021robust} and sun fish individuals~\citep{pedersen2022re} using machine learning and deep learning methods (Fig.~\ref{fig:example-monitoring-tasks}). While these are all important examples of types of marine surveys, this survey is focused primarily on automated analysis of underwater imagery for two key tasks: multi-species coral reef surveys and multi-species seagrass meadow surveys. We describe the different types of marine surveys methods in the following section.

\begin{figure}[t]
\centering
\setlength{\tabcolsep}{2pt}
\centerline{\begin{tabular}{ccc}
    \includegraphics[height=40mm]{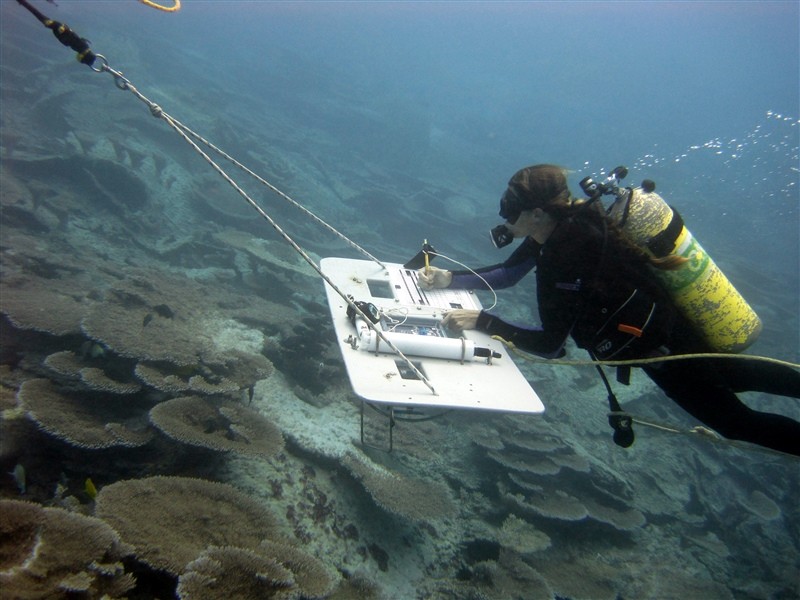} &
    \includegraphics[height=40mm]{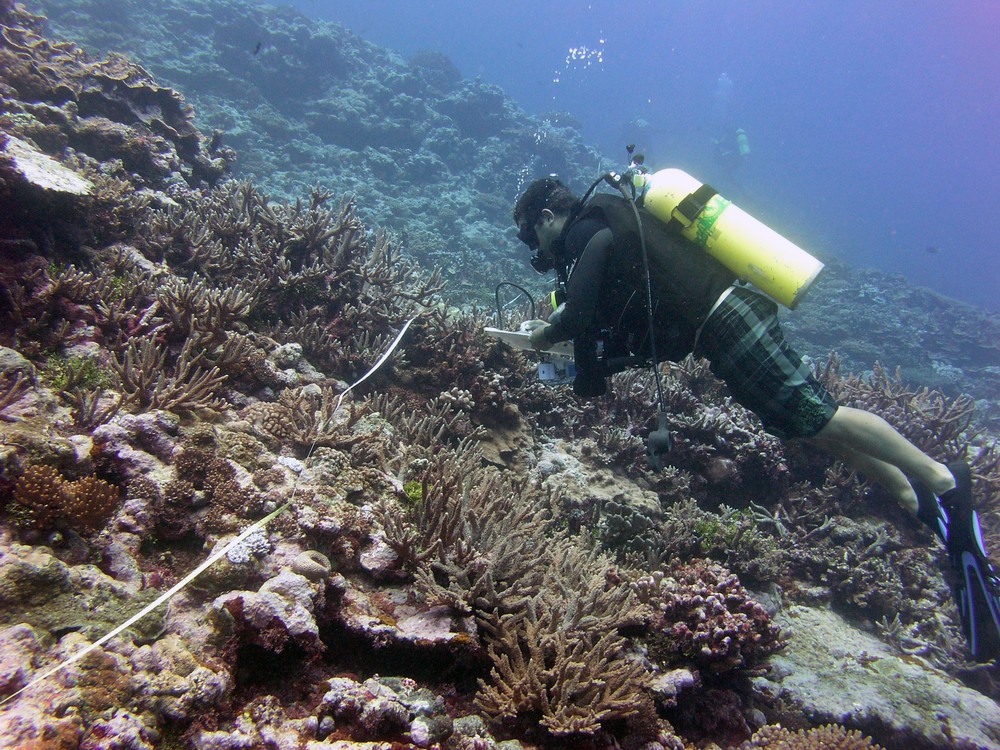} &
    \includegraphics[height=40mm]{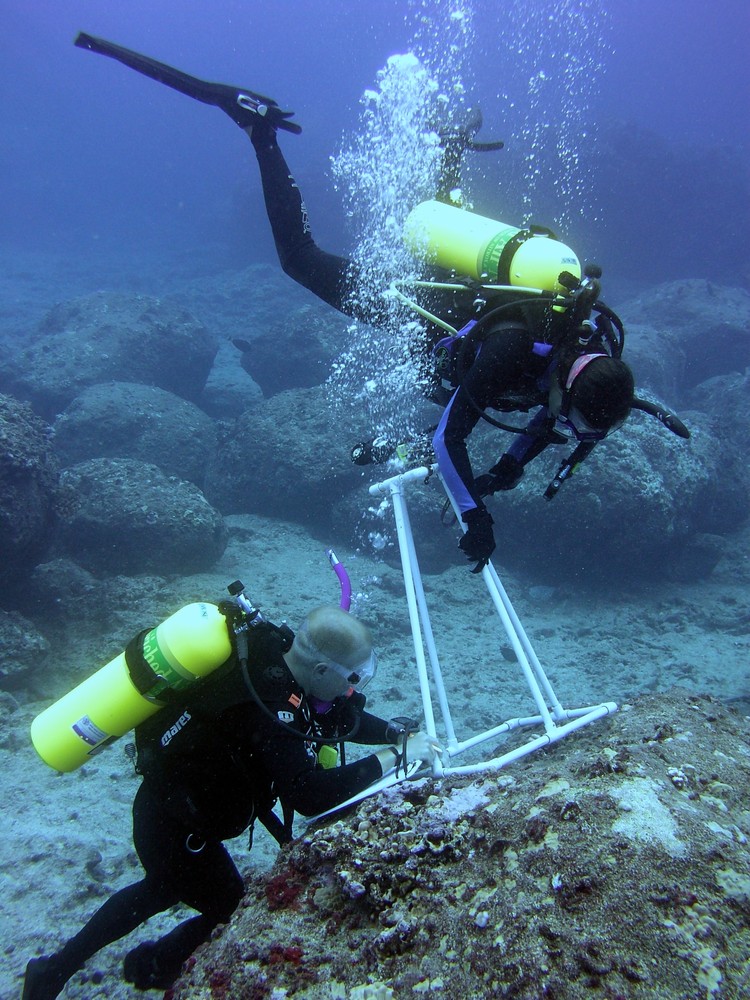} \\
    a) Manta tow & b) Transect & c) Photo-quadrat \\
\end{tabular}}
\caption[Examples of Diver Surveys]{Examples\footnotemark~of types of diver surveys, including: a) the manta tow technique, in which a diver observes the reef condition and structure while being towed by a manta board attached to a boat via a long rope; b) transect surveys, in which the diver makes visual observations at intervals along a linear transect tape; and c) photo-quadrat surveys, in which a diver collects photographs of the reef at regular intervals along a transect while using a highly controlled camera setup.}
\label{fig:diver-surveys}
\end{figure}

\footnotetext{These images are in the public domain and have been re-used without alteration from the \href{https://photolib.noaa.gov/About}{NOAA Photo Library}.}

\subsection{Marine Survey Methods}
\label{subsec:lit-marine-methods}
There are many different types of marine survey methods, ranging from manual survey methods performed by humans through to innovative applications of remote sensing and robotic technologies.  This section describes the common marine survey methods and discusses their benefits and limitations.

Divers play a critical role in marine monitoring and collect data through direct observation and/or photography. Manual diver surveys provide information on the percentage cover of hard corals and soft corals, measuring and monitoring populations of the Crown of Thorns Starfish, and detecting agents of coral mortality such as diseases or coral bleaching \citep{miller2018crown}.  Diver surveys can include manta tow, transect, photo-quadrat and SCUBA surveys (Fig.~\ref{fig:diver-surveys}). The manta tow survey technique involves towing a trained observer behind a tow vessel at low speed. The observer makes a visual assessment during a two minute tow period, and then records the observations on the manta board when the boat stops \citep{miller2018crown}.  Transect surveys involve a diver making observations at regular intervals along a linear reef transect of set distance, marked out by a tape measure \citep{jonker2020surveys}. Transect surveys can also be performed using photography, where images are collected at regular intervals along the transect. Photo-quadrat surveys are similar, but use a frame or structure to ensure photos are taken at a consistent height and scaled to a certain area. Photo-quadrat and transect images are later analysed using the CPC annotation technique to obtain estimates of the percent cover of benthic organisms present \citep{jonker2020surveys}.  Although diver surveys have been used widely for many decades, they are limited by availability of skilled personnel and dive time limits.

\begin{figure}[t]
\centering
\centerline{\includegraphics[width=140mm, clip, trim=0.2cm 1.6cm 1.4cm 2.2cm]{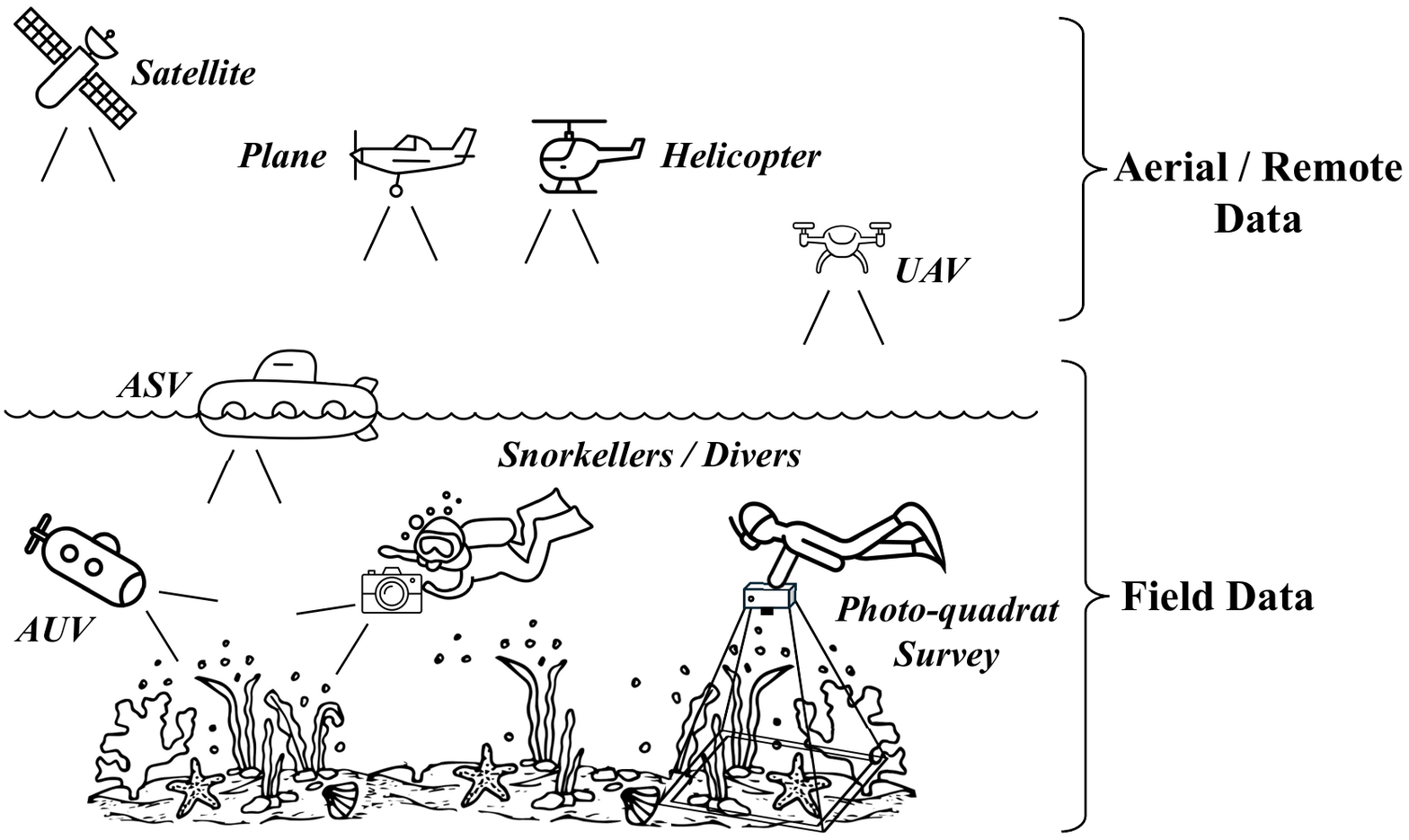}} 
\caption[Types of Monitoring Data]{A diagram visualising different common types of monitoring data\footnotemark. In this survey, the terms `aerial data' and `remote data' refer to monitoring imagery collected by Unmanned Aerial Vehicle (UAV), aircraft or satellite.  This review uses the term `field data' to encompass imagery collected by snorkellers, divers, underwater robots, surface vehicles and photo-quadrat surveys.}
\label{fig:data-types}
\end{figure}

\begin{figure}[t]
\centering
\centerline{\includegraphics[width=140mm, clip, trim=0cm 1.5cm 0cm 1.5cm]{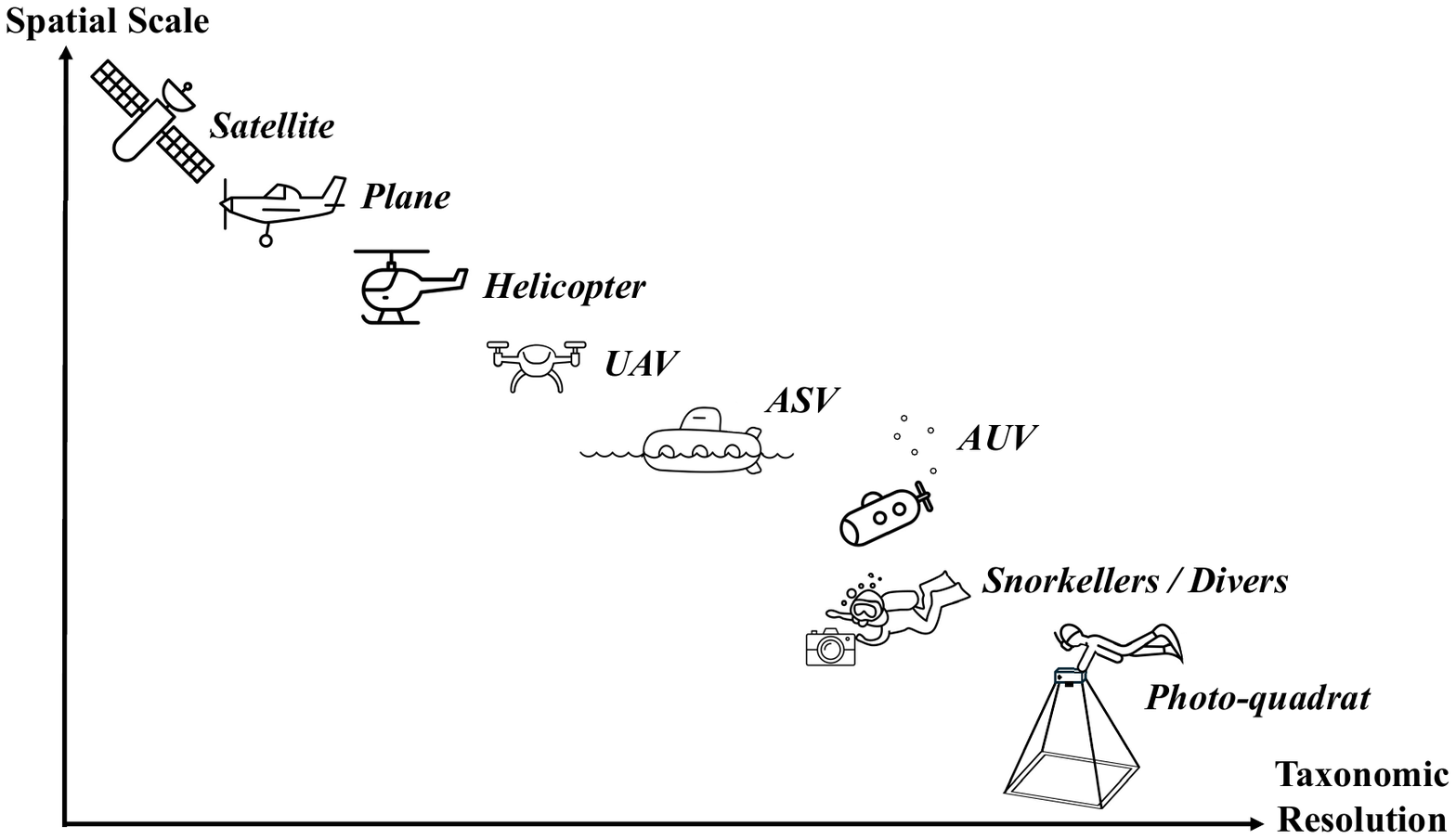}} 
\caption[Spatial Scale and Taxonomic Resolution of Survey Methods]{There are two factors to consider in comparing survey methods: spatial scale and taxonomic resolution. Spatial scale is the distance or area surveyed, while taxonomic resolution is the level of specificity at which organisms can be studied (high taxonomic resolution could be species level, low taxonomic resolution could be broad grouping \eg `coral', `substrate'). This diagram\footnotemark~visualises the spatial scale and taxonomic resolution for common monitoring methods. Photo-quadrat surveys offer the highest taxonomic resolution as they are static, controlled, close-range photos of the seafloor, while snorkellers, divers and Autonomous Underwater Vehicles (AUVs) offer similar resolution as they are dynamically moving through the water, typically within 1-2m of the seafloor.  AUVs and Autonomous Surface Vehicles (ASVs) offer greater spatial scale as they are not constrained to human or in-water limits, but slightly reduced taxonomic resolution as images may be taken at a greater distance from the seafloor, depending on operational conditions and water depth.}
\vspace{-0.2cm}
\label{fig:spatial-taxon}
\end{figure}

There has been research into use of remote sensing methods for mapping of coral reefs and seagrass meadows, specifically from imagery collected from UAVs, helicopters, planes or satellites (Fig.~\ref{fig:data-types})~\citep{hobley2021semi, roman2021using, tahara2022species, phinn2008mapping, roelfsema2010integrating}. These methods enable a large spatial area to be quickly surveyed, however they are typically only able to map the presence/absence of seagrass and are unable to discriminate between species (Fig.~\ref{fig:spatial-taxon})~\citep{reus2018looking}. The resolution can also be limited by changes in depth and the transparency of the water column~\citep{gonzalez2014catlin, lutzenkirchen2024exploring}.  In addition, maps generated with remote sensing imagery benefit from availability of finer-scale ground truth data for validation~\citep{roelfsema2010integrating, gonzalez2014catlin}.

`Field data' is a term used to describe observations, measurements or images collected directly in the natural environment, via transect surveys, spot checks, or use of autonomous underwater or surface vehicles (Fig.~\ref{fig:data-types})~\citep{roelfsema2010integrating}.  Analysis of underwater imagery collected in the field enables fine-grained mapping of species present, and also opens up opportunities for dynamic navigation and adaptation of survey paths~\citep{li2022real}.

\begin{figure}[t]
\centering
\setlength{\tabcolsep}{2pt}
\centerline{\begin{tabular}{cc}
    \includegraphics[height=32mm, clip, trim=9cm 0cm 5cm 0cm]{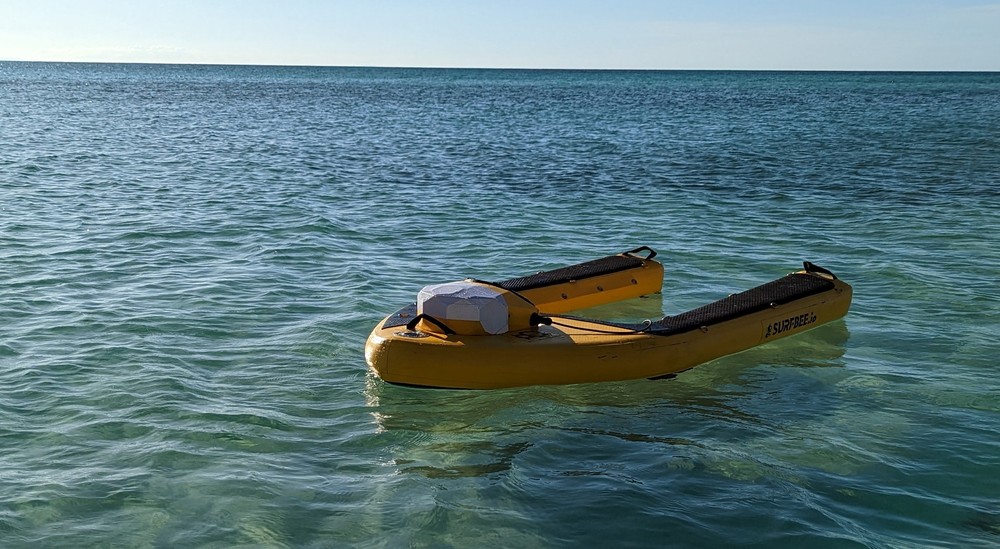} &
    \includegraphics[height=32mm]{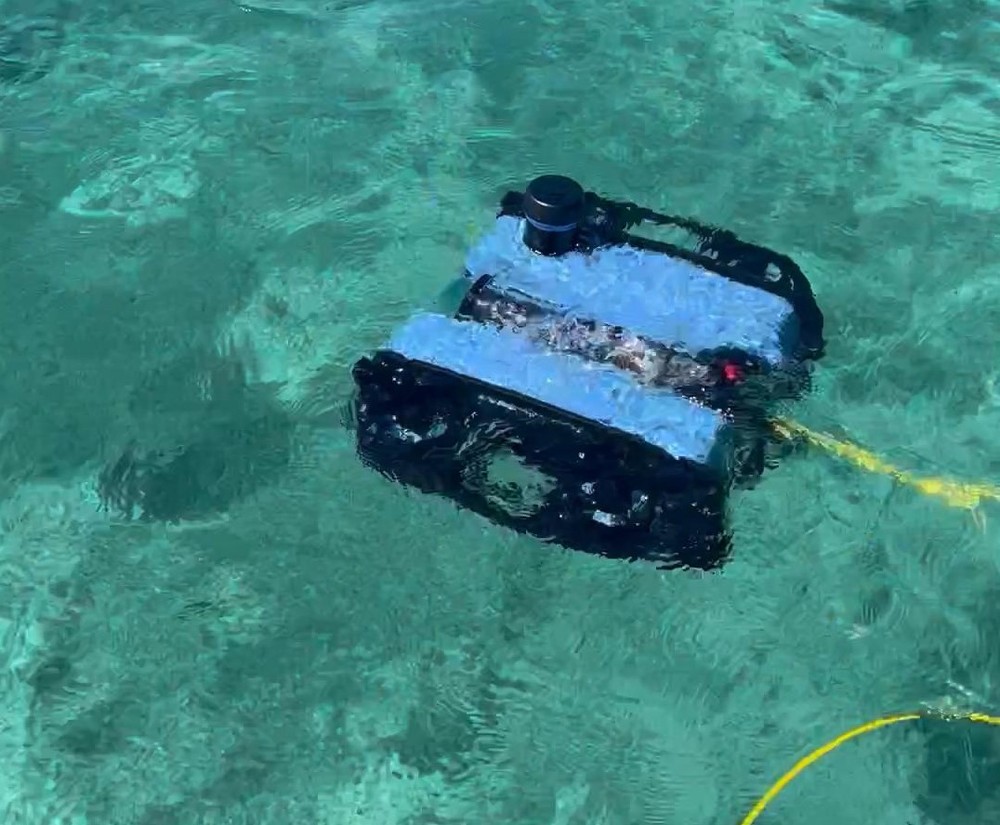} \\
    a) Autonomous Surface Vehicle & b) Remotely Operated Vehicle \\
    \citep{mou2022reconfigurable} & \\
    
    \includegraphics[height=32mm]{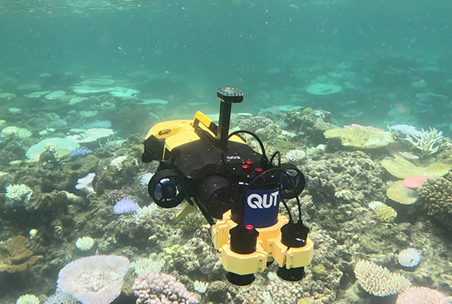} &
    \includegraphics[height=32mm]{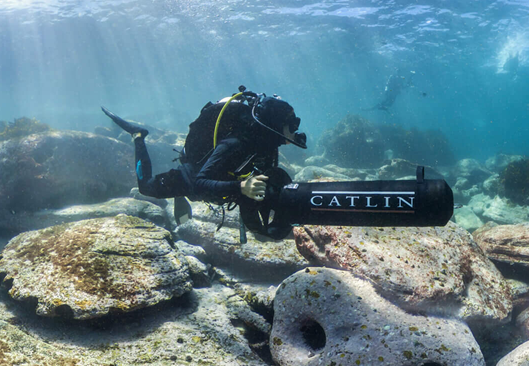} \\
    c) Autonomous Underwater Vehicle & d) Piloted Imaging System   \\
    \citep{dunbabin2020uncrewed}&\citep{gonzalez2014catlin, gonzalez2019seaview}
\end{tabular}}
\caption[Underwater Imaging Platforms]{Examples\footnotemark~of commonly used underwater imaging platforms, including a) an Autonomous Surface Vehicle, b) a Remotely Operated Vehicle, c) an Autonomous Underwater Vehicle and d) a Piloted Imaging System.}
\label{fig:example-imaging-systems}
\end{figure}

Recent technological advances in underwater imaging systems (Fig.~\ref{fig:example-imaging-systems}) include Remotely Operated Vehicles (ROVs)~\citep{sward2019systematic}, Towed Underwater Vehicles (TUVs)~\citep{bicknell2016camera}, Autonomous Surface Vehicles (ASVs)~\citep{mou2022reconfigurable}, Autonomous Underwater Vehicles (AUVs)~\citep{xu2019deep, monk2018marine, ferrari2018large} and long-range underwater gliders~\citep{gregorek2023long}.  These technologies are non-invasive and increase the extent, depth, efficiency and accuracy of marine surveys~\citep{murphy2010observational}. \footnotetext{The icons used to create this figure were created by Made, Olga, zaenul yahya, Amethyst Studio, Viral faisalovers, Isaac haq and Team iconify. They are released under \href{https://creativecommons.org/licenses/by/3.0/}{CC BY 3.0} and are available for download on
\url{https://thenounproject.com/}.} These platforms remove survey constraints caused by human diver limits and in-water time~\citep{williams2019leveraging}, and enable a greater level of taxonomic resolution~\citep{bryant2017comparison} (Fig.~\ref{fig:spatial-taxon}). Images collected can be used to ascertain marine species abundance and distribution~\citep{bacheler2015estimating}, perform substrate classification~\citep{gregorek2023long}, and can be used to build long-term archives of data for future analysis~\citep{williams2019leveraging}. Many of these vehicles are capable of collecting large quantities of videos or images, and thus can incur significant cost, time and effort if manually annotated~\citep{beijbom2015towards, xu2019deep}.  As a result, there is a need for automated image analysis approaches to process these large quantities of imagery~\citep{lutzenkirchen2024exploring}. The next section describes the characteristics of underwater imagery and outlines the availability of underwater datasets.

\section{Challenges in Underwater Imagery}
\label{sec:lit-underwater-imagery}

This section explores the unique challenges associated with computer vision for underwater imagery. First, the characteristics of underwater imagery are described in Section~\ref{subsec:lit-image-characteristics}. Section~\ref{subsec:lit-annotation} then provides a discussion of the challenges faced when annotating underwater imagery. Finally, Section~\ref{subsec:lit-underwater-datasets} reviews the publicly available datasets of underwater imagery.

\footnotetext{This figure was re-drawn and adapted from~\cite{bryant2017comparison}. The original is released under \href{https://creativecommons.org/licenses/by/3.0/}{CC BY 3.0}. The icons used to create this figure were created by Made, Olga, zaenul yahya, Amethyst Studio, Viral faisalovers, Isaac haq and Team iconify. They are released under \href{https://creativecommons.org/licenses/by/3.0/}{CC BY 3.0} and are available for download on
\url{https://thenounproject.com/}.}

\subsection{Image Characteristics}
\label{subsec:lit-image-characteristics}

This section describes the typical characteristcs of underwater imagery, with particular attention to how these characteristics impact the performance of automated image analysis approaches.  

Analysis of underwater imagery is challenging due to the dynamic, unrestricted environment of the ocean~\citep{jin2017deep}.  Typically images are deteriorated by noise, turbidity, scattering and attenuation of sunlight, low-light conditions, blur and changes in colouration due to depth~\citep{sun2018transferring, jin2017deep}, as seen in Fig.~\ref{fig:example-variation}.  For example, image characteristics vary significantly both within and between datasets due to lighting (whether artificial lighting or natural lighting is employed), camera settings used, and post-processing of images.  Furthermore, the type of imaging platform used for data collection (Fig.~\ref{fig:example-imaging-systems}) has a significant impact on the image appearance, due to changes in the camera angle (top-down, oblique or forward-facing), how close the image can be taken from the seafloor, and the movement and velocity of the platform through the water.

\footnotetext{Image a) is of the FloatyBoat ASV~\citep{mou2022reconfigurable} and was taken by Serena Mou and used with permission; image b) is of the `BlueRobotics' (\url{https://bluerobotics.com/}) ROV, and the image was taken by the first author; image c) is the `RangerBot' AUV~\citep{dunbabin2020uncrewed} and was taken by Serena Mou and used with permission; and image d) is available for educational reuse from \url{https://www.catlinseaviewsurvey.com/images-uwe} and is attributed to Underwater Earth, XL Catlin Seaview Survey and Christophe Bailhache.}

\begin{figure}[t]
\centering
\setlength{\tabcolsep}{1pt}
\centerline{\begin{tabular}{cc}
    \includegraphics[height=40mm]{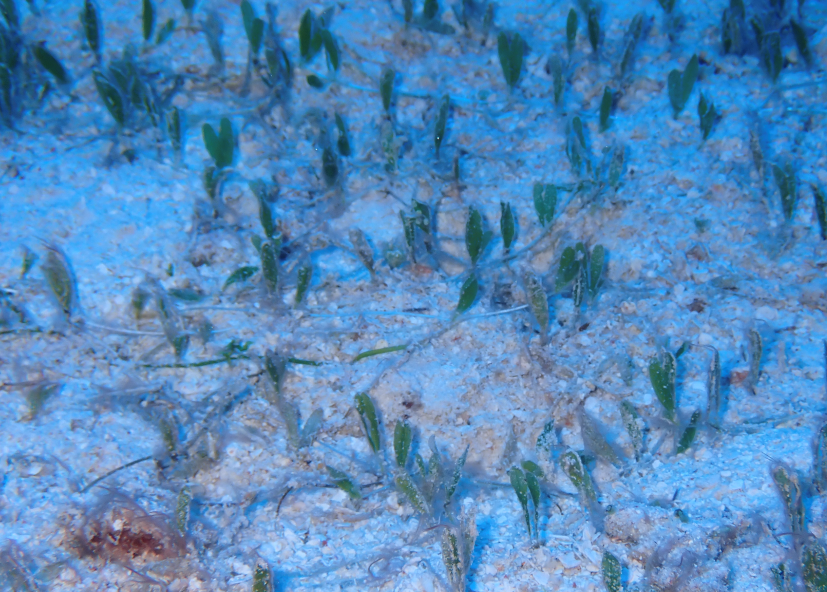} &
    \includegraphics[height=40mm]{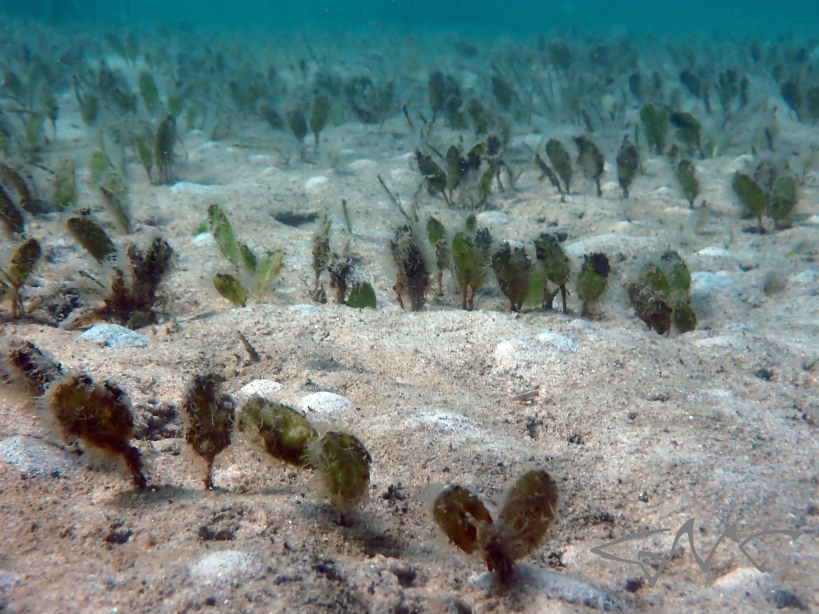} \\
    a) \textit{Halophila decipiens} & b) \textit{Halophila ovalis}\\
\end{tabular}}
\caption[Morphological Similarity of Seagrass Species]{Examples\footnotemark~of underwater images of seagrass species, demonstrating that different species can have visually similar characteristics.}
\label{fig:seagrass-species}
\end{figure}

For seagrass images, the season of image capture plays a significant role in the appearance as seasonality dictates the growth period and presence of algae~\citep{bonin-font2017visual}.  Seagrass images are also highly complex due to many overlapping leaves, and poor visibility and contrast~\citep{noman2023improving}.  In addition, many seagrass species are morphologically similar, making them visually difficult to distinguish~\citep{noman2023improving}, as seen in Fig.~\ref{fig:seagrass-species}.

Coral recognition is complicated by variation in the size, colour, texture and shape of corals within the same class, and by the difficulty in discerning boundaries between instances~\citep{beijbom2012automated, raphael2020deep}. The visual traits of corals are particularly challenging due to the plasticity of forms, lack of definition in shapes, intricacy of growth and the varying scale of discriminative features~\citep{raphael2020neural, gonzalez2020monitoring}. The label set used to annotate the data can introduce in-class variability as one label could describe several morphologically diverse species~\citep{gonzalez2020monitoring}. \footnotetext{Images re-used without alteration.  a) taken by Adam Smith and b) taken by Tony Strazzari. Both released under \href{https://creativecommons.org/licenses/by-nc/4.0/}{CC BY 4.0 licence} on the \href{https://bie.ala.org.au/search?q=halophila}{Atlas of Living Australia}.} The environmental conditions impact the morphology of coral reef benthos, as increased depth encourages growth structure to maximise light capture~\citep{gonzalez2020monitoring} and the edge of the reef, whether windward or leeward, impacts growth habits~\citep{jell1978guide}. 

The image characteristics discussed in this section are specific to analysis of underwater imagery, and therefore the task of automated underwater image analysis requires targeted approaches able to overcome these challenges. In the next section, we discuss how underwater image characteristics also have implications for data annotation, and the other considerations and difficulties of labelling data in this domain. 

\subsection{Data Annotation}
\label{subsec:lit-annotation}

The difficulty in correctly identifying the species present and the boundaries of each instance makes it necessary for domain experts to label underwater imagery for training deep learning models~\citep{zhang2024cnet, gonzalez2023survey}.  For precise evaluation of deep learning approaches, it is important that the labels are accurate and representative of the data. It can be difficult to find domain experts with sufficient time for data annotation, and then recompense their time~\citep{williams2019leveraging, katija2022fathomnet}. Specifically, it takes a domain expert more than 30 minutes to annotate a single coral image with 200 random points~\citep{mahmood2016automatic, beijbom2012automated}. By contrast, it takes on average 9 minutes to label a COCO image (which contains common objects) with a dense pixel-wise segmentation mask~\citep{cheng2022pointly}. Furthermore, the level of expertise required to label underwater imagery prevents usage of common computer vision tools such as the crowd-sourcing platform, Amazon Mechanical Turk~\citep{katija2022fathomnet}. 

In addition, labelling underwater imagery often has low annotator agreement or high `Inter-Observer Variability'~\citep{gonzalez2016scaling, friedman2013automated}, which is the degree of consensus amongst data annotators.  If the inter-observer variability is high, then a group of annotators labelling the same data sample may not label the data in the same way.  One study~\citep{gonzalez2016scaling} found the inter-observer error ranged between 1\% and 7\% for estimation of benthic composition for species of hard corals, soft corals and algae.  For underwater imagery, this is due to the difficulty and high level of expertise required to accurately identify the species present, and is also caused by low image quality, \ie blurry, turbid or poorly lit imagery which obscures the target and removes colour information~\citep{seiler2012image}.  

Conversely, traditional computer vision applications, \eg for autonomous vehicles or urban scenes, will typically exhibit high annotator agreement, as targets in images are more obvious and less ambiguous \citep{braylan2022measuring}.  Furthermore, the class labels used as more distinct and readily understood (apple \vs chair, road \vs sky; as opposed to differentiating between fine-grained marine species which may appear differently throughout the year and based on geographic location).  The combination of these factors mean that labelled datasets of underwater imagery are rare~\citep{zhang2024cnet} (this is further discussed in Section~\ref{subsec:lit-underwater-datasets} and Appendix~\ref{appendix:A}), and that computer vision for analysis of underwater imagery is a unique and challenging task~\citep{mahmood2017deep}.

There is one type of labelled underwater imagery which is more common: `photo-quadrat' data.  Photo-quadrat transect surveys involve taking photos of 1m by 1m quadrats at intervals along a linear transect of the reef or seagrass meadow~\citep{english1997survey}. Photo-quadrat images can be taken by divers, snorkellers or guided underwater platforms~\citep{gonzalez2019seaview}.  The transect surveys can be repeated at regular intervals in time to assess changes in reef health.  A frame can be used to take photos at a consistent height and angle along the transect~\citep{english1997survey}.  These photos are often labelled by ecologists using the CPC point-based method (Fig.~\ref{fig:example-cpc}), which involves assigning class labels to a certain number of points (\eg 100) randomly placed or spaced as an even grid over the image~\citep{kohler2006coral, murphy2010observational}, as seen in Fig.~\ref{fig:data-types}.  

The distinctions and difficulties identified in this section have implications for the availability of underwater datasets.  This is analysed in the next section, which performs a comprehensive review of underwater datasets and data platforms.

\subsection{Underwater Datasets and Platforms}
\label{subsec:lit-underwater-datasets}

Numerous reef data platforms have been developed in recent years \eg CoralNet\footnote{\url{https://coralnet.ucsd.edu/}}, Squidle+\footnote{\url{https://squidle.org/}}, ReefCloud\footnote{\url{https://reefcloud.ai/}}, Data Mermaid\footnote{\url{https://datamermaid.org/}} and FathomNet\footnote{\url{https://fathomnet.org/}}. These platforms are largely intended for storage and annotation of images, and visualisation of data~\citep{beijbom2012automated}.  However, as users upload and annotate their own data, the accuracy of the labels is unverified and some platforms only allow limited access to certain datasets.  

Apart from these platforms, there are other publicly available underwater datasets, however they are highly variable in the style and viewpoint of the imagery, type and quantity of annotations, number of classes and species present, and the amount of imagery. Availability of high quality data is important because demonstrating novel approaches on publicly available datasets with accompanying reliable ground truth labels enables recreation and verification of presented results. A survey of the publicly available datasets of underwater imagery is compiled in Appendix~\ref{appendix:A}. Specifically, seagrass datasets are analysed in Table~\ref{table:datasets-seagrass} and coral datasets are summarised in Table~\ref{table:datasets-coral}.

Our dataset survey indicated a shortage of multi-species datasets for seagrass imagery, with only one dataset providing multiple taxonomic morphotypes of seagrasses.  This dataset is called DeepSeagrass~\citep{raine2020multi} and contains 66,946 single-species patch images labelled into four classes. 

Our survey of coral datasets identified an abundance of photo-quadrat data annotated with the Coral Point Count method.  Many datasets of this type have been made publicly available, as seen in Appendix~\ref{appendix:A}, Table~\ref{table:datasets-coral}~\citep{roelfsema, beijbom2015towards, beijbom2012automated, alonso2019coralseg}.  Photo-quadrat data is typically collected by ecologists for estimating the percentage cover of species, and it is therefore not designed explicitly for training deep learning models for segmentation of underwater imagery. Point label propagation is a technique used to leverage this type of data, and we therefore review point label propagation algorithms in Section~\ref{subsubsec:lit-coral-weak}.

\begin{figure}[t]
\centering
\setlength{\tabcolsep}{10pt}
\centerline{\begin{tabular}{cc}
    \includegraphics[height=40mm]{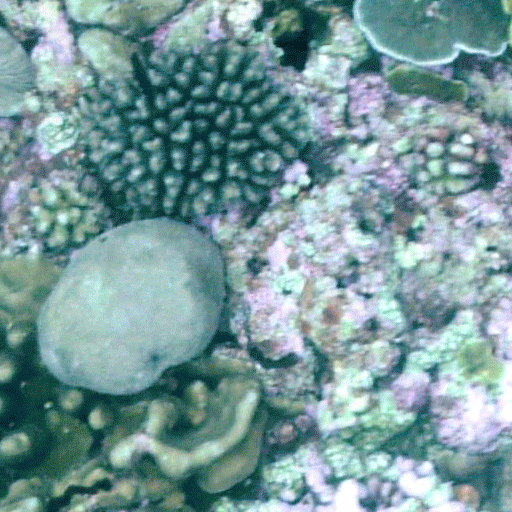} &
    \includegraphics[height=40mm]{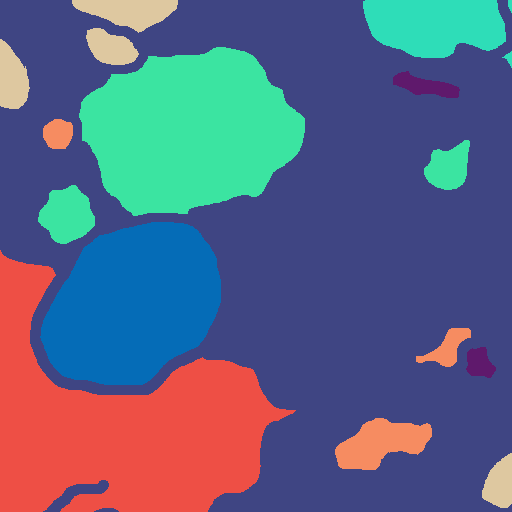} \\
    a) Example Image 1 & b) Ground Truth Mask for Image 1 \\
    \\
    \includegraphics[height=40mm]{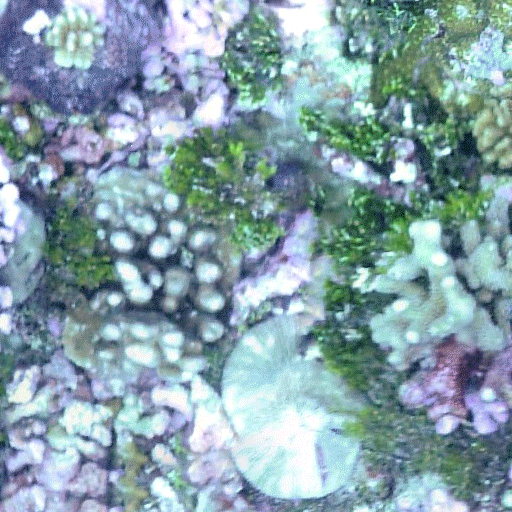} &
    \includegraphics[height=40mm]{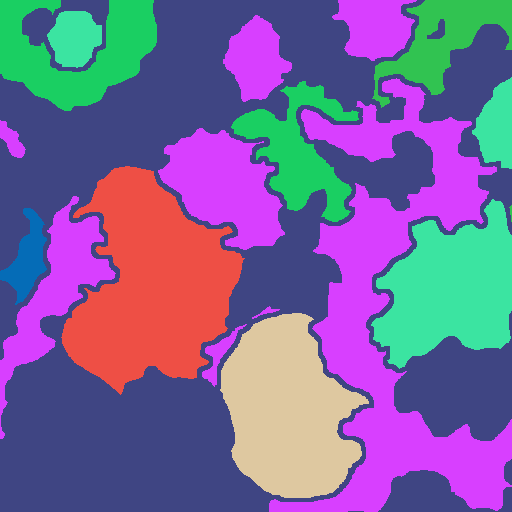} \\
    c) Example Image 2 & d) Ground Truth Mask for Image 2 \\
\end{tabular}}
\caption[UCSD Mosaics Dataset: Sample Images and Ground Truth Masks]{Sample images and ground truth masks publicly released as part of the UCSD Mosaics dataset\footnotemark~\citep{edwards2017large, alonso2019coralseg}.  This is an important dataset of multi-species coral images, accompanied by expert-labelled pixel-wise ground truth masks.}
\label{fig:ucsd-examples}
\end{figure}

While there is an abundance of point-labelled coral imagery, there is a shortage of data labelled with dense pixel-wise ground truth masks. The only publicly available dataset containing fine-grained multi-species coral images accompanied by dense ground truth masks is the University of California San Diego (UCSD) Mosaics dataset, originally from~\citep{edwards2017large} but re-released in~\citep{alonso2019coralseg} and cleaned of corrupted examples in~\citep{raine2024human}. Example images and ground truth masks are in Fig.~\ref{fig:ucsd-examples}. 

Other datasets outlined in Appendix~\ref{appendix:A} include the SUIM~\citep{islam2020semantic} dataset, which contains coarse-grained object classes such as human divers, plant, wrecks, robots, reef etc., and the `\#DeOlhoNosCorais' dataset is comprised of imagery obtained from the social media platform Instagram~\citep{furtado2023deolhonoscorais}. The quality, consistency and style of imagery varies significantly within the dataset and is substantially different to photo-quadrat imagery and imagery collected by underwater platforms.  The recent CoralVOS dataset~\citep{ziqiang2023coralvos} is a promising development in availability of video sequence data, however the pixel-wise annotations are only provided for the binary presence/absence of corals. Although the availability of underwater datasets is gradually improving, the datasets providing dense ground truth for multiple species remain limited, largely due to the requirement of annotation by domain experts~\citep{gonzalez2023survey}. 

This section has reviewed the underwater datasets in the literature and highlighted the gaps and limitations of the data available.  We provide further analysis of underwater data and recommendations for next steps in Section~\ref{subsubsec:disc-data}.

In the following sections, we step away from the underwater domain and provide background on the relevant computer vision tasks (Section~\ref{sec:lit-back-CV}) and deep learning architectures (Section~\ref{sec:lit-types-dl}). 

\section{Background on Computer Vision Tasks}
\label{sec:lit-back-CV}

Computer vision is a field of computer science which aims to analyse and understand images. This section describes some of the relevant tasks in computer vision for this survey. We first describe the image classification task (Section~\ref{subsec:lit-cls}), followed by patch classification (Section~\ref{subsec:lit-patch-cls}), coarse segmentation (Section~\ref{subsec:lit-coarse-seg}), semantic segmentation (Section~\ref{subsec:lit-seg}), and finally weakly supervised segmentation (Section~\ref{subsec:lit-weak-seg}).  

\subsection{Image Classification}
\label{subsec:lit-cls}

\footnotetext{These images are reproduced with permission from Springer Nature.}

\begin{figure}[t]
\centering
\setlength{\tabcolsep}{4pt}
\centerline{\begin{tabular}{ccccc}
    \includegraphics[height=30mm, clip, trim=11cm 8.5cm 11cm 8cm]{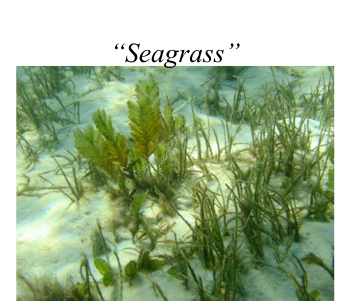} &
    \includegraphics[height=26mm, clip, trim=10.5cm 8.5cm 10.5cm 8cm]{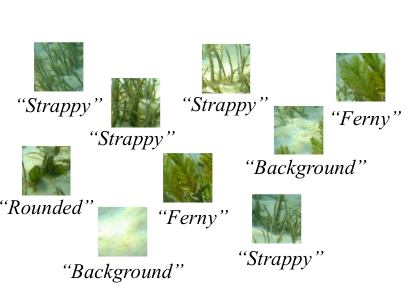} &
    \includegraphics[height=26mm, clip, trim=10.5cm 9cm 11.5cm 8cm]{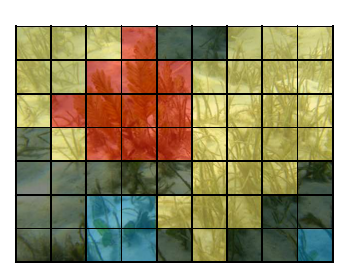} & 
    \includegraphics[height=23.5mm]{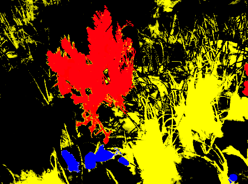} &
    \includegraphics[height=26mm, clip, trim=12cm 9cm 11cm 8cm]{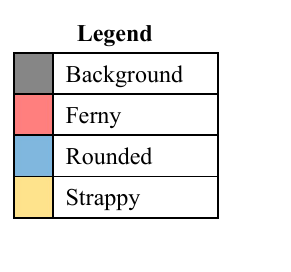} \\
    a) Image Classification & b) Patch Classification & c) Coarse Segmentation & d) Semantic Segmentation & \\
\end{tabular}}
\caption[Tasks in Computer Vision]{A comparison between computer vision tasks\footnotemark: image classification (a), in which the whole image is assigned a class label; patch classification (b) which involves classifying one small crop at a time; grid-based coarse segmentation (c) which describes inference as a grid over the whole image; and semantic segmentation (d) which is the task of predicting the class of every pixel in an input image, therefore partitioning the image into semantically meaningful regions.}
\label{fig:patch-grid}
\end{figure}

Image classification is the task of predicting the class label of an entire image, as seen in Fig.~\ref{fig:patch-grid} \citep{goodfellow2016deep}.  This task is not the focus of this survey, but included here for clarity. Image classification yields one label for an entire image, and does not provide any localisation of where the class occurs in the image.  Additionally, for underwater imagery, it does not provide any indication of the coverage of the class in the image.

\subsection{Patch Classification}
\label{subsec:lit-patch-cls}

Patch classification is the task of predicting the class label of a small sub-image crop of an original image.  In this survey, we define patches as rectangular or square crops taken from the original image by dividing the image into a grid with cells of equal size (Fig.~\ref{fig:patch-grid}). A classifier is trained on the patches and the class of the patch is predicted. The entire image is then analysed at inference time, as each cell in the original image is treated as a patch. 

Patch classification enables localisation of where the class is in the image and also facilitates estimation of species coverage based on how many patches in an image are classified as the species.  Patch classification also enables multiple classes to be predicted within the same image. Although patch classification improves on image classification with sub-image localisation, the individual patches are classified independently, meaning that any image context from outside the current patch is not taken into consideration during training and inference.

\footnotetext{Original image by Ellen Ditria and used with permission.}

\subsection{Coarse Segmentation}
\label{subsec:lit-coarse-seg}

\begin{figure}[t]
\centering
\setlength{\tabcolsep}{1pt}
\centerline{\begin{tabular}{cc}
    \includegraphics[height=40mm]{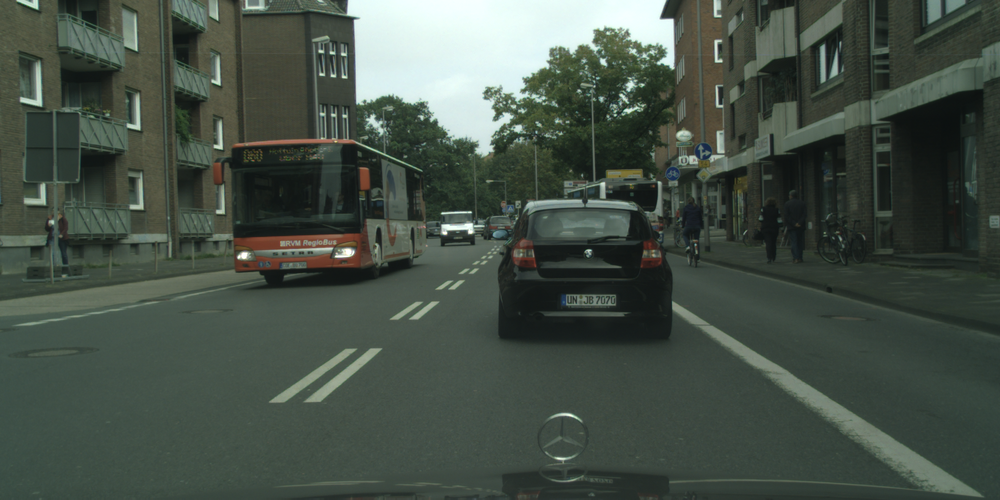} &
    \includegraphics[height=40mm]{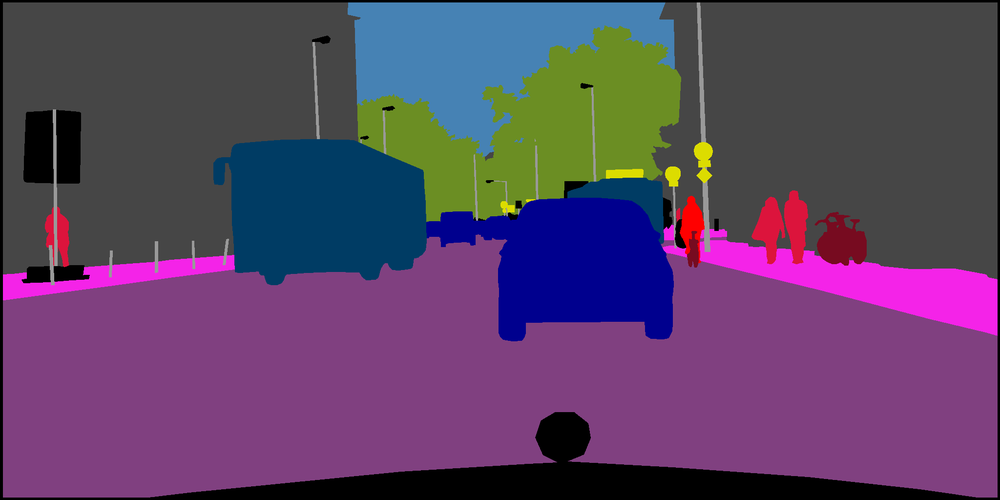} \\
    a) Cityscapes Image~\citep{cordts2016cityscapes} & b) Cityscapes GT Mask~\citep{cordts2016cityscapes} \\
    \\
    \includegraphics[height=42mm]{fig13_c.png} &
    \includegraphics[height=42mm, clip, trim={0 0 0 0.1cm}]{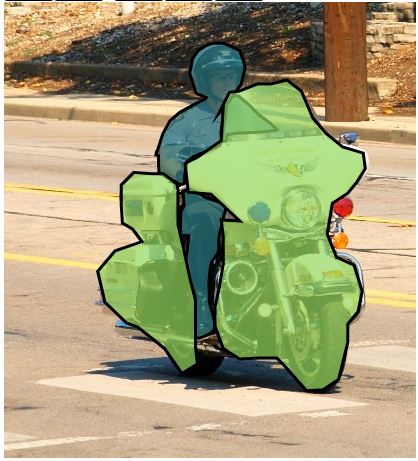} \\
    c) MS COCO Image~\citep{lin2014microsoft} & d) MS COCO GT Mask~\citep{lin2014microsoft} \\
\end{tabular}}
\caption[Examples of Typical Semantic Segmentation Images and Ground Truth Masks]{Sample images\footnotemark~and ground truth masks from the Cityscapes~\citep{cordts2016cityscapes} and MS COCO~\citep{lin2014microsoft} dataset, demonstrating clear boundaries of objects in images.}
\label{fig:regular-seg-examples}
\end{figure}

Coarse segmentation is similar to patch classification, but extends on patch classification as whole image context can be utilised during training and inference.  Although individual patches are assigned a class label, the entire image can be considered by the coarse segmentation approach (Fig.~\ref{fig:patch-grid}). 

\subsection{Semantic Segmentation}
\label{subsec:lit-seg}

\begin{figure}[t]
\centering
\setlength{\tabcolsep}{0.5pt}
\centerline{\begin{tabular}{ccc}
    \includegraphics[height=50mm,clip, trim=10.5cm 7.5cm 10.5cm 7.5cm]{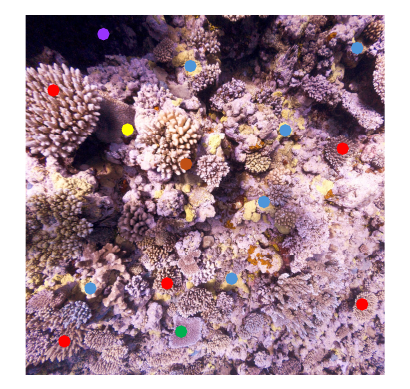} &
    \includegraphics[height=50mm,clip, trim=10.5cm 7.5cm 10.5cm 7.5cm]{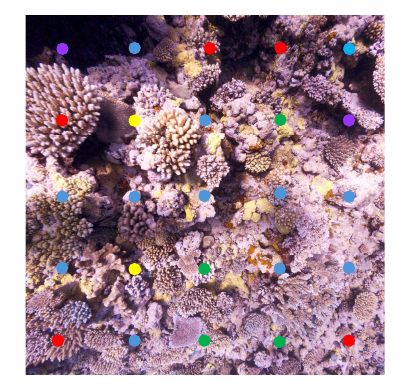} &  
    \includegraphics[height=50mm,clip, trim=10.5cm 7.5cm 10.5cm 7.5cm]{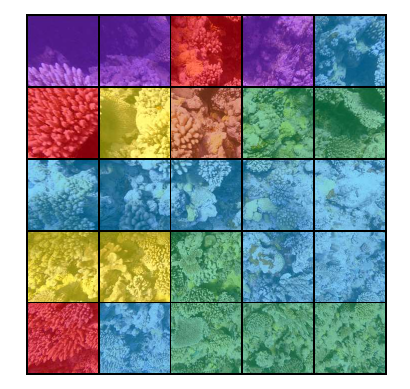}\\
    a) Random Points & b) Grid Points & c) Grid Cells \\
\end{tabular}}
\caption[Underwater Imagery Labelling Styles]{Common labelling styles used in underwater image datasets\footnotemark. The CPC annotation method involves labelling randomly-distributed (a) or grid-spaced (b) points. Some datasets also perform data annotation by assigning labels to patches based on the dominant class (c).}
\label{fig:label-types}
\end{figure}

Semantic segmentation describes the task of predicting the class of every pixel in a query image (Fig.~\ref{fig:patch-grid})~\citep{lateef2019survey}.  This is a common task in the field of computer vision, with considerable research effort dedicated to a variety of segmentation applications~\citep{liang2020weakly}, including segmentation of street scenes for autonomous driving applications~\citep{li2018real}; satellite and aerial imagery for land cover classification or environmental monitoring~\citep{kemker2018algorithms}; UAV and ground imagery for agricultural applications~\citep{milioto2018real}; and scenes for task planning and navigation in robotics~\citep{gupta2015indoor}. Fully supervised semantic segmentation is when a model is trained from image-mask pairs, where the mask contains a ground truth class label for every pixel in the corresponding image~\citep{zhang2019survey} (Fig.~\ref{fig:regular-seg-examples}). Typically these masks are created by humans manually inspecting the image and assigning class labels to each pixel, a laborious and time-consuming task~\citep{zhang2019survey}.  In mainstream computer vision, these masks can be labelled at scale by crowd-sourcing platforms such as Amazon Mechanical Turk.  This is because it is typically straightforward to identify objects in images and there is high annotator agreement between the labelled masks (as seen in Fig.~\ref{fig:regular-seg-examples}). This is not the case in the underwater domain, where images often contain many overlapping instances and forms can be difficult to distinguish, as outlined in Section~\ref{subsec:lit-image-characteristics}.

\footnotetext{Image a) was released publicly under the \href{https://www.cityscapes-dataset.com/license/}{Cityscapes}~\citep{cordts2016cityscapes} licence. Image b) was released publicly in the MS COCO~\citep{lin2014microsoft} dataset under \href{https://creativecommons.org/licenses/by/4.0/legalcode}{CC BY 4.0}.} 

\subsection{Weakly Supervised Semantic Segmentation}
\label{subsec:lit-weak-seg}

Weakly supervised semantic segmentation describes a model trained from weak labels which still outputs a dense pixel-wise mask at inference time~\citep{zhang2019survey}.  As compared to dense, pixel-wise ground truth masks in fully supervised segmentation, weakly supervised segmentation aims to reduce the amount of time and effort it takes to produce annotations used for training the model~\citep{liang2020weakly}.  The weak training labels could take the form of bounding boxes, polygons, scribbles, point labels or whole image labels~\citep{zhang2019survey, hua2021semantic, vernaza2017learning}. In underwater datasets, weak labels could take the form of grid points, random points or grid cells. Grid points are pixels arranged in a regular grid, where the species at the pixel location is labelled; random points are randomly distributed across the image and then labelled by the domain expert; and grid cells is where a grid is drawn over the image and each cell is labelled as the majority species. These annotation styles are shown in Fig.~\ref{fig:label-types}.  Note that the class of the grid point is not necessarily the same class as what would be assigned if using the grid cell labelling style.

\footnotetext{Image used in this figure was released publicly as part of the XL CATLIN Seaview Survey dataset~\citep{gonzalez2014catlin, gonzalez2019seaview} under \href{https://creativecommons.org/licenses/by/3.0/deed.en}{CC BY 3.0}.}

Section~\ref{sec:lit-back-CV} provided an overview of the relevant computer vision tasks.  The following section describes deep learning models which are trained to perform these computer vision tasks.

\section{Types of Deep Learning Models}
\label{sec:lit-types-dl}

In the previous section, various types of computer vision tasks for automated analysis of underwater imagery were outlined and described. This survey is focused on approaches which perform these tasks using deep learning.  This section therefore reviews relevant deep learning concepts and key architectures.  Deep learning is a type of artificial intelligence where large neural networks or `models' are trained to learn complex, meaningful representations and patterns in the data from a hierarchy of simpler concepts~\citep{goodfellow2016deep}. The deep learning system draws on the representations learned to perform some task.  This section provides background on Convolutional Neural Networks (CNNs) and Transformers, two key deep learning architectures which can be used to perform automated analysis of underwater imagery.

\subsection{Convolutional Neural Networks}
\label{subsec:lit-cnns}

\renewcommand{\topfraction}{.75}
\begin{figure}[t]
\centering
\setlength{\tabcolsep}{2pt}
\centerline{\begin{tabular}{cc}
    \multicolumn{2}{c}{\includegraphics[height=50mm, clip, trim=2cm 4cm 1.5cm 2.8cm]{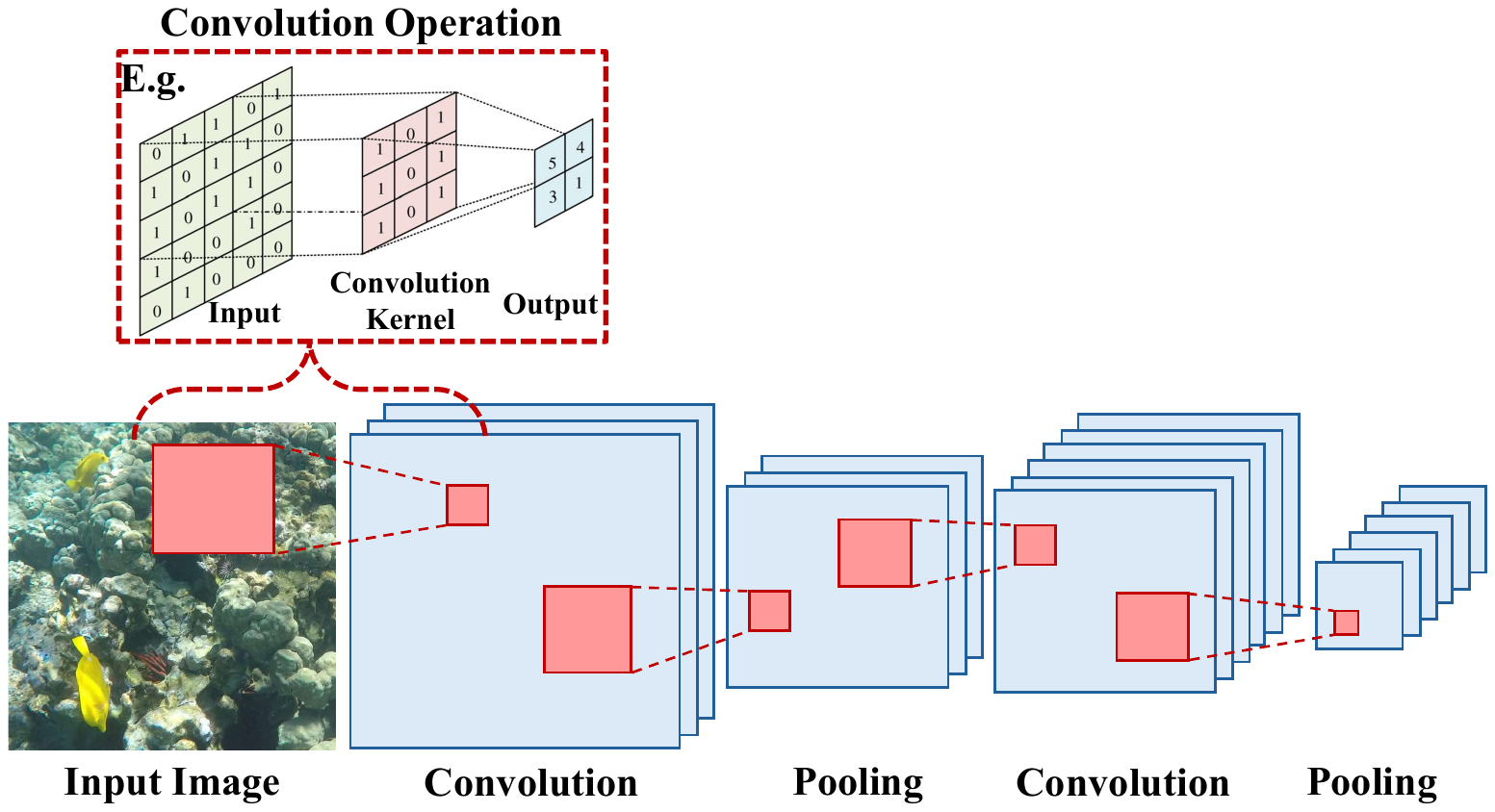}} \\
    \multicolumn{2}{c}{a) Convolution and Pooling Operations } \\
    \includegraphics[height=36mm, clip, trim=4cm 6cm 4cm 5cm]{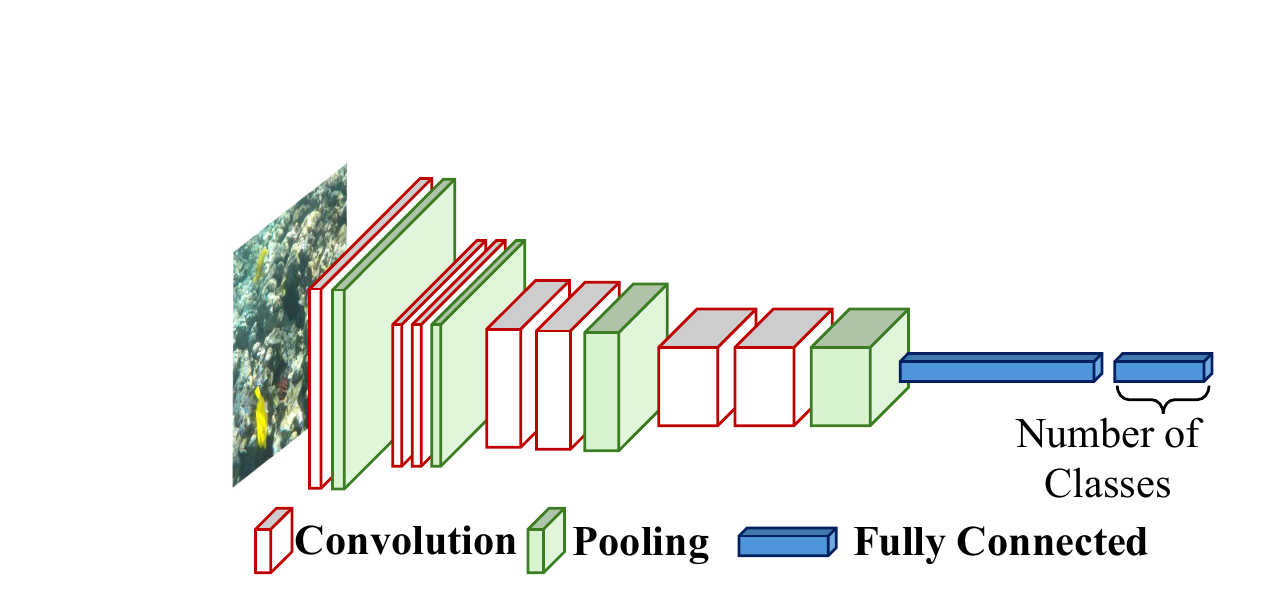} &
    \includegraphics[height=36mm, clip, trim=5cm 6.5cm 2cm 5cm]{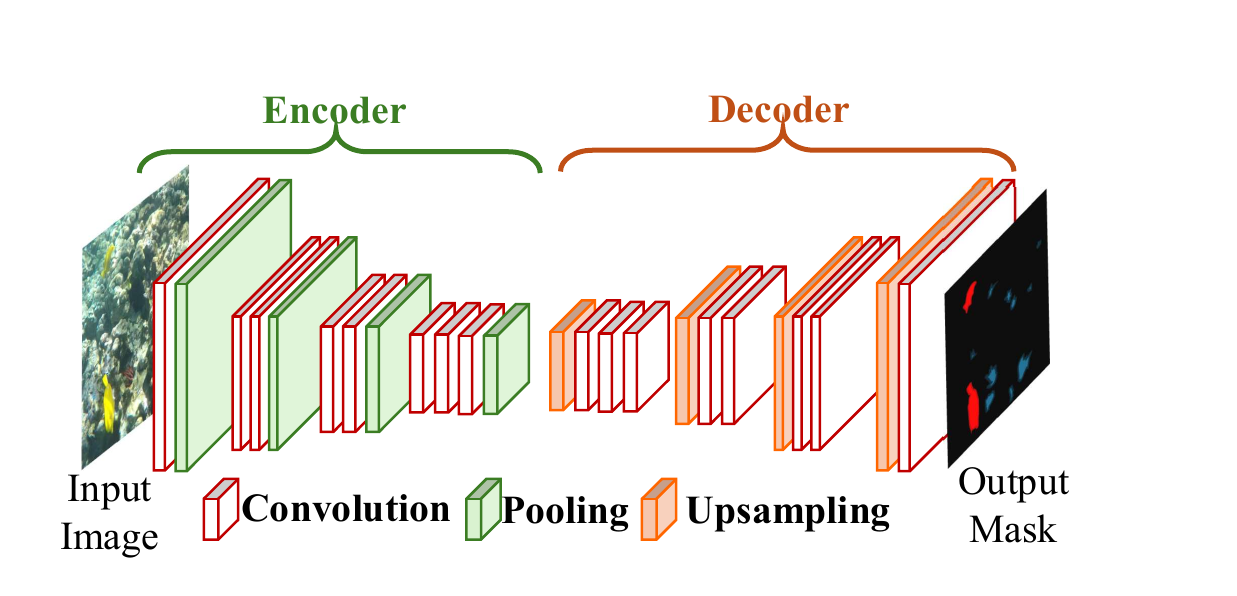} \\
    b) Classification & c) Segmentation \\
\end{tabular}}
\caption[Convolutional Neural Network Architecture]{Diagrams depicting: a) convolution\footnotemark~and pooling operations -- the feature maps decrease in spatial size while increasing in depth, and the number of layers and parameters used for convolution determine the effective receptive field of features; b) a CNN classifier -- the spatial dimensions are reduced and the depth is increased to obtain a deep embedding, and the final fully connected layer predicts the class of the input image; and c) an CNN segmentation architecture -- the encoder-decoder combines the feature extraction capabilities of convolutional layers with the spatial recovery capabilities of the upsampling layers, and predicts the class of each pixel at the final layer to obtain the segmentation mask.}
\label{fig:cnn}
\end{figure}

\footnotetext{Convolution diagram inset reused from~\cite{song2023use} under \href{https://creativecommons.org/licenses/by/4.0/}{CC BY 4.0}.  Example input image taken by the first author.}

CNNs use convolutional layers to repeatedly perform the convolution operation (a specialized type of linear operation) and extract features from data at different levels (Fig.~\ref{fig:cnn})~\citep{goodfellow2016deep}. This survey focuses on image data, and therefore the CNN would be used to extract hierarchical representations by sliding a kernel of learnable parameters across the input image~\citep{goodfellow2016deep}.  Through this process, the CNN can capture patterns and spatial features such as edges, textures, shapes and structures. As the data passes through the network, the convolutional layers will combine the simple features from earlier layers such that more complex and abstract concepts will be extracted by the deeper layers~\citep{albawi2017understanding}.  Features extracted by CNNs are local, \ie only a certain spatial region around the pixel in question will impact the computation of the output feature embedding of that pixel. This is known as the \textit{receptive field size}, and is determined by the depth of the CNN and the other parameters (kernel size, stride, padding) of the convolutional layers. The hierarchical, spatial feature extraction of CNNs make them powerful for tasks such as image classification, object detection and segmentation. 

\begin{figure}[t]
\centering
\centerline{\includegraphics[width=140mm, clip, trim=0.5cm 1.6cm 0.5cm 1cm]{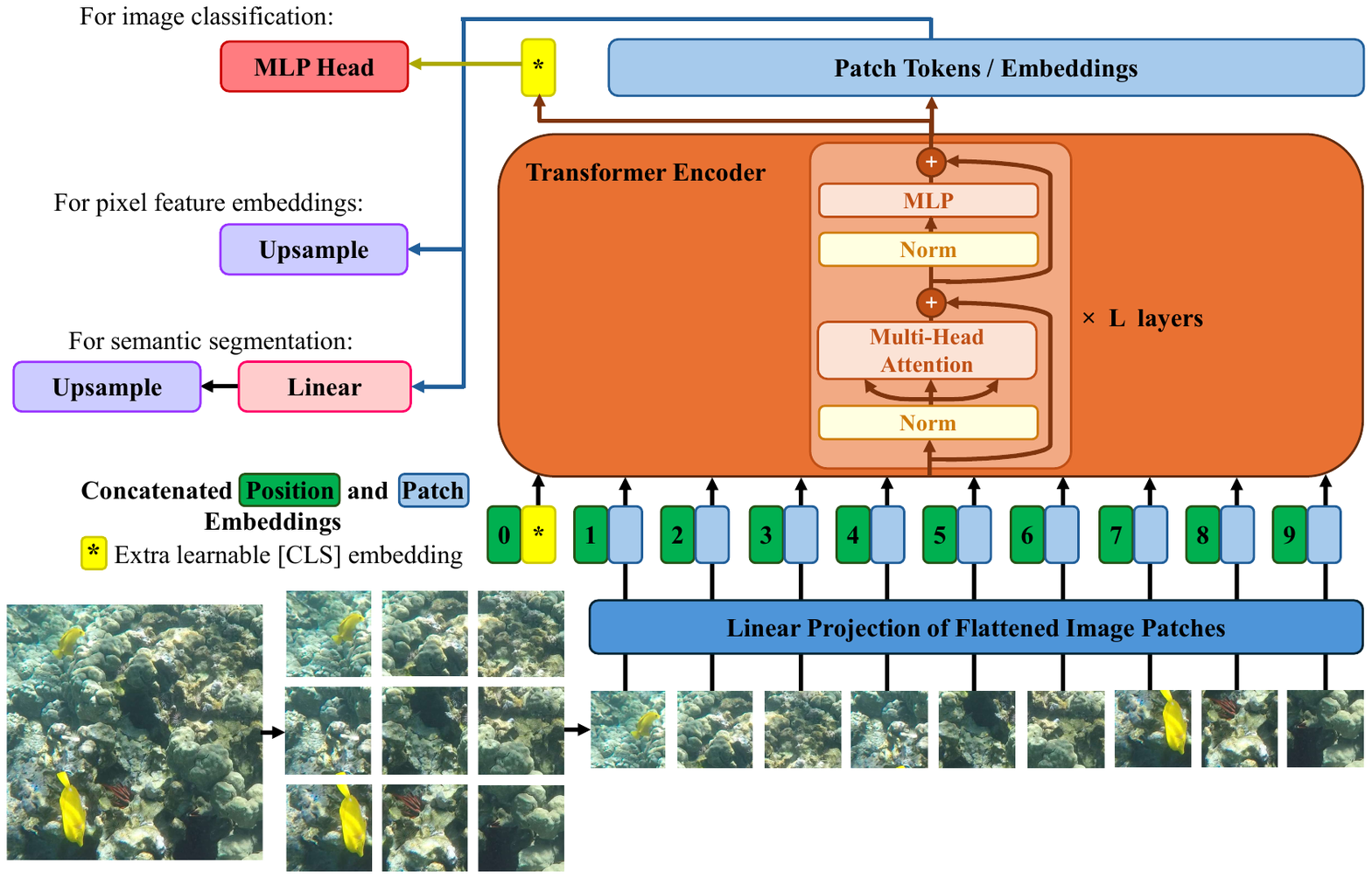}} 
\caption[Vision Transformer Architecture]{A diagram for the architecture of a Vision Transformer (ViT)~\citep{dosovitskiy2020image}.  ViTs divide the input image\footnotemark~into patches of equal size.  The position of the patches in the sequence is concatenated with the linear embedding of the patch before being input into the transformer.  There is also a learnable embedding (called the `CLS' embedding) which encodes global features for the whole image and is used for image classification.  The output patch tokens can be spatially upsampled to the original input size to obtain per-pixel feature embeddings. Figure was adapted and redrawn from~\cite{dosovitskiy2020image}.}
\label{fig:transformer}
\end{figure}

\footnotetext{Example input image taken by the first author.}

CNNs can be used to perform a range of computer vision tasks.  Fig.~\ref{fig:cnn} depicts common CNN architectures for image classification and semantic segmentation.  The classification architecture reduces the spatial dimensions while increasing the depth of the feature maps such that a single deep embedding is obtained for the input image.  The class of the image is then predicted as one of the pre-defined classes.  For semantic segmentation, the encoder of the architecture reduces the spatial dimensions and learns a deep feature space and then a decoder with upsampling layers increases the spatial dimensions back up to the original input size.  The class of every pixel in the image can then be predicted to obtain the output segmentation mask for the image~\citep{badrinarayanan2017segnet}.

For many years, CNNs were the gold standard for extracting meaningful feature embeddings for image data, however recently there have been advances in the use of transformers. There are settings, particularly for applied computer vision problems, for which CNN are preferred for their faster inference time and low memory constraints.  For example, \cite{raine2024image} demonstrates that training a CNN with supervision from the CLIP\footnote{Contrastive Language-Image Pretraining (CLIP)~\citep{radford2021learning}}~\citep{radford2021learning} large vision-language model (with a transformer architecture) performs inference 30 times faster than using the CLIP model directly.  The transformer architecture is described in the next section. 

\subsection{Transformers}
\label{subsec:lit-transformers}

Transformers were first introduced in the `Attention is All You Need' paper~\citep{vaswani2017attention}.  A transformer is comprised of an encoder and decoder of multi-head self-attention and feed-forward neural network layers~\citep{caron2021emerging}.  Self-attention is a mechanism which relates positions of a sequence and enables the decoder to dynamically `attend' to different parts of the input~\citep{khan2022transformers}. Transformers have led to significant advances in a range of tasks in Natural Language Processing (NLP) including text generation, translation and sentiment analysis~\citep{khan2022transformers}. 

Following the success of transformers for NLP tasks, \cite{dosovitskiy2020image} proposed the ViT and demonstrated that transformers could be applied directly on sequences of image patches, as seen in Fig.~\ref{fig:transformer}. These models perform well on computer vision tasks when combined with self-supervised pre-training on large quantities of images.  The resulting foundation models based on transformers outperform CNNs for many tasks and are able to model long range dependencies between image features more effectively than CNNs.

Recent foundation models based on transformers include iBOT\footnote{Image BERT Pre-Training with Online Tokenizer (iBOT)~\citep{zhou2021ibot}}~\citep{zhou2021ibot}, DINO\footnote{Self Distillation with NO Labels (DINO)~\citep{caron2021emerging}}~\citep{caron2021emerging}, DINOv2~\citep{oquab2023dinov2} and I-JEPA\footnote{Image-based Joint-Embedding Predictive Architecture (I-JEPA)~\citep{assran2023self}}~\citep{assran2023self}. Furthermore, large vision-language models, such as CLIP~\citep{radford2021learning} and ALIGN\footnote{A Large-scale ImaGe and Noisy-Text Embedding (ALIGN)~\citep{jia2021scaling}}~\citep{jia2021scaling}, enable both text and image data to be represented together in the feature space.  Section~\ref{sec:weakly-underwater} describes approaches which leverage large foundation and vision-language models for underwater scene understanding. 

Sections~\ref{sec:lit-back-CV} and \ref{sec:lit-types-dl} described the relevant computer vision tasks and the deep learning models that can be used to perform these tasks.  In the following two section, we describe approaches in the literature which perform automated image analysis specifically for each of the key ecosystems considered in this survey: seagrass meadows (Section~\ref{sec:lit-seagrass}) and coral reefs (Section~\ref{sec:lit-coral}).

\section{Automated Image Analysis for Seagrass Meadows}
\label{sec:lit-seagrass}

This section reviews works on automated image analysis for seagrass meadows. Related works on mapping seagrass meadows use UAVs~\citep{jeon2021semantic, tahara2022species, roman2021using, tallam2023application}, remotely piloted aircraft~\citep{hobley2021semi} and satellite imagery~\citep{phinn2008mapping, ferguson1997remote}. Recent approaches have also combined multiple types of data to improve the accuracy and resolution of seagrass maps~\citep{mckenzie2022improving}.  However, remote methods do not readily allow discrimination between different species of seagrass~\citep{pham2019review} and are often limited by cloud cover and weather conditions~\citep{tahara2022species, roman2021using}. UAV data is similarly limited by the turbidity of the water and environmental conditions, including the angle of the sun, wind and cloud cover~\citep{rende2020ultra}. It is also difficult to accurately discriminate between species of seagrass in UAV imagery~\citep{abid2024seagrass}, and models often exhibit a high false positive rate for macroalgae~\citep{tallam2023application}. It is necessary to quantify the species and extent of seagrass present for accurate estimation of blue carbon stocks~\citep{lavery2013variability}, thus requiring collection of field data.  Furthermore, field data is often needed to validate broad-scale maps generated from aerial or satellite data~\citep{rende2020ultra, mckenzie2022improving}. 

This section describes prior approaches for analysis of seagrass imagery collected in the field by divers, snorkellers, robotic platforms or remote cameras (Fig.~\ref{fig:data-types}). Many early approaches are based on classical machine learning (Section~\ref{subsec:lit-sea-ml}), while more recent approaches leverage deep learning, as described in Section~\ref{subsec:lit-seagrass-binary} (binary approaches) and Section~\ref{subsec:lit-seagrass-multi} (multi-species approaches). Although the primary focus of this section is approaches for seagrass imagery, some relevant works for monitoring kelp and macroalgae are also included because this problem shares some characteristics with seagrass analysis.

\subsection{Classical Machine Learning}
\label{subsec:lit-sea-ml}

Early approaches focused on hand-crafted features, \eg Local Binary Patterns (LBP)~\citep{bewley2012automated}, Gabor filtering~\citep{bonin-font2017visual, burgeura2016towards, bewley2012automated}, Haralick features~\citep{denuelle2010kelp} or gray-level co-occurrence matrices~\citep{massot-campos}, with classifiers such as Support Vector Machine (SVM)~\citep{gonzalez2017machine, bonin2016towards}. \cite{bewley2012automated} performed patch classification using LBP and an SVM, and \cite{denuelle2010kelp} performed segmentation using Mahalanobis distance and Haralick features, both for kelp presence/absence analysis in AUV imagery.  \cite{sengupta2020seagrassdetect} classified images based on the number of edges detected, assuming that images with seagrass would contain more edges than those without.  This approach relies on the assumptions that the morphology of the seagrass will result in many edges, and that when seagrass is not present, the seafloor will be free from any other objects. 

\cite{burguera2020segmentation} performed binary seagrass segmentation of imagery collected by a downward-facing camera on an AUV by classifying image patches, and compared machine learning, simple and deep neural networks~\citep{burguera2020segmentation}. The work demonstrated that for binary presence/absence segmentation of the \textit{Posidonia oceanica} seagrass, which is dark in colour and has a distinctive texture, a simple two-layer neural network is sufficient and achieves comparable performance to a CNN.

All other approaches used the hand-crafted features to perform binary patch classification for the presence/absence of a single seagrass species of interest~\citep{massot-campos, burgeura2016towards, bonin-font2017visual, bonin2016towards, sengupta2020seagrassdetect}.  The following section describes approaches which leverage deep learning for binary segmentation of seagrass imagery. 

\subsection{Deep Learning for Binary Segmentation of Seagrass Imagery}
\label{subsec:lit-seagrass-binary}

The majority of prior approaches for automated analysis of seagrass imagery with deep learning consider only the binary presence/absence problem~\citep{reus2018looking, weidmann2019closer}, or aim to detect only one species of seagrass~\citep{massot-campos, martin-abadal2018deep, pamungkas2021segmentation, marre2020fine, massot2023assessing}. These approaches are detailed in this section, followed by a discussion of the multi-species approaches in the next section.

Several seagrass segmentation approaches~\citep{reus2018looking, weidmann2019closer, wang2020real} have been evaluated on `Looking For Seagrass' (Appendix~\ref{appendix:A}, Table~\ref{table:datasets-seagrass}), a dataset of AUV images labelled with pixel-wise ground truth masks for the presence/absence of seagrass~\citep{reus2018looking}.  First, \cite{reus2018looking} contributed the dataset and proposed their approach in which image patches were extracted for the two classes based on the ground truth masks, and then LBP, Histogram of Gradients (HOG) and CNN features were compared for training a logistic regression classifier.  This work found that the CNN features, extracted from an InceptionNetV3~\citep{szegedy2016rethinking} pre-trained on ImageNet~\citep{deng2009imagenet}, yielded the best performance~\citep{reus2018looking}. Follow up works by~\cite{weidmann2019closer} and \cite{wang2020real}, trained semantic segmentation architectures end-to-end on the dense ground truth masks.  

\cite{ruscio2023autonomous} approach the problem of mapping seagrass meadows as a boundary inspection task and leverage binary image segmentation, boundary tracking and a loop closure detector based on visual and navigation information.  The segmentation model was trained using dense ground truth masks outlining the presence/absence of seagrass.

Acoustic sensors on-board underwater robotic vehicles have been successfully used to detect seagrass~\citep{ferretti2017towards, rende2020ultra}. \cite{rende2020ultra} combined UAV photo-mosaic maps, satellite multispectral imagery, acoustic data and underwater photogrammetry data for high resolution seafloor and seagrass mapping. These approaches can be operated when environmental conditions are poor (\eg if water turbidity makes visibility impossible for image-based levels), however they only consider one species of seagrass. 

There are also a number of works which perform object detection of individual leaves of the seagrass \textit{Halophila ovalis}, but these do not consider the multi-species seagrass problem and were trained using bounding boxes as supervision~\citep{moniruzzaman2019faster, noman2021seagrass, noman2023improving}.

This section has described deep learning approaches for binary segmentation of seagrass imagery, and the following section focuses on the approaches which perform multi-species analysis. Weakly and unsupervised approaches for seagrass image analysis are discussed in Section~\ref{subsubsec:lit-sea-weak}.

\subsection{Deep Learning for Multi-species Segmentation of Seagrass Imagery}
\label{subsec:lit-seagrass-multi}

This section discusses approaches for multi-species segmentation of seagrass imagery. Although there has been research on multi-class patch classification of macroalgae~\citep{mahmood2020automatic}, the first work on multi-species identification of seagrass with deep learning is \cite{raine2020multi}.  Since publication of this approach and contribution of the DeepSeagrass dataset (further detail in Section~\ref{subsec:lit-underwater-datasets}), there have been further works on the multi-species seagrass problem~\citep{noman2021multi, mehrubeoglu2021segmentation, balado2021semantic, ozaeta2023seagrass, paul2023lwds}, as outlined in the following paragraphs.

\cite{noman2021multi} proposed a semi-supervised approach for multi-species patch classification of seagrass images. They first trained a teacher model on the DeepSeagrass dataset and then used this model to predict class labels on a new dataset they collected\footnote{The authors are not able to share this dataset, and it was therefore not included in Table~\ref{table:datasets-seagrass}.} before retraining the model on both the labelled and pseudo-labelled data~\citep{noman2021multi}. 

\cite{mehrubeoglu2021segmentation} performed multi-class segmentation of seagrass blades, where individual seagrass blades were sampled and images taken of a single seagrass blade at a time, and then a model was trained with full supervision to segment regions of algae,  tubeworm, epiphytes, grass blade, and background.

\cite{ozaeta2023seagrass} recently proposed an approach for multi-species seagrass classification of the DeepSeagrass dataset based on differentiable architecture search, and which could be trained quickly (approximately four hours) on a commercially available Apple MacBook device. \cite{paul2023lwds} also proposed a lightweight patch classification architecture based on a Visual Geometry Group (VGG) architecture and evaluated their performance on DeepSeagrass. 

Finally, \cite{balado2021semantic} trained a semantic segmentation model end-to-end to differentiate between five species of macroalgae in photo-quadrat imagery, however this was trained on domain expert generated polygons.

All of these approaches are trained using either patch-level, polygon or dense pixel-wise masks.  Section~\ref{subsubsec:lit-sea-weak} describes approaches which further reduce the training signal.

 \section{Automated Image Analysis for Coral Reefs}
 \label{sec:lit-coral}

In this section, we consider the second ecosystem of interest in this survey: coral reefs. This section outlines the prior work in applying classical machine learning (Section~\ref{subsec:lit-coral-ml}) and deep learning (Section~\ref{subsec:lit-coral-dl}) to automated processing of coral imagery.

\subsection{Classical Machine Learning}
\label{subsec:lit-coral-ml}

As for early seagrass analysis approaches, the first approaches for analysing coral images were based on patch or image classification, using hand-crafted features and machine learning classifiers~\citep{xu2019deep}.  \cite{beijbom2012automated, beijbom2015towards} presented a method based on texture and colour descriptors over multiple scales, by using the Maximum Response filter bank and an SVM. A number of works employ variants of LBP to perform feature extraction for classification of underwater images~\citep{shihavuddin2013image, shakoor2018novel, mary2018classification, sotoodeh2019structural, dayoub2015robotic, bewley2015hierarchical}. \cite{bewley2015hierarchical} performed multi-species classification of imagery collected by an ROV using features via LBP and a hierarchical network of probabilistic graphical models.  \cite{paul2020gradient} proposed an Aura Matrix based on gradient based Cumulative Relative Difference as a feature extractor for classification of coral textures.  

More recently, \cite{ganesan2022novel} proposed a novel feature extractor for coral image classification based on local directional derivative binary pattern and systematic local discriminate binary pattern, achieving the state-of-the-art for the EILAT2~\citep{loya2004coral, shihavuddin2017dataset} dataset.  \cite{mohamed2022automatic} compared unsupervised segmentation algorithms K-Means, Otsu clustering, Fast and Robust Fuzzy C-Means, and Superpixel-based Fast Fuzzy C-Means to segment images into broad classes `corals', `blue corals', `brown algae', `other algae', `seagrass' and `sediments'.  

\subsection{Deep Learning} 
\label{subsec:lit-coral-dl}

Recent efforts focus on using deep learning models to extract features and perform classification, object detection, and semantic segmentation of coral images. In this section, we provide a brief overview of classification approaches as these works laid the groundwork for automated image analysis. This survey does not cover in detail approaches for object detection in underwater imagery \eg works including~\cite{han2020marine, li2015fast, modasshir2018coral, bijjahalli2023semi, modasshir2020augmenting, haixin2023marinedet, liu2021dataset}, but instead focuses on segmentation to enable species coverage estimation. Section~\ref{subsubsec:lit-coral-weak} describes approaches for weakly supervised segmentation of coral imagery. 

\subsubsection{Classification}
\label{subsubsec:lit-coral-class}

There are many prior works which perform multi-species classification of image patches with deep learning~\citep{mahmood2016coral, mahmood2016automatic, raphael2020deep, modasshir2018mdnet, gomez2019coral, gomez2019towards}. 

Early deep learning approaches combined deep and hand-crafted features for classification of corals. For example, \cite{mahmood2016coral} combined extracted CNN features with texture and colour-based hand-crafted features to classify coral images.  Their method was trained on square patches of varying sizes with a point label at the centre. Similarly, \cite{cao2015marine} combined CNN deep features with hand-crafted features when performing multi-class classification of marine animals and fish.

Later approaches focused solely on deep features and neural network architectures.  \cite{modasshir2018mdnet} presented their network, MD-Net, which extracted patches at multiple scales and aggregated the features before performing classification.  \cite{gomez2019towards} investigated different network architectures and analysed the impacts of data augmentation and transfer learning on texture image classification, where all approaches were trained in a fully supervised fashion.  \cite{gomez2019coral} have also presented a supervised two-stage classifier based on neural networks to classify corals using texture and structure. The first level determined if the presented image is a structure or texture image and then employed one of two trained networks to classify the image in the second stage. 

Other recent approaches for coral image classification based on deep learning include \cite{mahmood2020resfeats}, in which features extracted from multiple layers of a residual network pre-trained on ImageNet were used for coral image classification.  \cite{wyatt2022using} studied the impact of deep ensembles for underwater image classification, and found that the performance of classifiers degrades under significant data shift, but that using an ensemble of classifiers improves the robustness. Most recently, \cite{jackett2023benthic} performed image patch classification for six benthic classes by training a model on image patches extracted from random point labels.

\cite{szymak2020effectiveness} performed real-time underwater image classification on-board a robotic underwater vehicle of coarse-grained classes including divers, fish and underwater vehicles. \cite{mohamed2020semiautomated} combined patch classification of underwater images with satellite imagery to perform multi-class benthic and seagrass mapping. This approach was trained on a dataset of manually annotated patches. 

There are also numerous works which successfully fine-tune generic pre-trained CNNs for coral image or patch classification~\citep{gomez2019towards, lumini2020deep, gonzalez2020monitoring}.  Similarly, the deep learning engine behind the CoralNet data annotation platform is an EfficientNet-B0 backbone and multi-layer perceptron fine-tuned on patches contributed to the platform~\citep{chen2021new}.

The approaches mentioned in this section laid important groundwork for automated analysis of coral images, however they are limited in terms of resolution and granularity of model outputs. The following sections review approaches which perform semantic segmentation of underwater imagery.

\subsubsection{Binary Segmentation}
\label{subsubsec:lit-coral-binary}

Binary segmentation of corals has been performed in the literature, however these approaches are trained with supervision from dense pixel-wise ground truth or polygon masks~\citep{alonso2017coral, song2021development, arain2019improving, king2019deep, ziqiang2023coralvos}. 

\cite{arain2019improving} combined sparse stereo point clouds with monocular image segmentation for underwater obstacle avoidance. \cite{king2019deep} utilised multi-view stereoscopic images as input image pairs to improve accuracy of semantic segmentation of coral reef images.  The disparity map between the left and right images was added as a fourth channel to each coral reef image and their architecture was inspired by a Siamese network to generate the disparity map and relevant spatial features.  Both of these approaches required supervision from dense segmentation masks. 

Recently, \cite{ziqiang2023coralvos} contributed a large labelled dataset for dense segmentation of underwater video sequences.  They evaluated and compared six recent Video Object Segmentation approaches on this dataset, however they only considered binary segmentation.

\subsubsection{Multi-class Segmentation}
\label{subsubsec:lit-coral-multi}

Compared to patch classification and binary segmentation, there are fewer approaches in the literature for multi-class segmentation of coral imagery.  This section details approaches which are trained in a fully supervised manner.

Of these approaches, a number only consider very high-level coarse-grained classes instead of species. For example, \cite{islam2020semantic} developed a segmentation method and contributed a custom dataset with human-generated segmentation masks, where the segmentation classes were coarse-grained \eg `background', `robot', `plant', `human'. \cite{fu2023masnet} performed segmentation of marine animals using a fusion-based Siamese deep neural network and also contributed a dataset of images, where the classes are `sea products', `big fish', `small fish', `turtle' and `other'. \cite{li2021marine} performed marine animal segmentation, where object classes were coarse grained and only the animal targets were segmented from the rest of the image.

\cite{giles2023combining} performed multi-class segmentation of UAV RGB imagery for the classes `bleached coral', `healthy coral', `background' and `sun glint', but they were unable to accurately segment `seagrass', `algae-covered coral', and `newly dead coral'.  This suggests that aerial imagery can be used for segmentation of coarse-grained classes, but does not provide sufficient clarity or resolution for discriminating between fine-grained classes. 

Many approaches for multi-species segmentation are trained from dense pixel-wise masks or polygon annotations~\citep{picek2020coral, pavoni2019semantic, zhang2022deep, zhang2024cnet, fu2023masnet, islam2020semantic, vsiaulys2024coverage, zhong2023combining, sauder2023scalable}.  For example, \cite{pavoni2019semantic} performed semantic segmentation of ortho-mosaic benthic images using a Bayesian CNN trained on human-generated polygons. \cite{zhang2022deep} integrated underwater photogrammetry with deep learning and proposed a boundary-oriented U-Net for multi-species coral segmentation of ortho-mosaics, trained from dense ground truth masks. More recently, \cite{zhang2024cnet} proposed a novel architecture called CNet, with two branches: a pre-trained ResNet50 encoder for RGB information and a VGG encoder for depth information.  The network is trained using dense ground truth masks and they are able to perform segmentation of imagery into three classes: background, \textit{Pocillopora} and \textit{Acropora}.

To support training approaches on dense ground truth masks, TagLab~\citep{pavoni2022taglab} was proposed to enable AI-assisted dense annotation of ortho-mosaic images for semantic segmentation. TagLab is a CNN-based interactive annotation tool which speeds up manual annotation (by 42\% for experts and up to 90\% for non-expert annotators). The dense labelled masks can be used to train a model to perform semantic segmentation. While this approach speeds up dense annotation of ortho-mosaic images, it does not address the problem of leveraging the existing, large quantities of sparse point-labelled imagery.

Some recent approaches for multi-species segmentation of coral imagery have incorporated photogrammetry~\citep{zhang2022deep, zhong2023combining, remmers2023close, runyan2022automated, terayama2022cost, yuval2021repeatable, sauder2023scalable}.  Photogrammetry is a technique used to extract measurements from imagery using principles of optics, and enables conversion of 2D images into 3D maps~\citep{remmers2023close}.  From these 3D maps, scientists can extract reef health indicators including 3D rugosity, which assesses structural reef complexity, and volume of corals~\citep{zhang2022deep}.  Use of photogrammetry is reviewed in detail by \cite{remmers2023close} and is therefore considered out of scope for this survey. 

\section{Weakly and Self-Supervised Deep Learning in Other Domains}
\label{sec:lit-weakly}

Reducing label dependency is the key driver of this review. This section therefore reviews relevant prior approaches which perform weakly supervised and self-supervised deep learning developed outside of the underwater domain, and evaluates their suitability for this domain.  We first review weakly supervised approaches which use image-level labels (Section~\ref{subsec:lit-weak-image}), followed by approaches which use point labels (Section~\ref{subsec:lit-weak-point}), and self-supervised approaches (Section~\ref{subsec:lit-self-sup}).

\subsection{Image-level Labels}
\label{subsec:lit-weak-image}

Weakly supervised segmentation from image-level labels assumes that an image has been assigned one or multiple labels which describe the objects present in the image~\citep{zhu2023survey}.

Some prior approaches performed segmentation from image-level labels~\citep{kwak2017weakly, bae2020rethinking}. Many of these approaches use object saliency maps and Class Activation Maps (CAMs) to highlight regions of an image which are important for segmentation of a particular class. \cite{kwak2017weakly} described an approach for weakly supervised segmentation from image-level labels, in which they determined the superpixel boundaries for the image and fed this into their network along with the input image. When the CAM was combined with superpixels, the network was better able to segment the boundaries of objects in the image. \cite{rong2023boundary} performed boundary-enhanced co-training to perform segmentation from image-level labels, where the boundary enhancement technique was designed to improve pseudo-labelling from a CAM. 

Most recently, \cite{yang2024foundation} combined the CLIP large vision-language model~\citep{radford2021learning} and the Segment Anything~\citep{kirillov2023segment} model to perform foundation model-assisted segmentation. The Segment Anything model was used in this architecture to generate pseudo-labels of foreground objects for training a semantic segmentation model.

The approaches described in this section are designed to perform segmentation of foreground objects from the background, and do not consider tasks where the boundaries are hard to visually discern and the instances are overlapping and cluttered.  Even the approaches which perform \textit{panoptic} segmentation, \ie segmentation of all `things' and `stuff' in an image, are demonstrated on datasets of common objects (\eg Pascal VOC\footnote{Visual Object Classes (VOC)~\citep{everingham2010pascal}}~\citep{everingham2010pascal} and MS COCO\footnote{Microsoft Common Objects in COntext (COCO)~\citep{lin2014microsoft}}~\citep{lin2014microsoft}), which have different image characteristics as compared to underwater imagery~\citep{shen2021toward}. An example image from MS COCO can be seen in Fig.~\ref{fig:regular-seg-examples}.  

\subsection{Point Labels}
\label{subsec:lit-weak-point}

Bridging the gap between whole image labels and pixel-wise segmentation has attracted a large amount of research effort; however, the gap between point labels and segmentation has comparably fewer approaches~\citep{zhu2023survey}.

Early approaches for weakly supervised segmentation from point labels focused on `objectness' and CAMs.  \cite{bearman2016s} used a single point per object to train a segmentation architecture based on VGG-16. They incorporated an objectness prior and found an improvement in the boundaries of their object segmentation. Point Supervised Class Activation Maps was proposed in~\cite{mcever2020pcams} and also assumes that there is one point label for each object of interest in the image.  

For remote sensing imagery, \cite{wang2020weakly} used a U-Net architecture to segment satellite images of cropland.   Their method was trained using one labelled randomly placed point in each image and they determined that the architecture was able to achieve greater than 85\% accuracy when more than 100 training examples were provided.  \cite{hua2021semantic} compared point, line and polygon annotations for segmentation of urban remote sensing imagery, and found that encoding neighbourhood relations in both the spatial and feature space improves performance.  This assumed that nearby pixels belong to same class -- in underwater imagery, this is often not the case as species have fractal-like boundaries, overlapping instances, and the same species can appear in multiple areas of the image (Section~\ref{subsec:lit-image-characteristics}).

\renewcommand{\topfraction}{.75}
\begin{figure}[t]
\centering
\setlength{\tabcolsep}{2pt}
\centerline{\begin{tabular}{cc}
    \includegraphics[height=32mm]{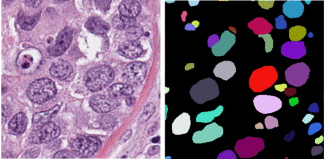} &
    \includegraphics[height=32mm]{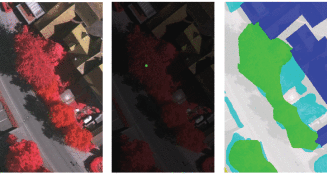} \\
    a) Nuclei Segmentation~\citep{qu2020nuclei} & b) Remote Sensing Imagery~\citep{hua2021semantic} \\
    \\
    \multicolumn{2}{c}{\includegraphics[height=28mm]{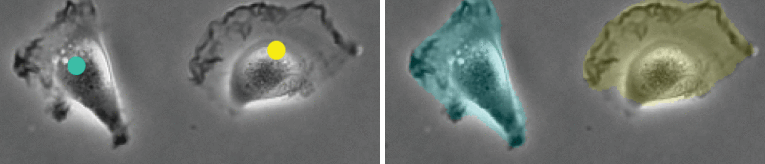}} \\
    \multicolumn{2}{c}{c) Cell Segmentation~\citep{zhao2020weakly}} \\
    \\
    \multicolumn{2}{c}{\includegraphics[height=30mm]{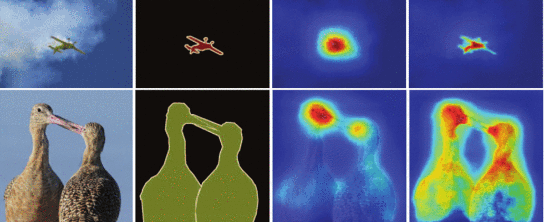}} \\
    \multicolumn{2}{c}{d) Segmentation of Common Objects~\citep{wang2020self}} \\
\end{tabular}}
\caption[Examples of Weakly Supervised Segmentation in Other Domains]{Examples of data\footnotemark~used for weakly supervised segmentation approaches in: a) and c) medical images, b) remote sensing imagery, and d) images of common objects.}
\label{fig:weak-seg-other}
\end{figure}

Various works have performed point-based, weakly labelled nucleus segmentation in histopathological images~\citep{obikane2019weakly, tian2020weakly, qu2020nuclei}; cell segmentation~\citep{zhao2020weakly}; and organ segmentation in medical imagery~\citep{roth2019weakly}. These tasks involve annotators clicking directly on targets of interest, which typically have a discernible shape or visual boundary. This setting differs from underwater imagery, in which existing quantities of data labelled with CPC have the points either randomly distributed or spaced as an equal grid over the image.

More recently, `Point2Mask'~\citep{li2023point2mask} proposed a unified framework, where one network performed mask pseudo-labelling from sparse points and the other network is trained with full supervision on the pseudo-labels. \cite{fan2022pointly} generated pseudo-masks from point labels by minimizing the pixel-to-point traversing costs, based on semantic similarity, low-level boundaries and instance-aware manifold knowledge. They also compared centre-biased vs border-biased point labelling, finding that placing point labels in the centre of instances improves performance. These approaches were trained from one point label, centrally located in each target instance~\citep{li2023point2mask, fan2022pointly}, opposed to the sparse random or grid labels encountered in underwater datasets.  \cite{cheng2022pointly} proposed a simple framework for weakly supervised segmentation, however they trained their method on both bounding boxes and also sparse points within each bounding box.

In all the cases presented here, \ie remote sensing imagery, urban or indoor scenes, and medical imagery, the target classes have clear boundaries and some semantic distinction between the target and the `background' (Fig.~\ref{fig:weak-seg-other}). The challenges faced for underwater imagery (turbid and blurry images, poorly defined boundaries, lack of foreground/background separation, morphologically indistinct forms etc.) mean that these approaches may not effectively transfer to the underwater domain (this is discussed further in Section~\ref{sec:discussion}).

\subsection{Self-Supervised Deep Learning in Other Domains}
\label{subsec:lit-self-sup}

\footnotetext{a) Reprinted with permission from~\cite{qu2020nuclei}, © 2020 IEEE. b) Reprinted with permission from~\cite{hua2021semantic}, © 2021 IEEE. c) Reprinted with permission from~\cite{zhao2020weakly}, © 2020 IEEE. d) Reprinted with permission from~\cite{wang2020self}, © 2020 IEEE.}

Self-supervised deep learning refers to approaches which are trained using the inherent structure or properties of the data as opposed to annotations.  Self-supervised learning can also involve some objective or pre-text task used to train the model.  The SimCLR~\citep{chen2020simple} framework uses contrastive self-supervision to learn a discriminative feature space, by taking an input image and performing a series of data augmentations or transforms to obtain positive samples.  These positive samples are combined with many randomly selected images from the rest of the dataset which act as the negative samples.  The model is then trained to maximise the cosine similarity between the deep representation of the input image and its positive samples, and minimise the similarity to the negative samples.  

An important self-supervised learning framework is the teacher-student model, in which outputs from a teacher model are used to train a student model~\citep{shin2020semi}. The student-teacher model is a form of knowledge distillation~\citep{hinton2015distilling}, in which the learnt knowledge of a larger teacher model is transferred to a smaller student model~\citep{gou2021knowledge}.  There is considerable research into the design of the student and teacher model, and the loss function used~\citep{gou2021knowledge}.

Recently, there have been advances in self-supervised approaches for transformers \citep{feng2023evolved,caron2021emerging,oquab2023dinov2}.  To train the DINO~\citep{caron2021emerging} model with self-supervision, the input image was transformed into a set of positive samples, and then a student model is trained to predict the output of the teacher model. Both `local' and `global', \ie spatially small and large crops of the input image, are created, with all views passed to the student, and only global views passed to the teacher, to encourage the model to relate the local crops to the global context from the teacher~\citep{caron2021emerging}.  Another recent self-supervised transformer, iBOT~\citep{zhou2021ibot} uses a tokenizer and target network structure, in which the tokenizer takes the input image and masks some patches before passing to the target network, which tries to recover the masked image patches. The DINOv2~\citep{oquab2023dinov2} foundation model was trained on a combination of both the iBOT~\citep{zhou2021ibot} and DINO~\citep{caron2021emerging} self-supervised loss functions. \cite{caron2024location} recently proposed a location-aware self-supervised approach, in which the position of patches in an input image is predicted.  

These approaches have yielded impressive results for mainstream semantic segmentation and training large foundation models for task-agnostic feature representations. Section~\ref{sec:weakly-underwater} describes the use of these methods for the underwater use case. 

\section{Weakly Supervised Underwater Image Analysis}
\label{sec:weakly-underwater}

This sections brings together the themes of marine environmental monitoring, computer vision, deep learning, and weakly supervised learning, and summarises the approaches at this intersection. Section~\ref{subsubsec:lit-sea-weak} describes weakly supervised approaches for analysis of seagrass images and Section~\ref{subsubsec:lit-coral-weak} describes approaches for coral reef imagery.

\subsection{Weakly Supervised Segmentation for Seagrass Images}
\label{subsubsec:lit-sea-weak}

Many approaches for seagrass segmentation focused on binary presence/absence segmentation and did not consider the multi-species problem. Early approaches \citep{raine2020multi} demonstrated that deep learning can be used to discriminate between species of seagrass. This work inspired further research interest in multi-species coarse seagrass segmentation, however these approaches (Section~\ref{subsec:lit-seagrass-multi}) still relied on availability of patch-level labels for training \citep{noman2021multi}.  \cite{raine2024image} extended this work and demonstrated that image-level labels instead of patch-level labels at training time can be leveraged while yielding multi-species coarse seagrass segmentation at inference time.  This work also investigated the utility of using large, general, vision-language models for domain-specific tasks and demonstrated that combining a supervisory signal from the CLIP vision-language model with weak domain expert labels was effective to train a coarse seagrass segmentation model. 

Most recently, \cite{abid2024seagrass} performed binary patch classification of the DeepSeagrass dataset, described in Section~\ref{subsec:lit-underwater-datasets}, using an unsupervised curriculum learning framework.  Their unsupervised approach outperformed all prior works for binary classification of seagrass, however they left the multi-species task for future work. 

Unsupervised classification of fine-grained categories including species of seagrass and classification of different substrates remains an important and open research problem.  The semantic similarity of these fine-grained classes makes unsupervised learning with contrastive or clustering-based approaches challenging.  It is likely that advances in powerful feature extractors will play a key role in accurately discriminating between fine-grained seagrass species in a weakly supervised or unsupervised fashion.

\subsection{Weakly Supervised Segmentation for Coral Images}
\label{subsubsec:lit-coral-weak}

The application of alternative deep learning approaches, such as self-supervised, weakly supervised, semi-supervised or unsupervised learning, to coral imagery has not been thoroughly investigated.  \cite{shields2020towards} presented an approach for adaptive benthic habitat mapping, which used a convolutional auto-encoder to learn a compressed feature representation of different habitats, including reef, sand and macroalgae. Their approach clustered the image features in the latent space and each cluster was assigned a habitat label by a biologist.  \cite{rao2017multimodal} performed unsupervised clustering of co-located AUV images (human-labelled at the image-level into five coarse classes) and remotely sensed bathymetric data.  These approaches did not perform pixel-wise segmentation of imagery.  

\cite{yu2019weakly} proposed a method in which a CNN trained on sparsely labelled coral images is used to generate features for unlabelled patches, and then the patch features are clustered with Latent Dirichlet Allocation and patches with higher probability features are selected as pseudo-labels for re-training the CNN. More recently, \cite{singh2023attention} performed self-supervised instance segmentation of underwater imagery, based on a fine-tuned DINO~\citep{caron2021emerging} ViT, however they did not perform semantic segmentation, \ie the classes of segments were not predicted.

The most common solution in the literature for weakly supervised segmentation of coral imagery is point label propagation.  \cite{friedman2013automated} first proposed using superpixel segments in coral images. Superpixels were generated using the mean-shift segmentation and edge detection algorithm, and then only the segments containing point labels were used to obtain LBP features and train a SVM classifier. \cite{alonso2017coral} used superpixels to propagate sparse point labels within segments and then used these augmented ground truth labels for fine-tuning a SegNet~\citep{badrinarayanan2017segnet} architecture for segmentation of coral images.  Similarly, \cite{yu2019fast} used superpixels to generate additional point labels, however they selected only a further 10 points to augment each original sparse label, and selected points within superpixel segments based on the intensity gradient of the surrounding pixels to prevent selection at the edge of an instance. 

\begin{figure}[t]
\centering
\setlength{\tabcolsep}{2pt}
\centerline{\begin{tabular}{ccccc}
    \includegraphics[height=31mm]{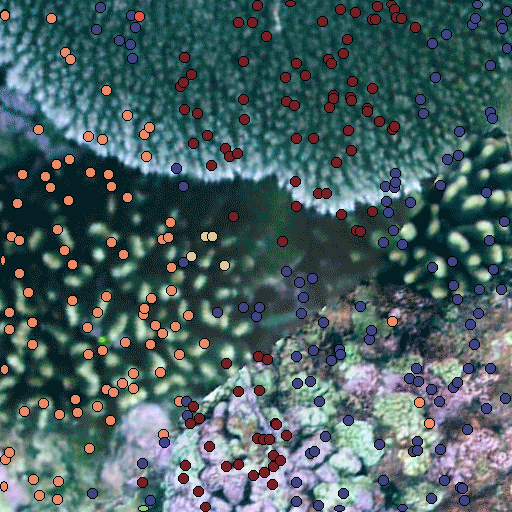} &
    \includegraphics[height=31mm]{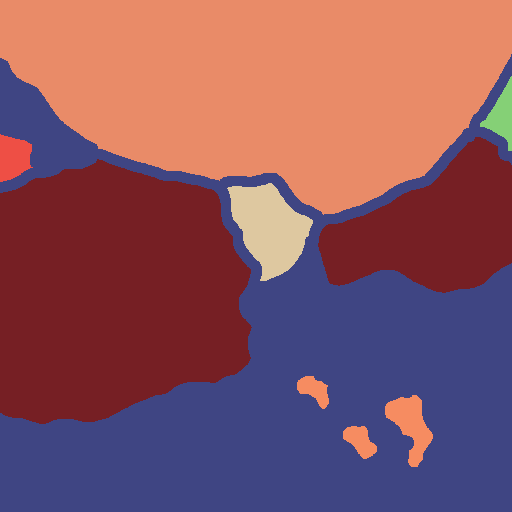} &  \includegraphics[height=31mm]{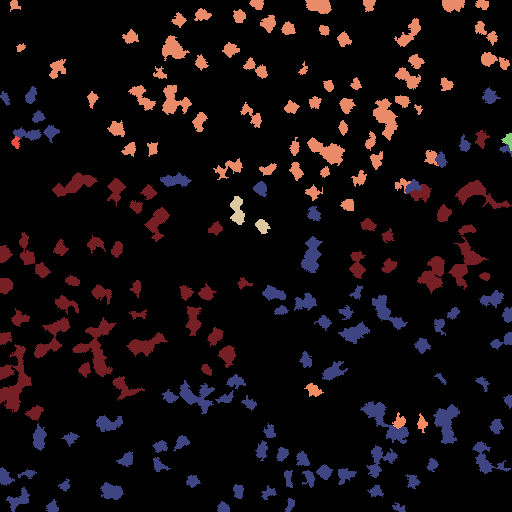} &
    \includegraphics[height=31mm]{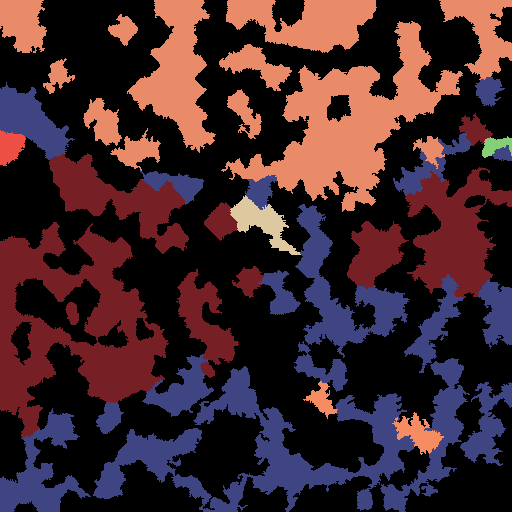}
    &
    \includegraphics[height=31mm]{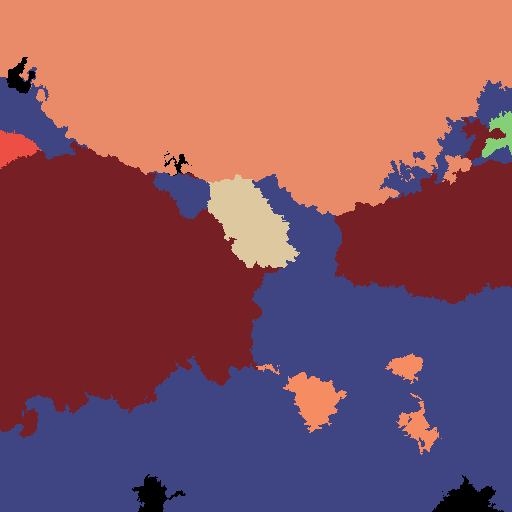}
    \\
    Image & Ground Truth & Step 1 & Step 10 & Step 20 \\
\end{tabular}}
\caption[The Multi-level Superpixel Algorithm]{The multi-level superpixel algorithm~\citep{alonso2018semantic, alonso2019coralseg} extracts superpixels at varying spatial scales and then combines the segments containing point labels using a join operation\footnotemark. This algorithm requires specifying the number of superpixels in the first step, the final step and the number of steps (a larger number of steps yields improved point label propagation but takes longer to execute).}
\label{fig:multi-level}
\end{figure}

\footnotetext{The image used in this visualisation was publicly released as part of the UCSD Mosaics dataset~\citep{edwards2017large, alonso2019coralseg} and reproduced with permission from Springer Nature.}

\cite{alonso2018semantic, alonso2019coralseg} proposed the popular multi-level superpixel algorithm.  Superpixels are extracted at varying spatial scales and then combined using a join operation to produce the final augmented ground truth mask (Fig.~\ref{fig:multi-level}).  Soon after, \cite{pierce2020reducing} re-implemented the same algorithm with a number of improvements for efficiency and computation time per image. These approaches~\citep{alonso2018semantic, alonso2019coralseg, pierce2020reducing} used the augmented ground truth masks to train segmentation models including DeepLabv3+, Fully Convolutional Networks and FC-DenseNet. \cite{yuval2021repeatable} also used the multi-level superpixel algorithm with photogrammetry to generate a 3D segmented map for coral ortho-mosaic imagery.

These approaches largely treat the point labels as a post-processing step following the generation of superpixel segments and the propagation is based only on RGB colour features. \cite{raine2022point} proposed a point label aware approach to superpixels which improved the accuracy and computation time for the propagation task by incorporating the point labels directly and using clustering of deep CNN features to produce segments which more closely conform to complex coral boundaries. 

These approaches for point label propagation are weakly supervised, however they assume that there are 100-300 point labels available in each image.  It is expensive and time-consuming for domain experts to label a large number of points in each image. Recent work \citep{raine2024human} investigates the extremely sparse label setting, \ie when there are only 5-25 point labels available in an image for point label propagation.  In this work, a human-in-the-loop labelling regime is used to propose the most informative points for annotation, significantly increasing the efficiency of the point labels. 

This section has described works for weakly supervised analysis of seagrass and coral imagery.  The discussion in Section~\ref{sec:discussion} provides an analysis of the literature and identifies key gaps, challenges and opportunities for future work. 

\section{Discussion}
\label{sec:discussion}

Although underwater computer vision and deep learning has received an increase in attention due to recent advances in the use of underwater platforms, it remains an under-researched field~\citep{gonzalez2023survey}.  Furthermore, the characteristics of underwater imagery, difficulty of the vision problem, and identified challenges with data availability make the problem of underwater segmentation distinct from the task of segmentation in other mainstream settings (Section~\ref{sec:lit-underwater-imagery}).  This section outlines the key findings of the review, including the challenges, barriers and opportunities for future work.

\subsection{Challenges and Barriers}
\label{sec:disc-challenges-barriers}

This section summarises the key challenges and barriers identified through the analysis of existing literature for weakly supervised segmentation of underwater imagery. 

\begin{figure}[t]
\centering
\centerline{\begin{tabular}{c@{\hspace{0.1cm}}c@{\hspace{0.1cm}}c@{\hspace{0.1cm}}c}
    \includegraphics[height=27mm]{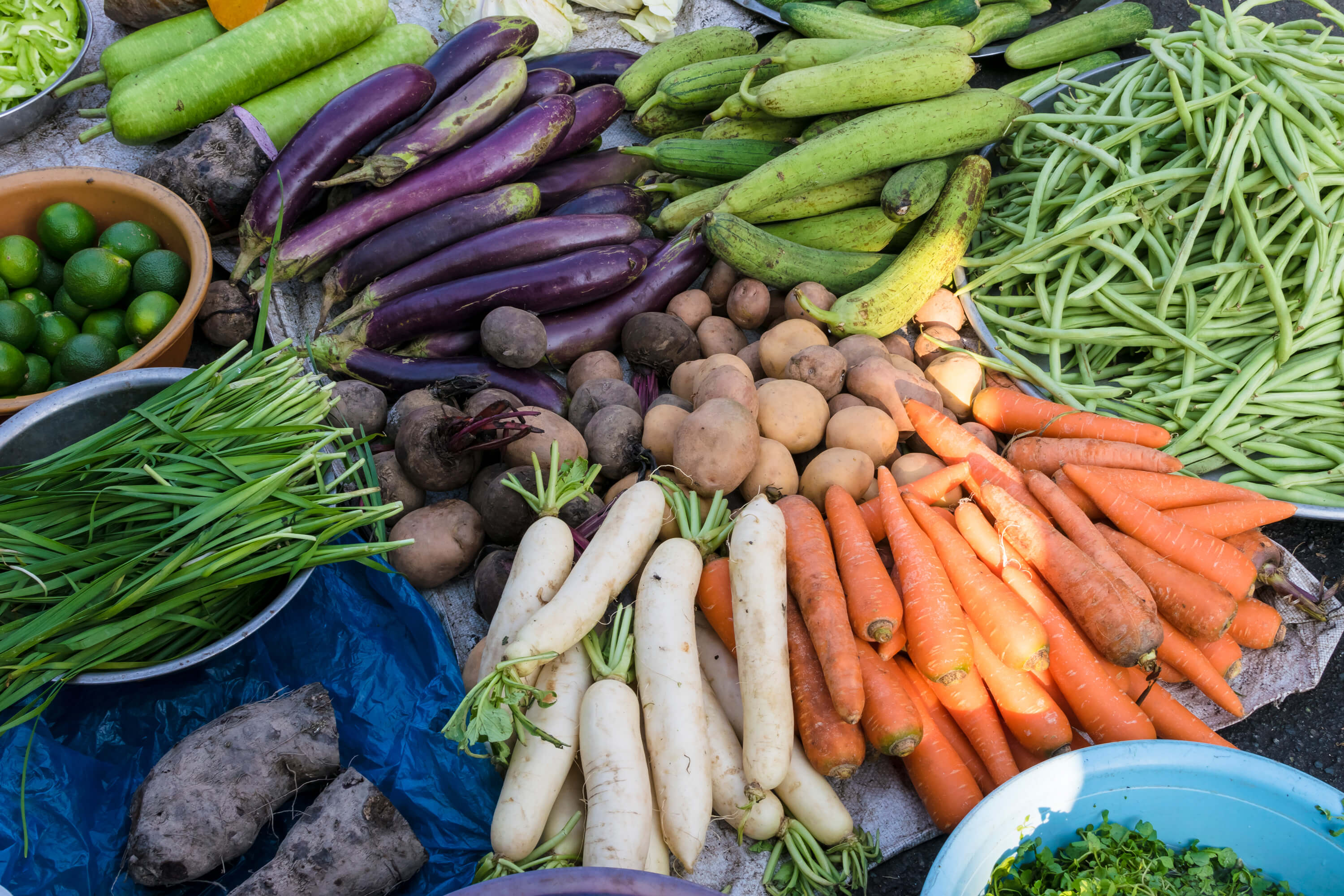} &
    \includegraphics[height=27mm]{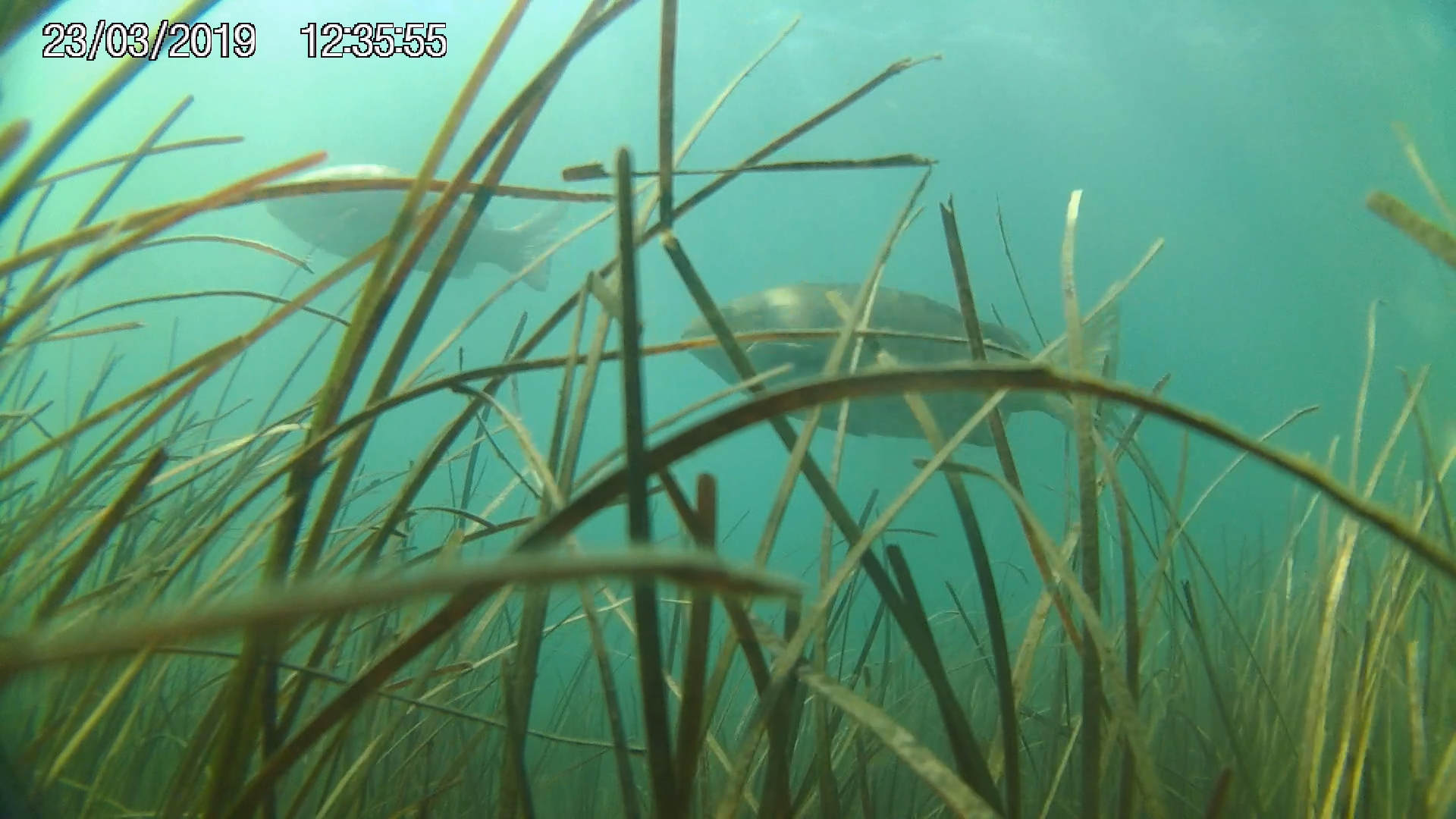} &
    \includegraphics[height=27mm]{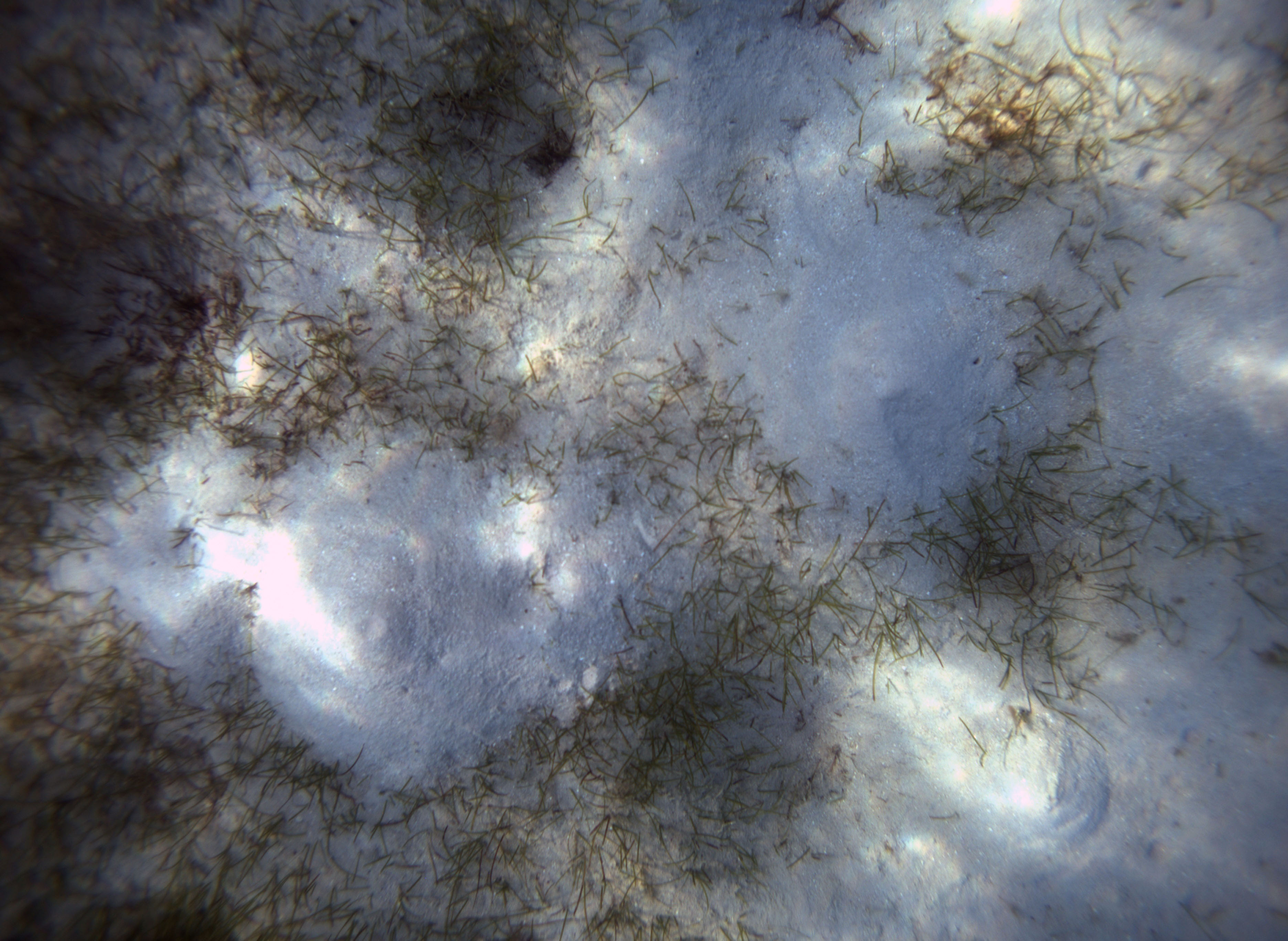} & 
    \includegraphics[height=27mm]{fig19_d1.jpg} 
    \\
    \includegraphics[height=27mm]{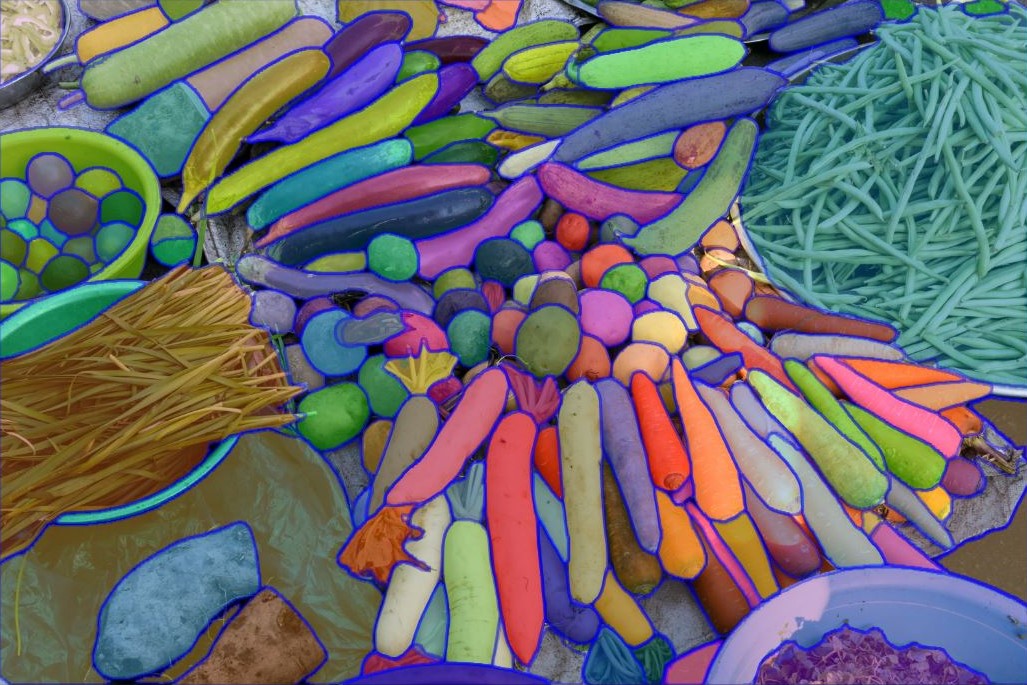} &  
    \includegraphics[height=27mm]{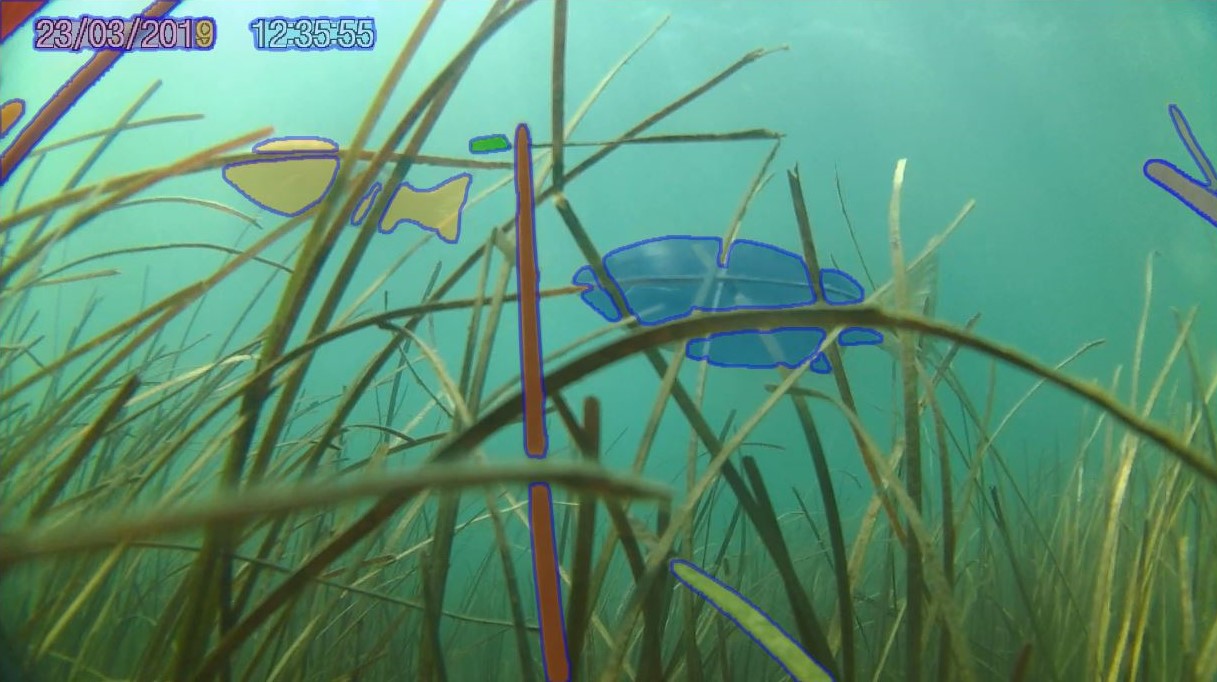} &  
    \includegraphics[height=27mm]{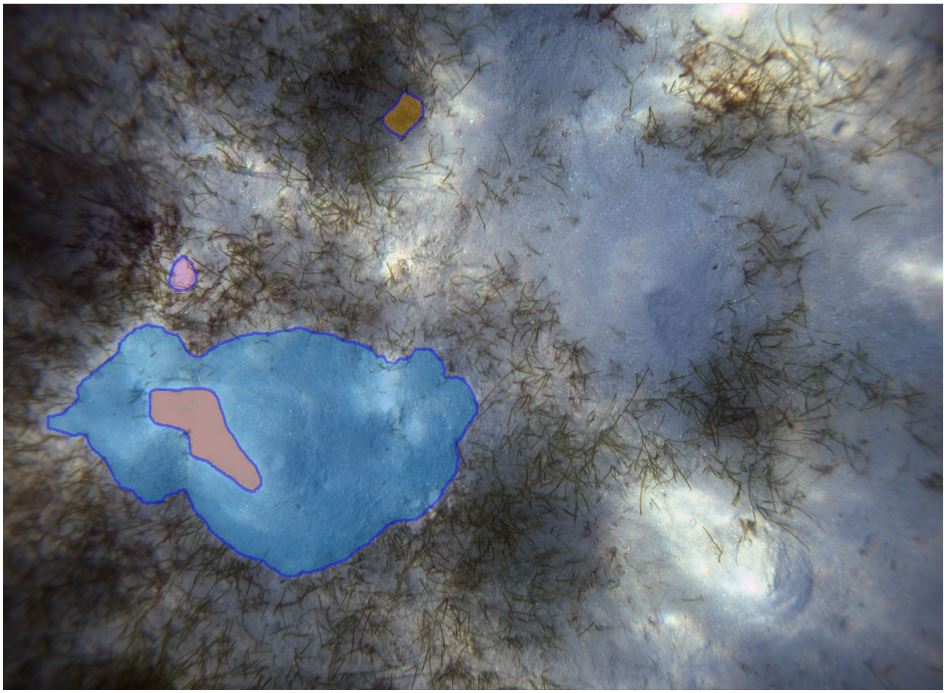} &   
    \includegraphics[height=27mm]{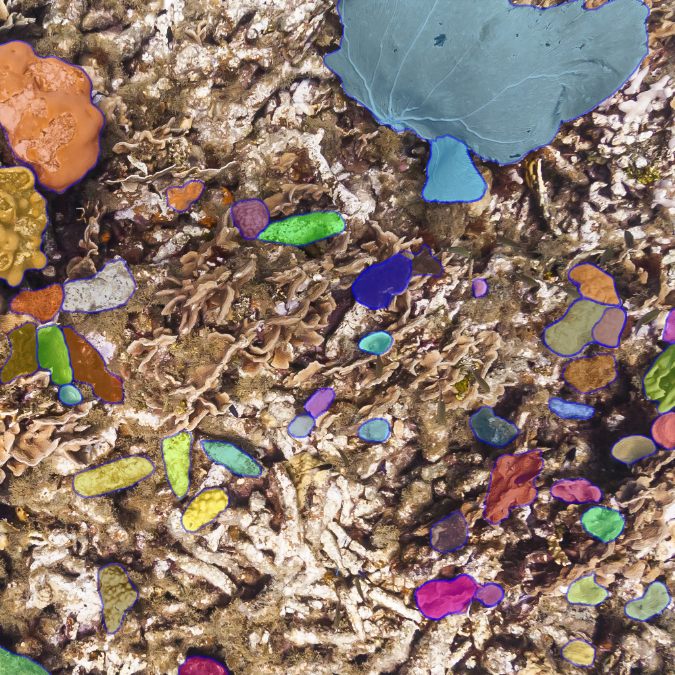}    
    \\
    a) & b) & c) & d)
\end{tabular}}
\caption[Example Predictions from Segment Anything]{A visualisation of the segments generated by the state-of-the-art Segment Anything~\citep{kirillov2023segment} model, demonstrating that underwater images do not exhibit the same hard boundaries or `objectness' as images from mainstream computer vision. The top row are the query images\footnotemark, and the bottom row depicts the segments predicted by the model, where each coloured region indicates a different object. In a), an example image supplied by~\cite{kirillov2023segment} is segmented into distinct objects, whereas in b) only the fish are segmented, and in c) and d) the pixels are not grouped in a meaningful way.}
\label{fig:sam}
\end{figure}

\footnotetext{Image a) released by Meta AI under \href{https://creativecommons.org/licenses/by-sa/4.0/}{CC BY-SA 4.0}, image b) is publicly released as part of the `Global Wetlands' dataset~\cite{ditria2021annotated} under a \href{https://creativecommons.org/licenses/by/4.0/}{CC BY 4.0 licence}, image c) was taken by Serena Mou using the `FloatyBoat' surface vehicle~\cite{mou2022reconfigurable} and used with permission, and d) is publicly released as part of the XL CATLIN Seaview Survey~\cite{gonzalez2014catlin, gonzalez2019seaview} under a \href{https://creativecommons.org/licenses/by/3.0/deed.en}{CC BY 3.0 licence}.}

\subsubsection{Semantic Characteristics}

Section~\ref{sec:lit-weakly} described a number of weakly supervised works in the mainstream field of computer vision, however these approaches are not well suited to the application of underwater imagery, where target classes are frequently overlapping, with unclear visual boundaries and no distinction between foreground and background. As outlined in Section~\ref{sec:lit-underwater-imagery}, underwater imagery is cluttered and complex, and the species boundaries are poorly defined. There is no clear foreground/background for underwater imagery, and the approaches designed for mainstream applications often fail to group pixels in a meaningful way.

\begin{figure}[t]
\centering
\centerline{\begin{tabular}{c@{\hspace{0.1cm}}c@{\hspace{0.1cm}}c@{\hspace{0.1cm}}c}
    \includegraphics[height=27mm]{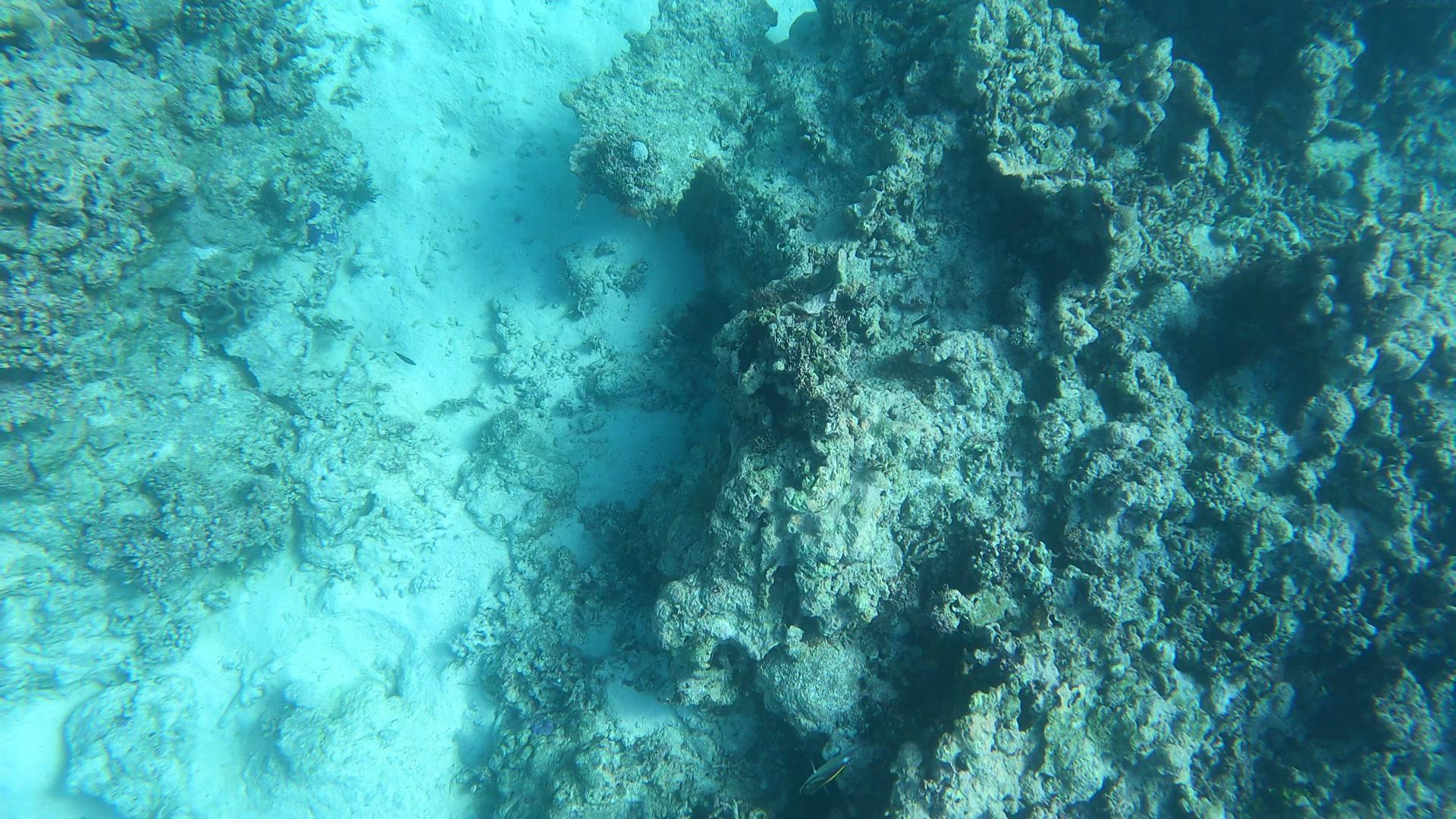} &
    \includegraphics[height=27mm]{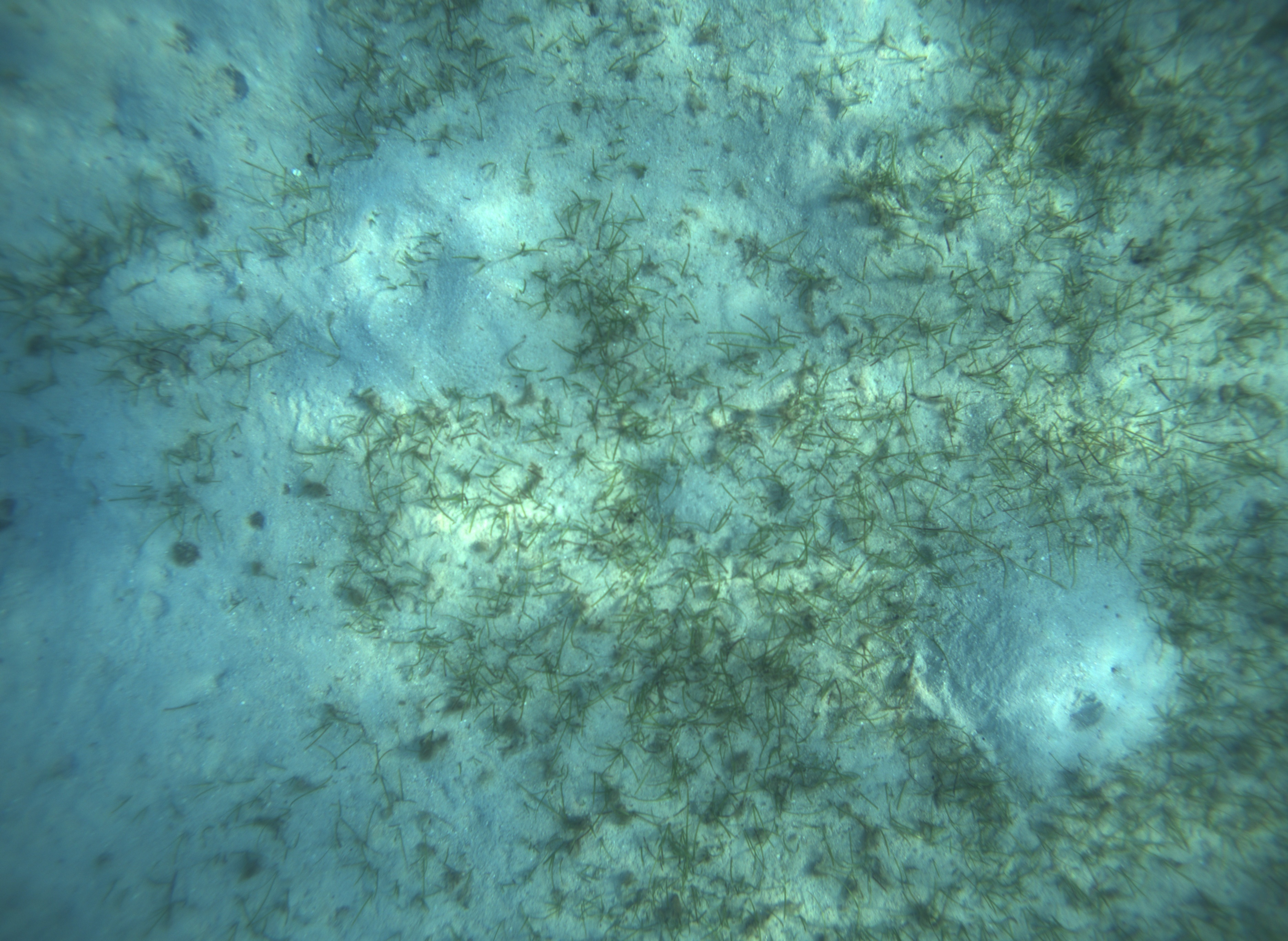} &
    \includegraphics[height=27mm]{fig20_c1.jpg} & 
    \includegraphics[height=27mm]{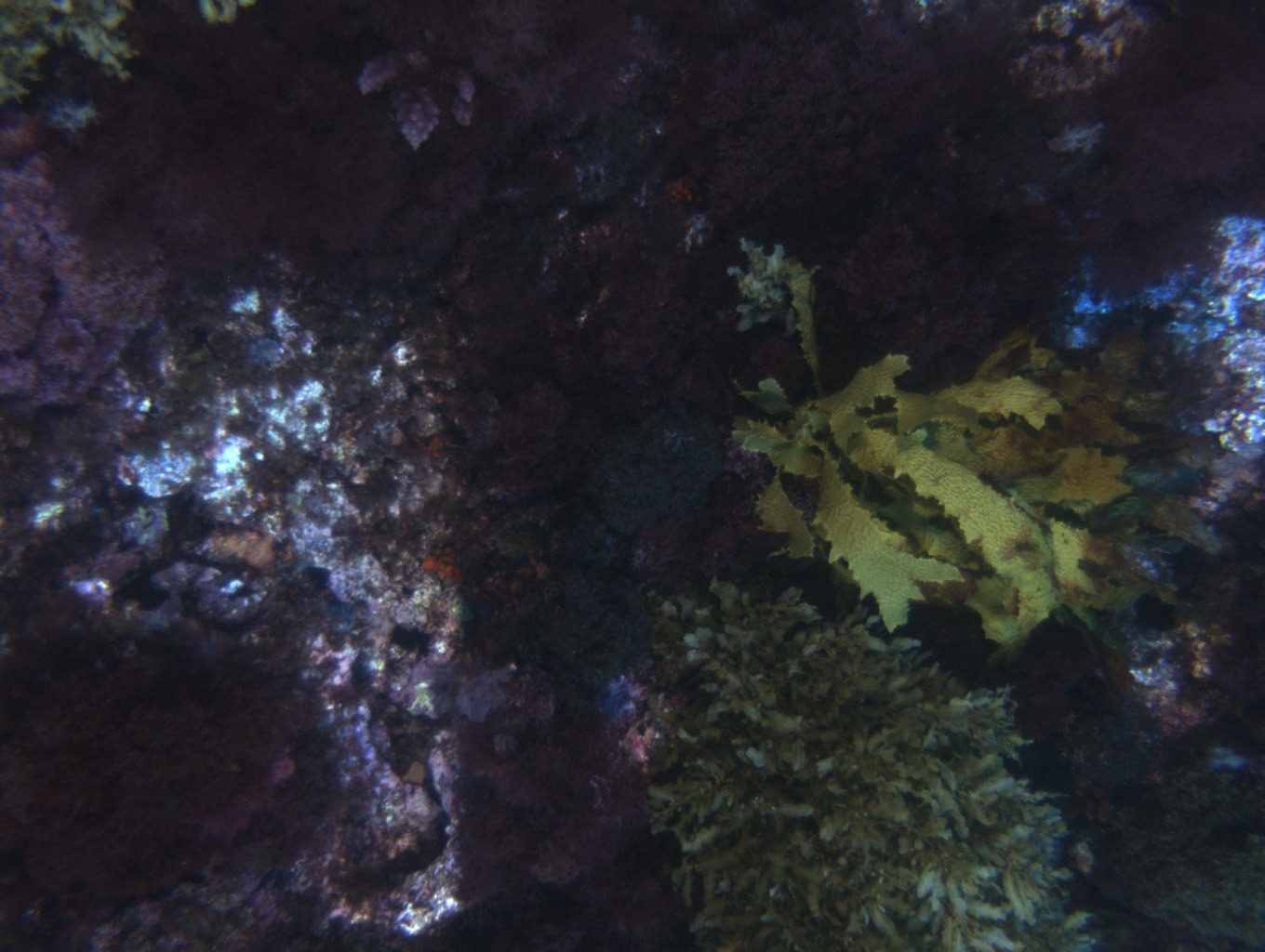} 
    \\
    \includegraphics[height=27mm]{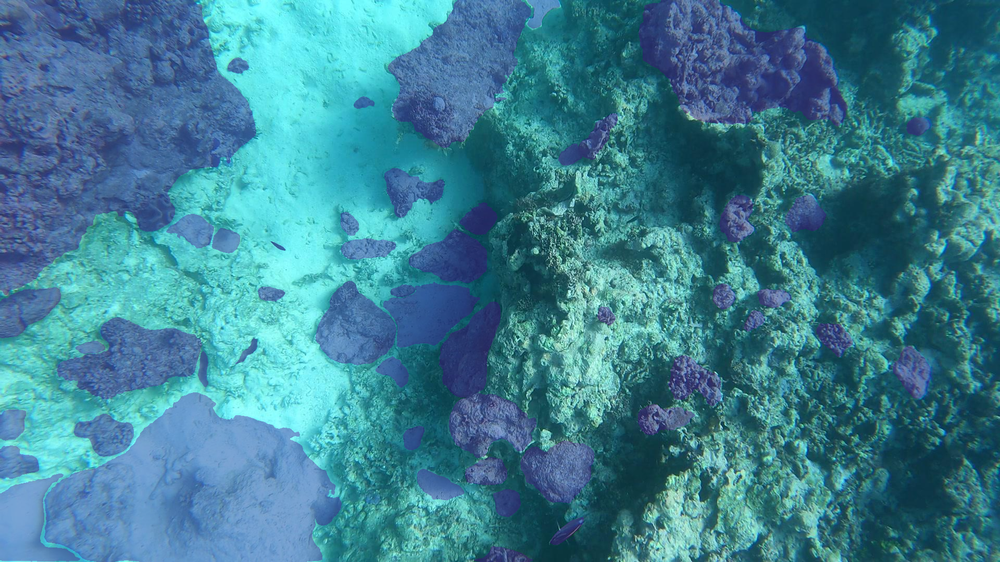} &  
    \includegraphics[height=27mm]{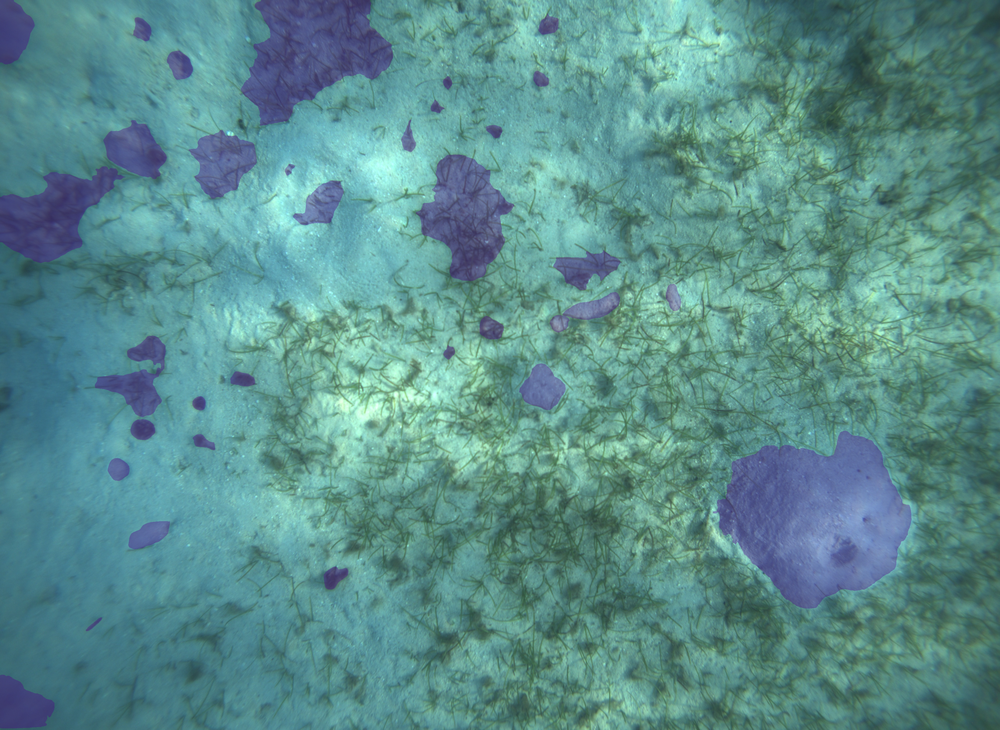} &  
    \includegraphics[height=27mm]{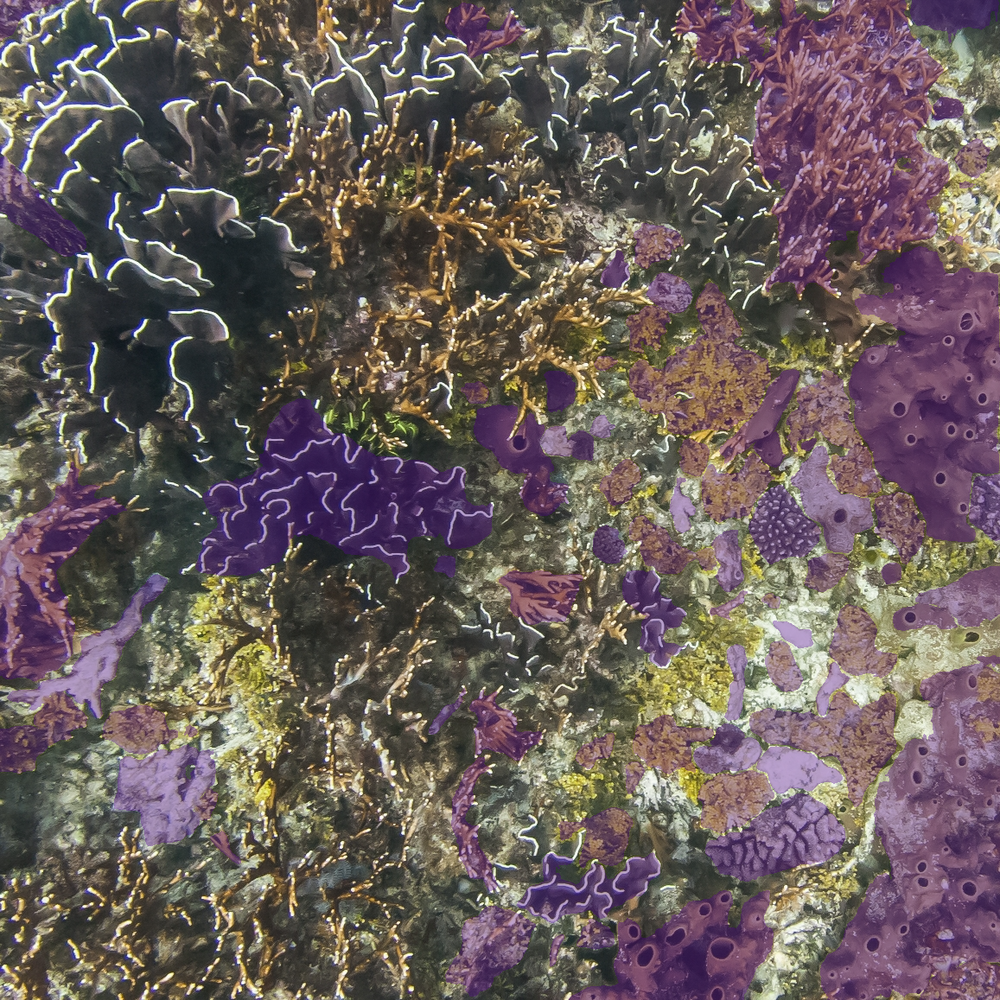} &   
    \includegraphics[height=27mm]{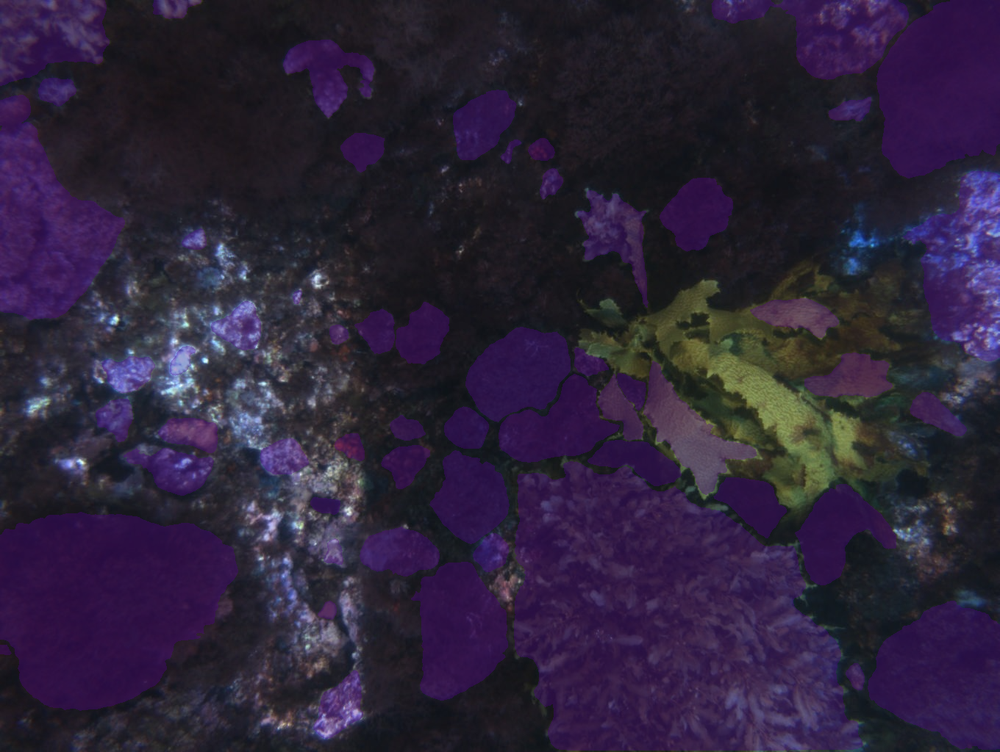}    
    \\
    a) & b) & c) & d)
\end{tabular}}
\caption[Example Predictions from CoralSCOP]{A visualisation of the segments generated by the CoralSCOP \citep{zheng2024coralscop} model, a coral foundation model designed to produce instance segments in any coral image\footnotemark. Even custom-designed approaches struggle to produce meaningful segments in deep water (a), poorly defined (b), cluttered (c), or dark (d) images.  The extreme variation in image characteristics for underwater imagery makes development of general models challenging.}
\label{fig:coralscop}
\end{figure}

\footnotetext{Image a) was publicly released in the `Underwater Crown-of-Thorn Starfish (COTS) imagery sample dataset'~\citep{liu2022underwater} released under a \href{https://creativecommons.org/licenses/by/4.0/}{CC BY 4.0 licence}, image b) was taken by Serena Mou using the `FloatyBoat' surface vehicle~\citep{mou2022reconfigurable} and used with permission, image c) is publicly released as part of the XL CATLIN Seaview Survey~\citep{gonzalez2014catlin, gonzalez2019seaview} under a \href{https://creativecommons.org/licenses/by/3.0/deed.en}{CC BY 3.0 licence} and image d) is part of the BENTHOZ 2015 dataset \citep{bewley2015australian} and released under a \href{https://creativecommons.org/licenses/by/4.0/}{CC BY 4.0 licence}.}

This is demonstrated in Fig.~\ref{fig:sam}, in which the state-of-the-art Segment Anything\footnote{\url{https://segment-anything.com/demo}}~\citep{kirillov2023segment} model is deployed on a range of underwater images.  The model effectively segments an image of vegetables (shown in Fig.~\ref{fig:sam}a), however it does not produce useful segments in the seagrass and coral images (Fig.~\ref{fig:sam}b-d). \cite{raine2024human} effectively used the DINOv2 foundation model with clustering in the deep embedding space, but this required a small quantity of sparse point labels provided by a domain expert as a weak supervisory signal.  The recent CoralSCOP \citep{zheng2024coralscop} model, which is a foundation model specifically for instance segmentation of coral images, suffers degraded performance on images impacted by depth, turbidity or poor lighting, as seen in Fig.~\ref{fig:coralscop}.

\subsubsection{Data Availability}
\label{subsubsec:disc-data}

A review of the available underwater image datasets revealed a shortage of high quality, annotated data for both multi-species seagrass and coral imagery. The only available multi-species seagrass datasets are labelled with patches. Additionally, Section~\ref{subsec:lit-underwater-datasets} highlighted the shortage of multi-species coral datasets annotated with dense ground truth labels for training and evaluating segmentation models. The dataset survey in Appendix~\ref{appendix:A} outlined numerous coral datasets labelled with sparse point labels, and therefore this type of data was identified as an important source of domain expert knowledge. Approaches which leverage this type of data for coral segmentation through point label propagation were described in Section~\ref{sec:weakly-underwater}. The recent CoralVOS dataset \citep{ziqiang2023coralvos} is the first video object segmentation dataset for the underwater domain, however it only contains binary masks for the presence/absence of corals. 

There is a significant gap between datasets of static, scaled and often corrected imagery, and the images collected by an underwater vehicles or towed camera system as it moves dynamically through the water. There is a distinct lack of temporal, realistic underwater imagery accompanied by dense, accurate ground truth labels for training and evaluation of approaches.  Of most critical importance is the available of labelled test sequences of imagery annotated densely at the pixel-level to enable fair and accurate evaluation of multi-species segmentation approaches.

Datasets of this type are needed for both seagrass and coral ecosystems.  Seagrass monitoring and mapping is a growing field of research which currently lacks any dataset with temporal images and dense, multi-species annotations.  Although coral data has attracted more attention and there are more coral datasets available generally, there still exists a clear lack of this type of temporal, densely annotated data to enable realistic and accurate comparison of approaches.  As coral segmentation attracts more attention, due to momentum in the research field, the increasing threats of climate change and coral bleaching, and the rising importance of reef restoration, availability of high quality, labelled data is critical.  

In addition to temporal video sequences of data, further research should focus on dataset design which incorporates multiple repeats of the same transect over a longer period of time \eg repeating the same transect at monthly intervals for a year, to capture the differences in seasonality and reef health over time. These datasets should have accompanying GPS information. This would enable evaluation of change detection and quantification approaches. 

Furthermore, availability of realistic video sequences annotated with dense, pixel-wise ground truth annotations could have implications for other tasks in underwater robotics including underwater visual place recognition (VPR), localisation and mapping~\citep{li2015high, boittiaux2023eiffel, keetha2023anyloc}.

\subsubsection{Desktop to Reality Gap}
\label{subsubsec:disc-desktop-reality}

The ``Desktop to Reality'' (also called the ``Sim2Real'') gap refers to the disconnect between models trained on datasets and their performance in real-world applications. This gap is particularly pronounced in underwater imagery, where conditions are highly variable and unstructured. Models often fail to generalise when deployed in the field, where factors like lighting, water clarity, and terrain differ dramatically. This gap poses a significant challenge for tasks like semantic segmentation, which require accurate, fine-grained identification and classification of underwater features.

Public datasets available online for underwater imagery often fail to reflect the real-world conditions encountered by underwater vehicles. These datasets are typically collected with high resolution cameras, controlled angles, zoom levels, scaling, artificial lighting, and post-processing corrections applied. However, when performing real-time inference, underwater vehicles cannot replicate these adjustments, leading to a mismatch between the training data and the actual environment the model encounters during deployment. Models trained on publicly available datasets may struggle to generalise and fail when faced with the unaltered, dynamic conditions of underwater environments.

Even when datasets are collected using the same platform and setup as planned for deployment, the conditions during testing are often not the same. Variations in location, water turbidity, natural lighting, caustic effects, and seasonal changes can significantly alter the visual characteristics of the scene. These environmental factors make it challenging for segmentation algorithms to accurately identify and delineate species in underwater images.  

\subsection{Future Work}
\label{subsec:disc-future-work}

This section outlines potential avenues for future work, based on the findings of this survey. 

\subsubsection{Leveraging Temporal Information}
\label{subsubsec:disc-future-work-temporal}

Many approaches for underwater semantic segmentation consider underwater imagery as a collection of independent images, largely due to data availability limitations. If datasets of labelled multi-species underwater videos become available, the incorporation of spatio-temporal continuity between consecutive frames could improve accuracy.  Forward filtering could improve model predictions when deployed on an underwater or surface vehicle, as the previous predictions of the model could be used to inform future predictions \ie if the model had previously seen a certain species, it would be likely to encounter it again during the survey~\citep{sunderhauf2018limits}.  

Targets in the distance can be difficult to distinguish, but as an AUV or ROV moves through the water and is closer to the target, the image clarity improves and the model performance improves.  This is known as backward smoothing, and incorporates temporal information to improve the accuracy of image analysis.  Furthermore, temporal information can be used to track targets through the frame, enabling accurate counting of individual instances.

\subsubsection{Large Language Models}
\label{subsubsec:disc-future-work-llm}
Large language models can be effectively leveraged for domain-specific tasks, as demonstrated in~\cite{raine2024image} and~\cite{zheng2023marinegpt}.  \cite{raine2024image} demonstrated that a large vision-language model could be used as a supervisory signal during training of model for multi-species coarse seagrass segmentation. Future work could investigate the potential of vision-language models to assist in recognition of previously unseen marine species where the ecologist is able to describe a target of interest.  One recent work in this area is by~\cite{zheng2023marinegpt}, who contribute MarineGPT, a vision-language model designed for the marine domain.  Their implementation performs image captioning for marine images sourced from the internet. However, they do not evaluate the performance of their approach on realistic imagery collected by underwater vehicles, and also do not use the model to detect targets in footage based on natural language descriptions provided by ecologists. 

A possible architecture could incorporate a large language model as the interpreter of the ecologist -- the ecologist describes the target of interest and the model converts this into some feature representation used to detect the target in the underwater imagery.  This framework was used in the SeaCLIP approach in \cite{raine2024image} for pseudo-labelling at training time, but it could be extended for detection of previously unseen targets during deployment. This could be transformative for open-set recognition and anomaly detection in the underwater domain, and significantly improve the flexibility and generalisation of approaches. 

\subsubsection{Improving Label Resolution}
\label{subsubsec:disc-future-work-labels}
Many approaches for coarse seagrass segmentation \citep{raine2020multi, raine2024image, noman2021multi, ozaeta2023seagrass, paul2023lwds} predict broad taxonomic morphotypes of seagrass \ie Strappy, Ferny, Rounded and Background. These taxonomic morphotypes group together species of seagrass which have the same morphology, \ie species which have round leaves of a similar appearance would be grouped together (\textit{Halophila ovalis} and \textit{Halophila decipiens}, depicted in Fig.~\ref{fig:seagrass-species}, were considered in the same `Rounded' morphotype). In future work, these classes could be un-grouped such that the model is trained on a wider range of seagrass species and performs inference at this higher taxonomic resolution.  This would necessitate a data collection effort to obtain sufficient samples of each species, but would provide more detail on the species present and their abundance.  A greater level of detail would enable more accurate estimation of blue carbon stocks. 

Furthermore, the imagery used to train approaches \citep{raine2020multi, raine2024image, noman2021multi, ozaeta2023seagrass, paul2023lwds} contained only seagrass growing densely and images with only a small area of seagrass were not used for training.  For accurate estimation of blue carbon sequestered, the density of the seagrass is a relevant factor.  Future work could include extending the approaches to enable estimation of seagrass growth stage, by training models on different densities of seagrass (\eg sparse, medium, dense coverage).

\subsubsection{Hierarchical Labels}

In the case of coral segmentation, the extremely large number of coral species (there are more than 800 hard coral species globally~\citep{dietzel2021population}) prevents training models to predict all species.  A hierarchical approach could be developed, in which predictions are made at different levels depending on the model's certainty.  The CATAMI scheme~\citep{althaus2015standardised} is a flexible, hierarchical classification taxonomy for Australian underwater species, which could be used to provide classes for hierarchical analysis of imagery.  For example, if the model was uncertain about the species present, it could predict a high-level grouping \eg `macroalgae', or if it had a high degree of certainty the model could predict the species of the macroalgae. 

\subsubsection{Continuous Labels}

In the natural environment, it is common for variables to be continuous.  For example, if a coral has experienced bleaching, there is a spectrum of severity in which it could recover quickly, or the individual could fail to recover and die.  Similiarly, the spectrum from dense seagrass coverage to sparse seagrass coverage is continuous, and defining discrete categories, and labelling images into these categories, is difficult. Another example is when an individual of some sort, \eg a coral polyp, goes from a early stage of development to a more developed form.  Ecologists often require some indication of these variables.  One interesting avenue for future work would be to perform segmentation of imagery into continuous classes (healthy--dead, dense--sparse coverage, early--mature stage of development).  This has the potential to provide a wealth of information for ecologists in a wide range of applications and settings. 

\subsubsection{Human-Robot Collaboration}
\label{subsubsec:disc-future-work-HRC}
\cite{raine2024human} proposed a Human-in-the-Loop approach to facilitate collaboration between the model and ecologist, significantly reducing the number of labels required during training. Future works could further leverage principles of human-robot collaboration to inform iterative improvement of models as species of interest are encountered during field trial deployment. In this scenario, a deep learning model could be deployed on a robotic platform and used to predict previously unseen targets in the footage. 

Recently, \cite{bijjahalli2023semi} performed semi-supervised anomaly detection in underwater images based on variational autoencoders and clustering features.  An extension of this idea would be to present identified anomalies to a domain expert in real-time. The expert determines whether the class is of interest and if so, specifies the class label. As further examples are detected, the examples could be incorporated in an active learning framework where the model iteratively incorporates additional classes.

Underwater anomaly detection could have significant implications for species discovery and would enable models to be quickly adapted to new locations and species encountered, rather than exhibit degraded performance when deployed in a different setting from the training data. 

\subsubsection{Geographical Prior Knowledge}
\label{subsubsec:disc-future-work-knowledge}

There can be significant variation between the visual characteristics of images collected at different geographic locations.  The visual differences can be caused by the physical growth of the species, as impacted by the environmental conditions at that location \eg whether the site is on the windward or leeward side of the reef, ocean temperatures, damage from tropical cyclones, algae growth rates or prevalence of the Crown of Thorns starfish.

Ecologists have significant prior knowledge about where species are likely to occur and how the visual traits of species change based on location and environmental factors.  Future work could involve incorporating this prior knowledge and other information such as GPS location as additional features into the segmentation model's training or by combining prior class probabilities with model outputs post-hoc.

\subsubsection{General Underwater Scene Understanding}
\label{subsubsec:disc-future-work-general}

Despite increasing attention on underwater perception in the research community and many researchers sharing large deep learning models and feature extractors publicly~\citep{chen2021new, zheng2024coralscop, gonzalez2020monitoring, raine2024human}, there is currently no publicly available, pre-trained, general segmentation model which can be downloaded and used out-of-the-box to perform semantic segmentation of any underwater imagery.  All approaches are specific to the image characteristics and classes present in the training data.  The CoralSCOP foundation model is the closest model to this in the literature, however it requires some input from the user in the form of points, a text description or  referring masks~\citep{zheng2024coralscop}. It would be beneficial to have a publicly available, pre-trained model able to perform inference at coarse-level categories on underwater imagery of any style and from anywhere in the world.  There are many contexts in underwater monitoring where it would be useful to quickly generate estimates of the coverage of broad categories including hard coral, soft coral, algae, macro-algae, rubble, sand, bleached coral, consolidated substrate.  

Future work could include designing and training a general model to perform semantic segmentation on a wide variety of image types at a coarse-grained category level.  This would enable fast quantification of changes in reef health \eg easily comparing the amount of bleaching or the growth of algae between consecutive surveys without requiring costly data annotation and model re-training every time.

\section{Conclusion}
\label{sec:conc}

This survey underscores the critical need for efficient underwater image analysis to meet the increasing demands of marine surveys performed by robotic underwater and surface vehicles. The complexity and diversity of underwater imagery, along with the reliance on domain experts for detailed annotations, present significant challenges to scaling current methods. By reviewing advances in deep learning and computer vision, particularly weakly supervised and unsupervised approaches, we highlight promising directions for reducing the dependency on fully labelled data. The taxonomy presented in this survey provides a structured framework for addressing these challenges, while the evaluation of existing datasets and platforms emphasizes the gaps that remain in the field. Future research must focus on developing more robust, scalable models that can operate with minimal human input, ultimately enhancing our ability to monitor coastal ecosystems and support sustainable management practices for underwater environments.  We hope this work forms a comprehensive overview of weakly supervised deep learning for coral and seagrass imagery, and that this encourages further interest in reducing label dependency for monitoring underwater environments.

\backmatter

\section*{Declarations}

\subsection*{Funding}
S.R., F.M., N.S., and T.F.~acknowledge continued support from the Queensland University of Technology (QUT) through the Centre for Robotics. T.F.~acknowledges funding from an Australian Research Council (ARC) Discovery Early Career Researcher Award (DECRA) Fellowship DE240100149 to T.F.  

S.R. and T.F.~ acknowledge support from the Reef Restoration and Adaptation Program (RRAP).  The Reef Restoration and Adaptation Program aims to develop effective interventions to help the Reef resist, adapt to, and recover from the impacts of climate change.  Partners include: the Australian Institute of Marine Science, the Great Barrier Reef Foundation, CSIRO, The University of Queensland, Queensland University of Technology, Southern Cross University and James Cook University.

\subsection*{Conflict of interest/Competing interests}
All authors certify that they have no affiliations with or involvement in any organisation or entity with any financial interest or non-financial interest in the subject matter or materials discussed in this manuscript.

\subsection*{Ethics approval and consent to participate}
Not applicable

\subsection*{Data availability}
No new datasets were generated or analysed in this survey paper.

\subsection*{Code availability}
Not applicable

\clearpage
\begin{appendices}

\section{Survey of Publicly Available Datasets of Underwater Imagery} 
\label{appendix:A}

This appendix outlines a survey of the publicly available datasets containing labelled underwater imagery.  First, a survey of the available underwater seagrass image datasets (Table \ref{table:datasets-seagrass}) is performed, followed by a survey of underwater coral imagery (Table \ref{table:datasets-coral}).  For each dataset, the name and citation, year of data collection, image viewpoint, classes used and quantity of data is provided.  Note that the quantity of data is the number of images which have accompanying labels.  The datasets are also grouped into their annotation types: image-level labels, patch-level labels, sparse point labels and dense pixel-wise masks.  

\begin{longtable}{>{\raggedright\arraybackslash}p{0.26\textwidth} >{\raggedright\arraybackslash}p{0.1\textwidth} >{\raggedright\arraybackslash}p{0.18\textwidth} >{\raggedright\arraybackslash}p{0.15\textwidth} >{\raggedright\arraybackslash}p{0.19\textwidth}}
\caption{Survey of Publicly Available Datasets of Seagrass Imagery} \\
\label{table:datasets-seagrass} \\
    \toprule
    \textbf{Dataset/Paper} & \textbf{Year} & \textbf{Viewpoint} & \textbf{Classes} & \textbf{Images} \\
    
    \midrule
    \multicolumn{5}{c}{\textbf{Image-Level Labels}} \\
    \midrule
    \arrayrulecolor{gray!30}

    Benthic and substrate cover data derived from a time series of photo-transect surveys for the Eastern Banks, Moreton Bay Australia \citep{roelfsema} & 2004-2015 & Mainly top-down & 41 classes (5 seagrass) & ~24,000 images \\
    
    \arrayrulecolor{black}
    \midrule
    \multicolumn{5}{c}{\textbf{Patch-Level Labels}} \\
    \midrule
    \arrayrulecolor{gray!30}

    DeepSeagrass \citep{raine2020multi, raine2021deepseagrass} & 2020 & Oblique & 5 classes &  66,946 patches \\
    \midrule
    Global Wetlands: Luderick Seagrass \citep{ditria2021annotated, raine2024image} & 2021 (original) and 2024 (test patches) & Forward & Binary fish polygons (original) and 3 classes for test patches (\textit{Background, Fish, Seagrass}) & 4280 images (original) and 38,200 labelled test patches \\

    \arrayrulecolor{black}
    \midrule
    \multicolumn{5}{c}{\textbf{Bounding Box Labels}} \\
    \midrule
    \arrayrulecolor{gray!30}
    
    ECUHO-1 \citep{ECUHO} & 2019 & Oblique & 1 class (\textit{Halophila ovalis}) & 2,699 images \\
    \midrule
    ECUHO-2 \citep{ECUHO} & 2019 & Oblique & 1 class (\textit{Halophila ovalis}) & 160 images \\   
    
    \arrayrulecolor{black}
    \midrule
    \multicolumn{5}{c}{\textbf{Pixel-wise Labels}} \\
    \midrule
    \arrayrulecolor{gray!30}

    Looking for Seagrass \citep{reus2018looking} & 2018 & Top-down & Binary &  6,037 images \\

    \arrayrulecolor{black} 
    \bottomrule
\end{longtable}

\begin{longtable}{>{\raggedright\arraybackslash}p{0.26\textwidth} >{\raggedright\arraybackslash}p{0.1\textwidth} >{\raggedright\arraybackslash}p{0.18\textwidth} >{\raggedright\arraybackslash}p{0.15\textwidth} >{\raggedright\arraybackslash}p{0.19\textwidth}}
\caption{Survey of Publicly Available Datasets of Coral Reef Imagery} \\
\label{table:datasets-coral} \\
    \toprule
    \textbf{Dataset/Paper} & \textbf{Year} & \textbf{Viewpoint} & \textbf{Classes} & \textbf{Images} \\
    
    \midrule
    \multicolumn{5}{c}{\textbf{Image-Level Labels}} \\
    \midrule
    \arrayrulecolor{gray!30}
    Benthic and substrate cover data derived from a time series of photo-transect surveys for the Eastern Banks, Moreton Bay Australia \citep{roelfsema} & 2004-2015 & Mainly top-down & 41 classes & ~24,000 images \\ 
    \arrayrulecolor{gray!30} %
    \midrule
    StructureRSMAS \citep{gomez2019coral} & 2019 & Mainly oblique, variety &  14 classes & 409 images\\
    
    \arrayrulecolor{black}
    \midrule
    \multicolumn{5}{c}{\textbf{Patch-Level Labels}} \\
    \midrule
    \arrayrulecolor{gray!30}
    EILAT \citep{loya2004coral, shihavuddin2017dataset} & 2004 & Top-down & 8 classes & 1,123 patches \\
    \midrule
    EILAT2 \citep{loya2004coral, shihavuddin2017dataset} & 2017 & Top-down & 5 classes & 303 patches \\
    \midrule
    RSMAS \citep{shihavuddin2013image} & 2017 & Top-down & 14 classes & 766 patches \\
    \midrule
    Benthic substrate classification ML training image snips \citep{benthic2023csiro, jackett2023benthic} & 2023 & Top-down & 6 classes & 69,846 patches \\ 

    \arrayrulecolor{black}
    \midrule
    \multicolumn{5}{c}{\textbf{Bounding Box Labels}} \\
    \midrule
    \arrayrulecolor{gray!30}

    FathomNet \citep{katija2022fathomnet} & 30+ years & Variety & 2244 hierarchical ``concepts" & 84,454 images \\
    \midrule
    DUO \citep{liu2021dataset} & 2021 & Variety & 4 classes & 7,782 images \\
    \midrule
    MAS3K \citep{li2020mas3k} & 2020 & Variety & 7 classes, 37 sub-classes & 3,103 images \\
    \midrule
    The CSIRO Crown of Thorns Starfish Detection Dataset \citep{liu2021csiro} & 2021 & Oblique & 1 class & 35,000 images \\
    
    \arrayrulecolor{black}
    \midrule
    \multicolumn{5}{c}{\textbf{Sparse Point Labels}} \\
    \midrule
    \arrayrulecolor{gray!30}
    BENTHOZ-2015 \citep{IMOS, bewley2015australian} & 2015 & Top-down &  CATAMI hierarchy (148 classes) &  9,874 images \\
    \midrule
    Moorea Labeled Corals (MLC) \citep{beijbom2012automated} & 2008 & Top-down & 9 classes &  2,055 images \\
    \midrule
    Pacific Labeled Corals (overlap with MLC) & 2005-2012 & Top-down & 20 classes &  5,090 images \\
    \midrule
    CoralNet \citep{beijbom2015towards} & To present & Mainly top-down & Dependent on specific set & Dependent on specific set \\
    \midrule
    XL CATLIN Seaview Survey \citep{gonzalez2019seaview, gonzalez2016scaling} & 2012-2018 & Top-down & 64 classes &  11,387 images  \\
    \midrule
    Tasmania Coral Point Count Dataset \citep{bewley2012automated} & 2010 & Top-down &  36 classes & ~1,000 images \\
    \midrule
    EILAT Fluorescence \citep{beijbom2016improving} & & Top-down & 10 classes & 212 images \\  
    
    \arrayrulecolor{black}
    \midrule
    \multicolumn{5}{c}{\textbf{Pixel-wise Labels}} \\
    \midrule
    \arrayrulecolor{gray!30}
    UCSD Mosaics Dataset \citep{edwards2017large} as processed by \citep{alonso2019coralseg} & 2017 & Top-down & 35 classes & 4,922 images \\
    \midrule
    SUIM dataset \citep{islam2020semantic} & 2020 & Oblique &  8 classes &  1,635 images \\
    \midrule
    CoralVOS \citep{ziqiang2023coralvos} & 2023 & Oblique & Binary & 150 videos (60,456 frames) \\
    \midrule
    \#DeOlhoNosCorais \citep{furtado2023deolhonoscorais}  & 2014-2021 & Variety & 21 classes & 1,411 images \\
    \midrule
    RMAS \citep{fu2023masnet} (images from: SUIM \citep{islam2020semantic}, UFO \citep{islam2020simultaneous}, DeepFish \citep{qin2016deepfish}) & Variety & Oblique & 5 classes & 3,041 images \\

    \arrayrulecolor{black} 
    \bottomrule
\end{longtable}

\end{appendices}

\clearpage

\bibliography{sn-bibliography}%

\begin{thebibliography}{249}
\providecommand{\natexlab}[1]{#1}
\providecommand{\url}[1]{{#1}}
\providecommand{\urlprefix}{URL }
\providecommand{\doi}[1]{\url{https://doi.org/#1}}
\providecommand{\eprint}[2][]{\url{#2}}
 \bibcommenthead

\bibitem[{Abid et~al(2024)Abid, Noman, Kov{\'a}cs, Islam, Adewumi, Lavery, Shafait, and Liwicki}]{abid2024seagrass}
Abid N, Noman MK, Kov{\'a}cs G, et~al (2024) Seagrass classification using unsupervised curriculum learning (ucl). Ecological Informatics 83

\bibitem[{Aeby et~al(2019)Aeby, Ushijima, Campbell, Jones, Williams, Meyer, H{\"a}se, and Paul}]{aeby2019pathogenesis}
Aeby GS, Ushijima B, Campbell JE, et~al (2019) Pathogenesis of a tissue loss disease affecting multiple species of corals along the florida reef tract. Frontiers in Marine Science 6

\bibitem[{Albawi et~al(2017)Albawi, Mohammed, and Al-Zawi}]{albawi2017understanding}
Albawi S, Mohammed TA, Al-Zawi S (2017) Understanding of a convolutional neural network. In: Proceedings of the International Conference on Engineering and Technology

\bibitem[{Alonso and Murillo(2018)}]{alonso2018semantic}
Alonso I, Murillo AC (2018) Semantic segmentation from sparse labeling using multi-level superpixels. In: Proceedings of the IEEE/RSJ International Conference on Intelligent Robots and Systems

\bibitem[{Alonso et~al(2017)Alonso, Cambra, Munoz, Treibitz, and Murillo}]{alonso2017coral}
Alonso I, Cambra A, Munoz A, et~al (2017) Coral-segmentation: Training dense labeling models with sparse ground truth. In: Proceedings of the IEEE International Conference on Computer Vision Workshops

\bibitem[{Alonso et~al(2019)Alonso, Yuval, Eyal, Treibitz, and Murillo}]{alonso2019coralseg}
Alonso I, Yuval M, Eyal G, et~al (2019) {CoralSeg}: Learning coral segmentation from sparse annotations. Journal of Field Robotics 36(8)

\bibitem[{Althaus et~al(2015)Althaus, Hill, Ferrari, Edwards, Przeslawski, Sch{\"o}nberg, Stuart-Smith, Barrett, Edgar, Colquhoun et~al}]{althaus2015standardised}
Althaus F, Hill N, Ferrari R, et~al (2015) A standardised vocabulary for identifying benthic biota and substrata from underwater imagery: the {CATAMI} classification scheme. PLOS One 10(10)

\bibitem[{Anthony et~al(2020)Anthony, Helmstedt, Bay, Fidelman, Hussey, Lundgren, Mead, McLeod, Mumby, Newlands et~al}]{anthony2020interventions}
Anthony KR, Helmstedt KJ, Bay LK, et~al (2020) Interventions to help coral reefs under global change—a complex decision challenge. PLOS One 15(8)

\bibitem[{Arain et~al(2019)Arain, McCool, Rigby, Cagara, and Dunbabin}]{arain2019improving}
Arain B, McCool C, Rigby P, et~al (2019) Improving underwater obstacle detection using semantic image segmentation. In: Proceedings of the IEEE International Conference on Robotics and Automation

\bibitem[{Assran et~al(2023)Assran, Duval, Misra, Bojanowski, Vincent, Rabbat, LeCun, and Ballas}]{assran2023self}
Assran M, Duval Q, Misra I, et~al (2023) Self-supervised learning from images with a joint-embedding predictive architecture. In: Proceedings of the IEEE/CVF Conference on Computer Vision and Pattern Recognition

\bibitem[{Bacheler and Shertzer(2015)}]{bacheler2015estimating}
Bacheler NM, Shertzer KW (2015) Estimating relative abundance and species richness from video surveys of reef fishes. Fishery Bulletin 113(1)

\bibitem[{Badrinarayanan et~al(2017)Badrinarayanan, Kendall, and Cipolla}]{badrinarayanan2017segnet}
Badrinarayanan V, Kendall A, Cipolla R (2017) Segnet: A deep convolutional encoder-decoder architecture for image segmentation. IEEE Transactions on Pattern Analysis and Machine Intelligence 39(12)

\bibitem[{Bae et~al(2020)Bae, Noh, and Kim}]{bae2020rethinking}
Bae W, Noh J, Kim G (2020) Rethinking class activation mapping for weakly supervised object localization. In: Proceedings of the European Conference on Computer Vision

\bibitem[{Balado et~al(2021)Balado, Olabarria, Mart{\'\i}nez-S{\'a}nchez, Rodr{\'\i}guez-P{\'e}rez, and Pedro}]{balado2021semantic}
Balado J, Olabarria C, Mart{\'\i}nez-S{\'a}nchez J, et~al (2021) Semantic segmentation of major macroalgae in coastal environments using high-resolution ground imagery and deep learning. International Journal of Remote Sensing 42(5)

\bibitem[{Bearman et~al(2016)Bearman, Russakovsky, Ferrari, and Fei-Fei}]{bearman2016s}
Bearman A, Russakovsky O, Ferrari V, et~al (2016) What’s the point: Semantic segmentation with point supervision. In: Proceedings of the European Conference on Computer Vision

\bibitem[{Beijbom et~al(2012)Beijbom, Edmunds, Kline, Mitchell, and Kriegman}]{beijbom2012automated}
Beijbom O, Edmunds PJ, Kline DI, et~al (2012) Automated annotation of coral reef survey images. In: Proceedings of the IEEE Conference on Computer Vision and Pattern Recognition

\bibitem[{Beijbom et~al(2015)Beijbom, Edmunds, Roelfsema, Smith, Kline, Neal, Dunlap, Moriarty, Fan, Tan et~al}]{beijbom2015towards}
Beijbom O, Edmunds PJ, Roelfsema C, et~al (2015) Towards automated annotation of benthic survey images: Variability of human experts and operational modes of automation. PLOS One 10(7)

\bibitem[{Beijbom et~al(2016)Beijbom, Treibitz, Kline, Eyal, Khen, Neal, Loya, Mitchell, and Kriegman}]{beijbom2016improving}
Beijbom O, Treibitz T, Kline DI, et~al (2016) Improving automated annotation of benthic survey images using wide-band fluorescence. Scientific Reports 6(1)

\bibitem[{Bewley et~al(2012)Bewley, Douillard, Nourani-Vatani, Friedman, Pizarro, and Williams}]{bewley2012automated}
Bewley M, Douillard B, Nourani-Vatani N, et~al (2012) Automated species detection: An experimental approach to kelp detection from sea-floor {AUV} images. In: Proceedings of the Australasian Conference on Robotics and Automation

\bibitem[{Bewley et~al(2015{\natexlab{a}})Bewley, Friedman, Ferrari, Hill, Hovey, Barrett, Marzinelli, Pizarro, Figueira, Meyer et~al}]{bewley2015australian}
Bewley M, Friedman A, Ferrari R, et~al (2015{\natexlab{a}}) Australian sea-floor survey data, with images and expert annotations. Scientific Data 2(1)

\bibitem[{Bewley et~al(2015{\natexlab{b}})Bewley, Nourani-Vatani, Rao, Douillard, Pizarro, and Williams}]{bewley2015hierarchical}
Bewley M, Nourani-Vatani N, Rao D, et~al (2015{\natexlab{b}}) Hierarchical classification in {AUV} imagery. In: Field and Service Robotics

\bibitem[{Bicknell et~al(2016)Bicknell, Godley, Sheehan, Votier, and Witt}]{bicknell2016camera}
Bicknell AW, Godley BJ, Sheehan EV, et~al (2016) Camera technology for monitoring marine biodiversity and human impact. Frontiers in Ecology and the Environment 14(8)

\bibitem[{Bijjahalli et~al(2023)Bijjahalli, Pizarro, and Williams}]{bijjahalli2023semi}
Bijjahalli S, Pizarro O, Williams SB (2023) A semi-supervised object detection algorithm for underwater imagery. arXiv preprint arXiv:230604834

\bibitem[{Boittiaux et~al(2023)Boittiaux, Dune, Ferrera, Arnaubec, Marxer, Matabos, Van~Audenhaege, and Hugel}]{boittiaux2023eiffel}
Boittiaux C, Dune C, Ferrera M, et~al (2023) {Eiffel Tower: A} deep-sea underwater dataset for long-term visual localization. The International Journal of Robotics Research 42(9)

\bibitem[{Bonin-Font et~al(2016)Bonin-Font, Campos, and Codina}]{bonin2016towards}
Bonin-Font F, Campos MM, Codina GO (2016) Towards visual detection, mapping and quantification of {Posidonia} oceanica using a lightweight {AUV}. International Federation of Automatic Control 49(23)

\bibitem[{Bonin-Font et~al(2017)Bonin-Font, Burguera, and Lisani}]{bonin-font2017visual}
Bonin-Font F, Burguera A, Lisani JL (2017) Visual discrimination and large area mapping of {P}osidonia oceanica using a lightweight {AUV}. IEEE Access 5

\bibitem[{Boom et~al(2012)Boom, Huang, He, and Fisher}]{boom2012supporting}
Boom BJ, Huang PX, He J, et~al (2012) Supporting ground-truth annotation of image datasets using clustering. In: Proceedings of the International Conference on Pattern Recognition

\bibitem[{Boom et~al(2014)Boom, He, Palazzo, Huang, Beyan, Chou, Lin, Spampinato, and Fisher}]{boom2014research}
Boom BJ, He J, Palazzo S, et~al (2014) A research tool for long-term and continuous analysis of fish assemblage in coral-reefs using underwater camera footage. Ecological Informatics 23

\bibitem[{Braylan et~al(2022)Braylan, Alonso, and Lease}]{braylan2022measuring}
Braylan A, Alonso O, Lease M (2022) Measuring annotator agreement generally across complex structured, multi-object, and free-text annotation tasks. In: Proceedings of the ACM Web Conference

\bibitem[{Brooks et~al(2010)Brooks, Rowat, Pierce, Jouannet, and Vely}]{brooks2010seeing}
Brooks K, Rowat D, Pierce SJ, et~al (2010) Seeing spots: photo-identification as a regional tool for whale shark identification. Western Indian Ocean Journal of Marine Science 9(2)

\bibitem[{Bryant et~al(2017)Bryant, Rodriguez-Ramirez, Phinn, Gonz{\'a}lez-Rivero, Brown, Neal, Hoegh-Guldberg, and Dove}]{bryant2017comparison}
Bryant D, Rodriguez-Ramirez A, Phinn S, et~al (2017) Comparison of two photographic methodologies for collecting and analyzing the condition of coral reef ecosystems. Ecosphere 8(10)

\bibitem[{Burguera(2020)}]{burguera2020segmentation}
Burguera A (2020) Segmentation through patch classification: A neural network approach to detect {Posidonia} oceanica in underwater images. Ecological Informatics 56

\bibitem[{Burguera et~al(2016)Burguera, Bonin-Font, Lisani, Petro, and Oliver}]{burgeura2016towards}
Burguera A, Bonin-Font F, Lisani JL, et~al (2016) Towards automatic visual seagrass detection in underwater areas of ecological interest. In: Proceedings of the IEEE International Conference on Emerging Technologies and Factory Automation

\bibitem[{Campagne et~al(2015)Campagne, Salles, Boissery, and Deter}]{campagne2015seagrass}
Campagne CS, Salles JM, Boissery P, et~al (2015) The seagrass {Posidonia} oceanica: ecosystem services identification and economic evaluation of goods and benefits. Marine Pollution Bulletin 97(1-2)

\bibitem[{Cao et~al(2015)Cao, Principe, Ouyang, Dalgleish, and Vuorenkoski}]{cao2015marine}
Cao Z, Principe JC, Ouyang B, et~al (2015) Marine animal classification using combined {CNN} and hand-designed image features. In: Proceedings of the OCEANS Conference

\bibitem[{Caron et~al(2021)Caron, Touvron, Misra, J{\'e}gou, Mairal, Bojanowski, and Joulin}]{caron2021emerging}
Caron M, Touvron H, Misra I, et~al (2021) Emerging properties in self-supervised vision transformers. In: Proceedings of the IEEE/CVF International Conference on Computer Vision

\bibitem[{Caron et~al(2024)Caron, Houlsby, and Schmid}]{caron2024location}
Caron M, Houlsby N, Schmid C (2024) Location-aware self-supervised transformers for semantic segmentation. In: Proceedings of the IEEE/CVF Winter Conference on Applications of Computer Vision

\bibitem[{Chen et~al(2021)Chen, Beijbom, Chan, Bouwmeester, and Kriegman}]{chen2021new}
Chen Q, Beijbom O, Chan S, et~al (2021) A new deep learning engine for {CoralNet}. In: Proceedings of the IEEE/CVF International Conference on Computer Vision

\bibitem[{Chen et~al(2020)Chen, Kornblith, Norouzi, and Hinton}]{chen2020simple}
Chen T, Kornblith S, Norouzi M, et~al (2020) A simple framework for contrastive learning of visual representations. In: Proceedings of the International Conference on Machine Learning

\bibitem[{Cheng et~al(2022)Cheng, Parkhi, and Kirillov}]{cheng2022pointly}
Cheng B, Parkhi O, Kirillov A (2022) Pointly-supervised instance segmentation. In: Proceedings of the IEEE/CVF Conference on Computer Vision and Pattern Recognition

\bibitem[{Combs et~al(2021)Combs, Studivan, Eckert, and Voss}]{combs2021quantifying}
Combs IR, Studivan MS, Eckert RJ, et~al (2021) Quantifying impacts of stony coral tissue loss disease on corals in southeast florida through surveys and {3D} photogrammetry. PLOS One 16(6)

\bibitem[{Cordts et~al(2016)Cordts, Omran, Ramos, Rehfeld, Enzweiler, Benenson, Franke, Roth, and Schiele}]{cordts2016cityscapes}
Cordts M, Omran M, Ramos S, et~al (2016) The cityscapes dataset for semantic urban scene understanding. In: Proceedings of the IEEE Conference on Computer Vision and Pattern Recognition

\bibitem[{Cornwall et~al(2021)Cornwall, Comeau, Kornder, Perry, van Hooidonk, DeCarlo, Pratchett, Anderson, Browne, Carpenter et~al}]{cornwall2021global}
Cornwall CE, Comeau S, Kornder NA, et~al (2021) Global declines in coral reef calcium carbonate production under ocean acidification and warming. Proceedings of the National Academy of Sciences 118(21)

\bibitem[{Dayoub et~al(2015)Dayoub, Dunbabin, and Corke}]{dayoub2015robotic}
Dayoub F, Dunbabin M, Corke P (2015) Robotic detection and tracking of crown-of-thorns starfish. In: Proceedings of the IEEE/RSJ International Conference on Intelligent Robots and Systems

\bibitem[{Deng et~al(2009)Deng, Dong, Socher, Li, Li, and Fei-Fei}]{deng2009imagenet}
Deng J, Dong W, Socher R, et~al (2009) Imagenet: A large-scale hierarchical image database. In: Proceedings of the IEEE Conference on Computer Vision and Pattern Recognition

\bibitem[{Denuelle and Dunbabin(2010)}]{denuelle2010kelp}
Denuelle A, Dunbabin M (2010) Kelp detection in highly dynamic environments using texture recognition. In: Proceedings of the Australasian Conference on Robotics and Automation

\bibitem[{Dietzel et~al(2021)Dietzel, Bode, Connolly, and Hughes}]{dietzel2021population}
Dietzel A, Bode M, Connolly SR, et~al (2021) The population sizes and global extinction risk of reef-building coral species at biogeographic scales. Nature Ecology and Evolution 5(5)

\bibitem[{Ditria et~al(2021)Ditria, Connolly, Jinks, and Lopez-Marcano}]{ditria2021annotated}
Ditria EM, Connolly RM, Jinks EL, et~al (2021) Annotated video footage for automated identification and counting of fish in unconstrained seagrass habitats. Frontiers in Marine Science 8

\bibitem[{Ditria et~al(2022)Ditria, Buelow, Gonzalez-Rivero, and Connolly}]{ditria2022artificial}
Ditria EM, Buelow CA, Gonzalez-Rivero M, et~al (2022) Artificial intelligence and automated monitoring for assisting conservation of marine ecosystems: A perspective. Frontiers in Marine Science 9

\bibitem[{Doropoulos et~al(2019)Doropoulos, Elzinga, ter Hofstede, van Koningsveld, and Babcock}]{doropoulos2019optimizing}
Doropoulos C, Elzinga J, ter Hofstede R, et~al (2019) Optimizing industrial-scale coral reef restoration: Comparing harvesting wild coral spawn slicks and transplanting gravid adult colonies. Restoration Ecology 27(4)

\bibitem[{Dosovitskiy et~al(2020)Dosovitskiy, Beyer, Kolesnikov, Weissenborn, Zhai, Unterthiner, Dehghani, Minderer, Heigold, Gelly et~al}]{dosovitskiy2020image}
Dosovitskiy A, Beyer L, Kolesnikov A, et~al (2020) An image is worth 16x16 words: Transformers for image recognition at scale. arXiv preprint arXiv:201011929

\bibitem[{Duarte et~al(2013)Duarte, Losada, Hendriks, Mazarrasa, and Marb{\`a}}]{duarte2013role}
Duarte CM, Losada IJ, Hendriks IE, et~al (2013) The role of coastal plant communities for climate change mitigation and adaptation. Nature Climate Change 3(11)

\bibitem[{Dunbabin et~al(2020)Dunbabin, Manley, and Harrison}]{dunbabin2020uncrewed}
Dunbabin M, Manley J, Harrison PL (2020) Uncrewed maritime systems for coral reef conservation. In: Proceedings of the OCEANS Conference

\bibitem[{Edwards et~al(2017)Edwards, Eynaud, Williams, Pedersen, Zgliczynski, Gleason, Smith, and Sandin}]{edwards2017large}
Edwards CB, Eynaud Y, Williams GJ, et~al (2017) Large-area imaging reveals biologically driven non-random spatial patterns of corals at a remote reef. Coral Reefs 36(4)

\bibitem[{English et~al(1997)English, Wilkinson, Baker et~al}]{english1997survey}
English S, Wilkinson C, Baker V, et~al (1997) Survey manual for tropical marine resources. Tech. rep., Townsville (Australia) AIMS

\bibitem[{Everingham et~al(2010)Everingham, Van~Gool, Williams, Winn, and Zisserman}]{everingham2010pascal}
Everingham M, Van~Gool L, Williams CK, et~al (2010) The {Pascal Visual Object Classes (VOC)} challenge. International Journal of Computer Vision 88

\bibitem[{Fabricius et~al(2010)Fabricius, Okaji, and De’Ath}]{fabricius2010three}
Fabricius K, Okaji K, De’Ath G (2010) Three lines of evidence to link outbreaks of the crown-of-thorns seastar {Acanthaster} planci to the release of larval food limitation. Coral Reefs 29

\bibitem[{Fan et~al(2022)Fan, Zhang, and Tan}]{fan2022pointly}
Fan J, Zhang Z, Tan T (2022) Pointly-supervised panoptic segmentation. In: Proceedings of the European Conference on Computer Vision

\bibitem[{Feng and Zhang(2023)}]{feng2023evolved}
Feng Z, Zhang S (2023) Evolved part masking for self-supervised learning. In: Proceedings of the IEEE/CVF Conference on Computer Vision and Pattern Recognition

\bibitem[{Ferguson and Korfmacher(1997)}]{ferguson1997remote}
Ferguson RL, Korfmacher K (1997) Remote sensing and {GIS} analysis of seagrass meadows in {North Carolina, USA}. Aquatic Botany 58(3-4)

\bibitem[{Ferrari et~al(2018)Ferrari, Marzinelli, Ayroza, Jordan, Figueira, Byrne, Malcolm, Williams, and Steinberg}]{ferrari2018large}
Ferrari R, Marzinelli EM, Ayroza CR, et~al (2018) Large-scale assessment of benthic communities across multiple marine protected areas using an autonomous underwater vehicle. PLOS One 13(3)

\bibitem[{Ferretti et~al(2017)Ferretti, Bibuli, Caccia, Chiarella, Odetti, Ranieri, Zereik, and Bruzzone}]{ferretti2017towards}
Ferretti R, Bibuli M, Caccia M, et~al (2017) Towards {Posidonia} meadows detection, mapping and automatic recognition using unmanned marine vehicles. International Federation of Automatic Control 50(1)

\bibitem[{Friedman(2013)}]{friedman2013automated}
Friedman AL (2013) Automated interpretation of benthic stereo imagery. PhD thesis, University of Sydney

\bibitem[{Fu et~al(2023)Fu, Chen, Huang, Cheng, Ding, and Ma}]{fu2023masnet}
Fu Z, Chen R, Huang Y, et~al (2023) Masnet: A robust deep marine animal segmentation network. IEEE Journal of Oceanic Engineering

\bibitem[{Furtado et~al(2023)Furtado, Vieira, Nascimento, Inagaki, Bleuel, Alves, Longo, and Oliveira}]{furtado2023deolhonoscorais}
Furtado DP, Vieira EA, Nascimento WF, et~al (2023) \# {DeOlhoNosCorais: A} polygonal annotated dataset to optimize coral monitoring. PeerJ 11

\bibitem[{Ganesan and Santhanam(2022)}]{ganesan2022novel}
Ganesan A, Santhanam SM (2022) A novel feature descriptor based coral image classification using extreme learning machine with ameliorated chimp optimization algorithm. Ecological Informatics 68

\bibitem[{Gao et~al(2022)Gao, Li, Kong, Yu, Guo, and Ren}]{gao2022algaenet}
Gao L, Li X, Kong F, et~al (2022) Algaenet: A deep-learning framework to detect floating green algae from optical and {SAR} imagery. IEEE Journal of Selected Topics in Applied Earth Observations and Remote Sensing 15

\bibitem[{Giles et~al(2023)Giles, Ren, Davies, Abrego, and Kelaher}]{giles2023combining}
Giles AB, Ren K, Davies JE, et~al (2023) Combining drones and deep learning to automate coral reef assessment with rgb imagery. Remote Sensing 15(9)

\bibitem[{G{\'o}mez-R{\'\i}os et~al(2019{\natexlab{a}})G{\'o}mez-R{\'\i}os, Tabik, Luengo, Shihavuddin, and Herrera}]{gomez2019coral}
G{\'o}mez-R{\'\i}os A, Tabik S, Luengo J, et~al (2019{\natexlab{a}}) Coral species identification with texture or structure images using a two-level classifier based on convolutional neural networks. Knowledge-Based Systems 184

\bibitem[{G{\'o}mez-R{\'\i}os et~al(2019{\natexlab{b}})G{\'o}mez-R{\'\i}os, Tabik, Luengo, Shihavuddin, Krawczyk, and Herrera}]{gomez2019towards}
G{\'o}mez-R{\'\i}os A, Tabik S, Luengo J, et~al (2019{\natexlab{b}}) Towards highly accurate coral texture images classification using deep convolutional neural networks and data augmentation. Expert Systems with Applications 118

\bibitem[{Gonzalez-Cid et~al(2017)Gonzalez-Cid, Burguera, Bonin-Font, and Matamoros}]{gonzalez2017machine}
Gonzalez-Cid Y, Burguera A, Bonin-Font F, et~al (2017) Machine learning and deep learning strategies to identify {Posidonia} meadows in underwater images. In: Proceedings of the OCEANS Conference

\bibitem[{Gonz{\'a}lez-Rivero et~al(2014)Gonz{\'a}lez-Rivero, Bongaerts, Beijbom, Pizarro, Friedman, Rodriguez-Ramirez, Upcroft, Laffoley, Kline, Bailhache et~al}]{gonzalez2014catlin}
Gonz{\'a}lez-Rivero M, Bongaerts P, Beijbom O, et~al (2014) The {Catlin} seaview survey--kilometre-scale seascape assessment, and monitoring of coral reef ecosystems. Aquatic Conservation: Marine and Freshwater Ecosystems 24

\bibitem[{Gonz{\'a}lez-Rivero et~al(2016)Gonz{\'a}lez-Rivero, Beijbom, Rodriguez-Ramirez, Holtrop, Gonz{\'a}lez-Marrero, Ganase, Roelfsema, Phinn, and Hoegh-Guldberg}]{gonzalez2016scaling}
Gonz{\'a}lez-Rivero M, Beijbom O, Rodriguez-Ramirez A, et~al (2016) Scaling up ecological measurements of coral reefs using semi-automated field image collection and analysis. Remote Sensing 8(1)

\bibitem[{Gonz{\'a}lez-Rivero et~al(2019)Gonz{\'a}lez-Rivero, Rodriguez-Ramirez, Beijbom, Dalton, Kennedy, Neal, Vercelloni, Bongaerts, Ganase, Bryant et~al}]{gonzalez2019seaview}
Gonz{\'a}lez-Rivero M, Rodriguez-Ramirez A, Beijbom O, et~al (2019) Seaview survey photo-quadrat and image classification dataset. \urlprefix\url{https://doi.org/10.14264/uql.2019.930}

\bibitem[{Gonz{\'a}lez-Rivero et~al(2020)Gonz{\'a}lez-Rivero, Beijbom, Rodriguez-Ramirez, Bryant, Ganase, Gonzalez-Marrero, Herrera-Reveles, Kennedy, Kim, Lopez-Marcano et~al}]{gonzalez2020monitoring}
Gonz{\'a}lez-Rivero M, Beijbom O, Rodriguez-Ramirez A, et~al (2020) Monitoring of coral reefs using artificial intelligence: A feasible and cost-effective approach. Remote Sensing 12(3)

\bibitem[{Gonz{\'a}lez-Sabbagh and Robles-Kelly(2023)}]{gonzalez2023survey}
Gonz{\'a}lez-Sabbagh SP, Robles-Kelly A (2023) A survey on underwater computer vision. ACM Computing Surveys 55

\bibitem[{Goodfellow et~al(2016)Goodfellow, Bengio, and Courville}]{goodfellow2016deep}
Goodfellow I, Bengio Y, Courville A (2016) Deep learning. MIT press

\bibitem[{Goreau et~al(1979)Goreau, Goreau, and Goreau}]{goreau1979corals}
Goreau TF, Goreau NI, Goreau TJ (1979) Corals and coral reefs. Scientific American 241(2)

\bibitem[{Gou et~al(2021)Gou, Yu, Maybank, and Tao}]{gou2021knowledge}
Gou J, Yu B, Maybank SJ, et~al (2021) Knowledge distillation: A survey. International Journal of Computer Vision 129(6)

\bibitem[{Gregorek et~al(2023)Gregorek, Tibebu, Caudet, Barrera, and Bachmayer}]{gregorek2023long}
Gregorek D, Tibebu A, Caudet E, et~al (2023) Long-endurance optical seafloor imaging using underwater gliders: Concept, development and initial trials. In: Proceedings of the IEEE/RSJ International Conference on Intelligent Robots and Systems

\bibitem[{Guo et~al(2018)Guo, Liu, Georgiou, and Lew}]{guo2018review}
Guo Y, Liu Y, Georgiou T, et~al (2018) A review of semantic segmentation using deep neural networks. International Journal of Multimedia Information Retrieval 7

\bibitem[{Gupta et~al(2015)Gupta, Arbel{\'a}ez, Girshick, and Malik}]{gupta2015indoor}
Gupta S, Arbel{\'a}ez P, Girshick R, et~al (2015) Indoor scene understanding with rgb-d images: Bottom-up segmentation, object detection and semantic segmentation. International Journal of Computer Vision 112

\bibitem[{Haixin et~al(2023)Haixin, Ziqiang, Zeyu, and Yeung}]{haixin2023marinedet}
Haixin L, Ziqiang Z, Zeyu M, et~al (2023) {MarineDet: T}owards open-marine object detection. arXiv preprint arXiv:231001931

\bibitem[{Han et~al(2020)Han, Yao, Zhu, and Wang}]{han2020marine}
Han F, Yao J, Zhu H, et~al (2020) Marine organism detection and classification from underwater vision based on the deep {CNN} method. Mathematical Problems in Engineering

\bibitem[{Hinton et~al(2015)Hinton, Vinyals, and Dean}]{hinton2015distilling}
Hinton G, Vinyals O, Dean J (2015) Distilling the knowledge in a neural network. arXiv preprint arXiv:150302531

\bibitem[{Hobley et~al(2021)Hobley, Arosio, French, Bremner, Dolphin, and Mackiewicz}]{hobley2021semi}
Hobley B, Arosio R, French G, et~al (2021) Semi-supervised segmentation for coastal monitoring seagrass using {RPA} imagery. Remote Sensing 13(9)

\bibitem[{Hua et~al(2021)Hua, Marcos, Mou, Zhu, and Tuia}]{hua2021semantic}
Hua Y, Marcos D, Mou L, et~al (2021) Semantic segmentation of remote sensing images with sparse annotations. IEEE Geoscience and Remote Sensing Letters 19

\bibitem[{Igbinenikaro et~al(2024)Igbinenikaro, Adekoya, and Etukudoh}]{igbinenikaro2024emerging}
Igbinenikaro OP, Adekoya OO, Etukudoh EA (2024) Emerging underwater survey technologies: A review and future outlook. Open Access Research Journal of Science and Technology 10(02)

\bibitem[{{Integrated Marine Observing System (IMOS)}(2011)}]{IMOS}
{Integrated Marine Observing System (IMOS)} (2011) {IMOS} - {AUV} {S}irius, campaign: Great {B}arrier {R}eef, {F}ebruary 2011. \urlprefix\url{https://catalogue-imos.aodn.org.au/geonetwork/srv/eng/catalog.search#/metadata/ae70eb18-b1f0-4012-8d62-b03daf99f7f2}

\bibitem[{Islam et~al(2020{\natexlab{a}})Islam, Edge, Xiao, Luo, Mehtaz, Morse, Enan, and Sattar}]{islam2020semantic}
Islam MJ, Edge C, Xiao Y, et~al (2020{\natexlab{a}}) Semantic segmentation of underwater imagery: Dataset and benchmark. In: Proceedings of the IEEE/RSJ International Conference on Intelligent Robots and Systems, IEEE

\bibitem[{Islam et~al(2020{\natexlab{b}})Islam, Luo, and Sattar}]{islam2020simultaneous}
Islam MJ, Luo P, Sattar J (2020{\natexlab{b}}) Simultaneous enhancement and super-resolution of underwater imagery for improved visual perception. In: Proceedings of the Robotics: Science and Systems

\bibitem[{Jackett et~al(2023{\natexlab{a}})Jackett, Althaus, Maguire, Farazi, Scoulding, Untiedt, Ryan, Shanks, Brodie, and Williams}]{jackett2023benthic}
Jackett C, Althaus F, Maguire K, et~al (2023{\natexlab{a}}) A benthic substrate classification method for seabed images using deep learning: Application to management of deep-sea coral reefs. Journal of Applied Ecology 60(7)

\bibitem[{Jackett et~al(2023{\natexlab{b}})Jackett, Maguire, Untiedt, Althaus, and Shanks}]{benthic2023csiro}
Jackett C, Maguire K, Untiedt C, et~al (2023{\natexlab{b}}) Benthic substrate classification {ML} training image snips. v2. \urlprefix\url{https://doi.org/10.25919/jfsp-zd42}

\bibitem[{Jell and Flood(1978)}]{jell1978guide}
Jell JS, Flood PG (1978) Guide to the geology of reefs of the {C}apricorn and {Bunker Groups, Great Barrier Reef Province, with special reference to Heron Reef}. Tech. rep., University of Queensland Press

\bibitem[{Jeon et~al(2021)Jeon, Kim, Park, Kwak, and Choi}]{jeon2021semantic}
Jeon Ei, Kim S, Park S, et~al (2021) Semantic segmentation of seagrass habitat from drone imagery based on deep learning: A comparative study. Ecological Informatics 66

\bibitem[{Jia et~al(2021)Jia, Yang, Xia, Chen, Parekh, Pham, Le, Sung, Li, and Duerig}]{jia2021scaling}
Jia C, Yang Y, Xia Y, et~al (2021) Scaling up visual and vision-language representation learning with noisy text supervision. In: Proceedings of the International Conference on Machine Learning

\bibitem[{Jin and Liang(2017)}]{jin2017deep}
Jin L, Liang H (2017) Deep learning for underwater image recognition in small sample size situations. In: Proceedings of the OCEANS Conference

\bibitem[{Jonker et~al(2020)Jonker, Bray, Johns, and Osborne}]{jonker2020surveys}
Jonker MJ, Bray PE, Johns KA, et~al (2020) Surveys of benthic reef communities using underwater digital photography and counts of juvenile corals: Long term monitoring of the {Great Barrier Reef} standard operational procedure number 10. Australian Institute of Marine Science

\bibitem[{Katija et~al(2022)Katija, Orenstein, Schlining, Lundsten, Barnard, Sainz, Boulais, Cromwell, Butler, Woodward et~al}]{katija2022fathomnet}
Katija K, Orenstein E, Schlining B, et~al (2022) Fathomnet: A global image database for enabling artificial intelligence in the ocean. Scientific Reports 12(1)

\bibitem[{Keetha et~al(2023)Keetha, Mishra, Karhade, Jatavallabhula, Scherer, Krishna, and Garg}]{keetha2023anyloc}
Keetha N, Mishra A, Karhade J, et~al (2023) {AnyLoc: T}owards universal visual place recognition. IEEE Robotics and Automation Letters

\bibitem[{Kemker et~al(2018)Kemker, Salvaggio, and Kanan}]{kemker2018algorithms}
Kemker R, Salvaggio C, Kanan C (2018) Algorithms for semantic segmentation of multispectral remote sensing imagery using deep learning. ISPRS Journal of Photogrammetry and Remote Sensing 145

\bibitem[{Khan et~al(2022)Khan, Naseer, Hayat, Zamir, Khan, and Shah}]{khan2022transformers}
Khan S, Naseer M, Hayat M, et~al (2022) Transformers in vision: A survey. ACM Computing Surveys 54(10s)

\bibitem[{King et~al(2019)King, M~Bhandarkar, and Hopkinson}]{king2019deep}
King A, M~Bhandarkar S, Hopkinson BM (2019) Deep learning for semantic segmentation of coral reef images using multi-view information. In: Proceedings of the IEEE Conference on Computer Vision and Pattern Recognition Workshops

\bibitem[{Kirillov et~al(2023)Kirillov, Mintun, Ravi, Mao, Rolland, Gustafson, Xiao, Whitehead, Berg, Lo et~al}]{kirillov2023segment}
Kirillov A, Mintun E, Ravi N, et~al (2023) Segment anything. In: Proceedings of the IEEE/CVF International Conference on Computer Vision

\bibitem[{Kohler and Gill(2006)}]{kohler2006coral}
Kohler KE, Gill SM (2006) {Coral Point Count with Excel extensions (CPCe): A Visual Basic }program for the determination of coral and substrate coverage using random point count methodology. Computers and Geosciences 32(9)

\bibitem[{Kwak et~al(2017)Kwak, Hong, Han et~al}]{kwak2017weakly}
Kwak S, Hong S, Han B, et~al (2017) Weakly supervised semantic segmentation using superpixel pooling network. In: Proceedings of the AAAI Conference on Artificial Intelligence

\bibitem[{Lateef and Ruichek(2019)}]{lateef2019survey}
Lateef F, Ruichek Y (2019) Survey on semantic segmentation using deep learning techniques. Neurocomputing 338

\bibitem[{Lavery et~al(2013)Lavery, Mateo, Serrano, and Rozaimi}]{lavery2013variability}
Lavery PS, Mateo M{\'A}, Serrano O, et~al (2013) Variability in the carbon storage of seagrass habitats and its implications for global estimates of blue carbon ecosystem service. PLOS One 8(9)

\bibitem[{Li et~al(2018)Li, Liu, Xu, and Qiu}]{li2018real}
Li B, Liu S, Xu W, et~al (2018) Real-time object detection and semantic segmentation for autonomous driving. In: Proceedings of the Multispectral Image Acquisition, Processing, and Analysis Conference

\bibitem[{Li et~al(2015{\natexlab{a}})Li, Eustice, and Johnson-Roberson}]{li2015high}
Li J, Eustice RM, Johnson-Roberson M (2015{\natexlab{a}}) High-level visual features for underwater place recognition. In: Proceedings of the IEEE International Conference on Robotics and Automation

\bibitem[{Li et~al(2020)Li, Rigall, Dong, and Chen}]{li2020mas3k}
Li L, Rigall E, Dong J, et~al (2020) {MAS3K: An} open dataset for marine animal segmentation. In: Proceedings of the International Symposium on Benchmarking, Measuring and Optimization

\bibitem[{Li et~al(2021)Li, Dong, Rigall, Zhou, Dong, and Chen}]{li2021marine}
Li L, Dong B, Rigall E, et~al (2021) Marine animal segmentation. IEEE Transactions on Circuits and Systems for Video Technology 32(4)

\bibitem[{Li et~al(2024)Li, Zhang, Gruen, and Li}]{li2024survey}
Li M, Zhang H, Gruen A, et~al (2024) A survey on underwater coral image segmentation based on deep learning. Geo-spatial Information Science

\bibitem[{Li et~al(2023)Li, Yuan, Wang, Zhu, Li, Liu, and Zhang}]{li2023point2mask}
Li W, Yuan Y, Wang S, et~al (2023) {Point2Mask: P}oint-supervised panoptic segmentation via optimal transport. In: Proceedings of the IEEE/CVF International Conference on Computer Vision

\bibitem[{Li et~al(2015{\natexlab{b}})Li, Shang, Qin, and Chen}]{li2015fast}
Li X, Shang M, Qin H, et~al (2015{\natexlab{b}}) Fast accurate fish detection and recognition of underwater images with {Fast R-CNN}. In: Proceedings of the OCEANS Conference

\bibitem[{Li et~al(2022)Li, Liu, Kusy, Marchant, Do, Merz, Crosswell, Steven, Tychsen-Smith, Ahmedt-Aristizabal et~al}]{li2022real}
Li Y, Liu J, Kusy B, et~al (2022) A real-time edge-{AI} system for reef surveys. In: Proceedings of the International Conference on Mobile Computing and Networking

\bibitem[{Liang et~al(2020)Liang, Liu, He, and Li}]{liang2020weakly}
Liang B, Liu Y, He L, et~al (2020) Weakly supervised semantic segmentation based on deep learning. In: Proceedings of the International Conference on Modelling, Identification and Control

\bibitem[{Lin et~al(2014)Lin, Maire, Belongie, Hays, Perona, Ramanan, Doll{\'a}r, and Zitnick}]{lin2014microsoft}
Lin TY, Maire M, Belongie S, et~al (2014) Microsoft {COCO: Common Objects in Context}. In: Proceedings of the European Conference on Computer Vision

\bibitem[{Lindenmayer and Likens(2009)}]{lindenmayer2009adaptive}
Lindenmayer DB, Likens GE (2009) Adaptive monitoring: A new paradigm for long-term research and monitoring. Trends in Ecology and Evolution 24(9)

\bibitem[{Liu et~al(2021{\natexlab{a}})Liu, Li, Wang, Zhu, Wang, Fan, and Wang}]{liu2021dataset}
Liu C, Li H, Wang S, et~al (2021{\natexlab{a}}) A dataset and benchmark of underwater object detection for robot picking. In: Proceedings of the IEEE International Conference on Multimedia and Expo Workshops

\bibitem[{Liu et~al(2021{\natexlab{b}})Liu, Kusy, Marchant, Do, Merz, Crosswell, Steven, Heaney, von Richter, Tychsen-Smith et~al}]{liu2021csiro}
Liu J, Kusy B, Marchant R, et~al (2021{\natexlab{b}}) The {CSIRO} {Crown-of-Thorn Starfish} detection dataset. arXiv preprint arXiv:211114311

\bibitem[{Liu et~al(2022)Liu, Li, Crosswell, and Do}]{liu2022underwater}
Liu J, Li Y, Crosswell J, et~al (2022) {Underwater Crown-of-Thorn Starfish (COTS)} imagery sample dataset, v2. \urlprefix\url{https://doi.org/10.25919/nf6m-2k87}

\bibitem[{Loya(2004)}]{loya2004coral}
Loya Y (2004) The coral reefs of {Eilat}—past, present and future: three decades of coral community structure studies. In: Coral Health and Disease. Springer

\bibitem[{Lumini et~al(2020)Lumini, Nanni, and Maguolo}]{lumini2020deep}
Lumini A, Nanni L, Maguolo G (2020) Deep learning for plankton and coral classification. Applied Computing and Informatics 19(3-4)

\bibitem[{Lutzenkirchen et~al(2024)Lutzenkirchen, Duce, and Bellwood}]{lutzenkirchen2024exploring}
Lutzenkirchen LL, Duce SJ, Bellwood DR (2024) Exploring benthic habitat assessments on coral reefs: a comparison of direct field measurements versus remote sensing. Coral Reefs

\bibitem[{Macreadie et~al(2014)Macreadie, Baird, Trevathan-Tackett, Larkum, and Ralph}]{macreadie2014quantifying}
Macreadie P, Baird M, Trevathan-Tackett S, et~al (2014) Quantifying and modelling the carbon sequestration capacity of seagrass meadows--a critical assessment. Marine Pollution Bulletin 83(2)

\bibitem[{Mahmood et~al(2016{\natexlab{a}})Mahmood, Bennamoun, An, Sohel, Boussaid, Hovey, Kendrick, and Fisher}]{mahmood2016automatic}
Mahmood A, Bennamoun M, An S, et~al (2016{\natexlab{a}}) Automatic annotation of coral reefs using deep learning. In: Proceedings of the OCEANS Conference

\bibitem[{Mahmood et~al(2016{\natexlab{b}})Mahmood, Bennamoun, An, Sohel, Boussaid, Hovey, Kendrick, and Fisher}]{mahmood2016coral}
Mahmood A, Bennamoun M, An S, et~al (2016{\natexlab{b}}) Coral classification with hybrid feature representations. In: Proceedings of the IEEE International Conference on Image Processing

\bibitem[{Mahmood et~al(2017)Mahmood, Bennamoun, An, Sohel, Boussaid, Hovey, Kendrick, and Fisher}]{mahmood2017deep}
Mahmood A, Bennamoun M, An S, et~al (2017) Deep learning for coral classification. In: Handbook of Neural Computation. Elsevier

\bibitem[{Mahmood et~al(2020{\natexlab{a}})Mahmood, Bennamoun, An, Sohel, and Boussaid}]{mahmood2020resfeats}
Mahmood A, Bennamoun M, An S, et~al (2020{\natexlab{a}}) {ResFeats: R}esidual network based features for underwater image classification. Image and Vision Computing 93

\bibitem[{Mahmood et~al(2020{\natexlab{b}})Mahmood, Ospina, Bennamoun, An, Sohel, Boussaid, Hovey, Fisher, and Kendrick}]{mahmood2020automatic}
Mahmood A, Ospina AG, Bennamoun M, et~al (2020{\natexlab{b}}) Automatic hierarchical classification of kelps using deep residual features. Sensors 20(2)

\bibitem[{Marre et~al(2020)Marre, Deter, Holon, Boissery, and Luque}]{marre2020fine}
Marre G, Deter J, Holon F, et~al (2020) Fine-scale automatic mapping of living {Posidonia} oceanica seagrass beds with underwater photogrammetry. Marine Ecology Progress Series 643

\bibitem[{Martin-Abadal et~al(2018)Martin-Abadal, Guerrero-Font, Bonin-Font, and Gonzalez-Cid}]{martin-abadal2018deep}
Martin-Abadal M, Guerrero-Font E, Bonin-Font F, et~al (2018) Deep semantic segmentation in an {AUV} for online {Posidonia} oceanica meadows identification. IEEE Access 6

\bibitem[{Mary and Dejey(2018)}]{mary2018classification}
Mary NAB, Dejey D (2018) Classification of coral reef submarine images and videos using a novel {Z} with tilted {Z} local binary pattern. Wireless Personal Communications 98(3)

\bibitem[{Massot-Campos et~al(2013)Massot-Campos, Oliver-Codina, Ruano-Amengual, and Mir{\'o}-Juli{\'a}}]{massot-campos}
Massot-Campos M, Oliver-Codina G, Ruano-Amengual L, et~al (2013) Texture analysis of seabed images: Quantifying the presence of {Posidonia} oceanica at {Palma Bay}. In: Proceedings of the OCEANS Conference

\bibitem[{Massot-Campos et~al(2023)Massot-Campos, Bonin-Font, Guerrero-Font, Martorell-Torres, Abadal, Muntaner-Gonzalez, Nordfeldt-Fiol, Oliver-Codina, Cappelletto, and Thornton}]{massot2023assessing}
Massot-Campos M, Bonin-Font F, Guerrero-Font E, et~al (2023) Assessing benthic marine habitats colonized with {Posidonia} oceanica using autonomous marine robots and deep learning: A {Eurofleets} campaign. Estuarine, Coastal and Shelf Science 291

\bibitem[{Mazarrasa et~al(2018)Mazarrasa, Samper-Villarreal, Serrano, Lavery, Lovelock, Marb{\`a}, Duarte, and Cort{\'e}s}]{mazarrasa2018habitat}
Mazarrasa I, Samper-Villarreal J, Serrano O, et~al (2018) Habitat characteristics provide insights of carbon storage in seagrass meadows. Marine Pollution Bulletin 134

\bibitem[{McEver and Manjunath(2020)}]{mcever2020pcams}
McEver RA, Manjunath B (2020) {PCAMs}: Weakly supervised semantic segmentation using point supervision. arXiv preprint arXiv:200705615

\bibitem[{McKenzie et~al(2022)McKenzie, Langlois, and Roelfsema}]{mckenzie2022improving}
McKenzie LJ, Langlois LA, Roelfsema CM (2022) Improving approaches to mapping seagrass within the {Great Barrier Reef}: From field to spaceborne {Earth} observation. Remote Sensing 14(11)

\bibitem[{McLeod et~al(2022)McLeod, Hein, Babcock, Bay, Bourne, Cook, Doropoulos, Gibbs, Harrison, Lockie et~al}]{mcleod2022coral}
McLeod IM, Hein MY, Babcock R, et~al (2022) Coral restoration and adaptation in {Australia}: The first five years. PLOS One 17(11)

\bibitem[{Mehrubeoglu et~al(2021)Mehrubeoglu, Vargas, Huang, and Cammarata}]{mehrubeoglu2021segmentation}
Mehrubeoglu M, Vargas I, Huang C, et~al (2021) Segmentation of seagrass blade images using deep learning. In: Real-time Image Processing and Deep Learning

\bibitem[{Milioto et~al(2018)Milioto, Lottes, and Stachniss}]{milioto2018real}
Milioto A, Lottes P, Stachniss C (2018) Real-time semantic segmentation of crop and weed for precision agriculture robots leveraging background knowledge in {CNN}s. In: Proceedings of the IEEE International Conference on Robotics and Automation

\bibitem[{Miller et~al(2018)Miller, Jonker, and Coleman}]{miller2018crown}
Miller IR, Jonker MJ, Coleman G (2018) Crown-of-thorns starfish and coral surveys using the manta tow technique: Long-term monitoring of the {Great Barrier Reef} standard operational procedure number 9. Australian Institute of Marine Science

\bibitem[{Mittal et~al(2022)Mittal, Srivastava, and Jayanth}]{mittal2022survey}
Mittal S, Srivastava S, Jayanth JP (2022) A survey of deep learning techniques for underwater image classification. IEEE Transactions on Neural Networks and Learning Systems 34

\bibitem[{Modasshir and Rekleitis(2020)}]{modasshir2020augmenting}
Modasshir M, Rekleitis I (2020) Augmenting coral reef monitoring with an enhanced detection system. In: Proceedings of the IEEE International Conference on Robotics and Automation

\bibitem[{Modasshir et~al(2018{\natexlab{a}})Modasshir, Li, and Rekleitis}]{modasshir2018mdnet}
Modasshir M, Li AQ, Rekleitis I (2018{\natexlab{a}}) {MDNet}: Multi-patch dense network for coral classification. In: Proceedings of the OCEANS Conference

\bibitem[{Modasshir et~al(2018{\natexlab{b}})Modasshir, Rahman, Youngquist, and Rekleitis}]{modasshir2018coral}
Modasshir M, Rahman S, Youngquist O, et~al (2018{\natexlab{b}}) Coral identification and counting with an autonomous underwater vehicle. In: Proceedings of the IEEE International Conference on Robotics and Biomimetics

\bibitem[{Mohamed et~al(2020)Mohamed, Nadaoka, and Nakamura}]{mohamed2020semiautomated}
Mohamed H, Nadaoka K, Nakamura T (2020) Semiautomated mapping of benthic habitats and seagrass species using a convolutional neural network framework in shallow water environments. Remote Sensing 12(23)

\bibitem[{Mohamed et~al(2022)Mohamed, Nadaoka, and Nakamura}]{mohamed2022automatic}
Mohamed H, Nadaoka K, Nakamura T (2022) Automatic semantic segmentation of benthic habitats using images from towed underwater camera in a complex shallow water environment. Remote Sensing 14(8)

\bibitem[{Moniruzzaman et~al(2019{\natexlab{a}})Moniruzzaman, Islam, and Lavery}]{ECUHO}
Moniruzzaman M, Islam SM, Lavery P (2019{\natexlab{a}}) {ECUHO1 (Edith Cowan University Halophila Ovalis 1)}. \urlprefix\url{https://doi.org/10.25958/5dd7566dfaabb}

\bibitem[{Moniruzzaman et~al(2019{\natexlab{b}})Moniruzzaman, Islam, Lavery, and Bennamoun}]{moniruzzaman2019faster}
Moniruzzaman M, Islam SMS, Lavery P, et~al (2019{\natexlab{b}}) Faster {R-CNN} based deep learning for seagrass detection from underwater digital images. In: Proceedings of the Digital Image Computing: Techniques and Applications

\bibitem[{Monk et~al(2018)Monk, Barrett, Bridge, Carroll, Friedman, Jordan, Kendrick, Lucieer et~al}]{monk2018marine}
Monk J, Barrett N, Bridge T, et~al (2018) Marine sampling field manual for {AUV's (Autonomous Underwater Vehicles)}. Tech. rep., NESP Marine Biodiversity Hub

\bibitem[{Moskvyak et~al(2021)Moskvyak, Maire, Dayoub, Armstrong, and Baktashmotlagh}]{moskvyak2021robust}
Moskvyak O, Maire F, Dayoub F, et~al (2021) Robust re-identification of manta rays from natural markings by learning pose invariant embeddings. In: Proceedings of the Digital Image Computing: Techniques and Applications

\bibitem[{Mou et~al(2022)Mou, Tsai, and Dunbabin}]{mou2022reconfigurable}
Mou S, Tsai D, Dunbabin M (2022) Reconfigurable robots for scaling reef restoration. arXiv preprint arXiv:220504612

\bibitem[{Mtwana~Nordlund et~al(2016)Mtwana~Nordlund, Koch, Barbier, and Creed}]{mtwana2016seagrass}
Mtwana~Nordlund L, Koch EW, Barbier EB, et~al (2016) Seagrass ecosystem services and their variability across genera and geographical regions. PLOS One 11(10)

\bibitem[{Muntaner-Gonzalez et~al(2023)Muntaner-Gonzalez, Martin-Abadal, and Gonzalez-Cid}]{muntaner2023deep}
Muntaner-Gonzalez C, Martin-Abadal M, Gonzalez-Cid Y (2023) A deep learning approach to estimate {Halimeda incrassata} invasive stage in the mediterranean sea. Journal of Marine Science and Engineering 12(1)

\bibitem[{Murphy and Jenkins(2010)}]{murphy2010observational}
Murphy HM, Jenkins GP (2010) Observational methods used in marine spatial monitoring of fishes and associated habitats: A review. Marine and Freshwater Research 61(2)

\bibitem[{Noman et~al(2021{\natexlab{a}})Noman, Islam, Abu-Khalaf, and Lavery}]{noman2021multi}
Noman MK, Islam SMS, Abu-Khalaf J, et~al (2021{\natexlab{a}}) Multi-species seagrass detection using semi-supervised learning. In: Proceedings of the International Conference on Image and Vision Computing New Zealand

\bibitem[{Noman et~al(2021{\natexlab{b}})Noman, Islam, Abu-Khalaf, and Lavery}]{noman2021seagrass}
Noman MK, Islam SMS, Abu-Khalaf J, et~al (2021{\natexlab{b}}) Seagrass detection from underwater digital images using {Faster R-CNN} with {NASNet}. In: Proceedings of the Digital Image Computing: Techniques and Applications

\bibitem[{Noman et~al(2023)Noman, Islam, Abu-Khalaf, Jalali, and Lavery}]{noman2023improving}
Noman MK, Islam SMS, Abu-Khalaf J, et~al (2023) Improving accuracy and efficiency in seagrass detection using state-of-the-art {AI} techniques. Ecological Informatics 76

\bibitem[{Obikane and Aoki(2019)}]{obikane2019weakly}
Obikane S, Aoki Y (2019) Weakly supervised domain adaptation with point supervision in histopathological image segmentation. In: Proceedings of the Asian Conference on Pattern Recognition

\bibitem[{Ondiviela et~al(2014)Ondiviela, Losada, Lara, Maza, Galv{\'a}n, Bouma, and van Belzen}]{ondiviela2014role}
Ondiviela B, Losada IJ, Lara JL, et~al (2014) The role of seagrasses in coastal protection in a changing climate. Coastal Engineering 87

\bibitem[{Oquab et~al(2023)Oquab, Darcet, Moutakanni, Vo, Szafraniec, Khalidov, Fernandez, Haziza, Massa, El-Nouby et~al}]{oquab2023dinov2}
Oquab M, Darcet T, Moutakanni T, et~al (2023) {DINOv2: L}earning robust visual features without supervision. arXiv preprint arXiv:230407193

\bibitem[{Ozaeta et~al(2023)Ozaeta, Fajardo, Brazas, and Cantal}]{ozaeta2023seagrass}
Ozaeta MAA, Fajardo AC, Brazas FP, et~al (2023) Seagrass classification using differentiable architecture search. In: Proceedings of the International Joint Conference on Computer Science and Software Engineering

\bibitem[{Pamungkas et~al(2021)Pamungkas, Jaya, and Iqbal}]{pamungkas2021segmentation}
Pamungkas S, Jaya I, Iqbal M (2021) Segmentation of {Enhalus} acoroides seagrass from underwater images using the {Mask R-CNN} method. In: Proceedings of the IOP Conference Series: Earth and Environmental Science

\bibitem[{Papke et~al(2024)Papke, Carreiro, Dennison, Deutsch, Isma, Meiling, Rossin, Baker, Brandt, Garg et~al}]{papke2024stony}
Papke E, Carreiro A, Dennison C, et~al (2024) Stony coral tissue loss disease: a review of emergence, impacts, etiology, diagnostics, and intervention. Frontiers in Marine Science 10

\bibitem[{Paul et~al(2020)Paul, Rani, and Manopriya}]{paul2020gradient}
Paul MA, Rani PAJ, Manopriya JL (2020) Gradient based aura feature extraction for coral reef classification. Wireless Personal Communications 114(1)

\bibitem[{Paul et~al(2023)Paul, Kumar, Sagar, and Sreeji}]{paul2023lwds}
Paul MA, Kumar KS, Sagar S, et~al (2023) {LWDS}: lightweight {DeepSeagrass} technique for classifying seagrass from underwater images. Environmental Monitoring and Assessment 195(5)

\bibitem[{Pavoni et~al(2019)Pavoni, Corsini, Callieri, Palma, and Scopigno}]{pavoni2019semantic}
Pavoni G, Corsini M, Callieri M, et~al (2019) Semantic segmentation of benthic communities from ortho-mosaic maps. International Archives of the Photogrammetry, Remote Sensing and Spatial Information Sciences

\bibitem[{Pavoni et~al(2022)Pavoni, Corsini, Ponchio, Muntoni, Edwards, Pedersen, Sandin, and Cignoni}]{pavoni2022taglab}
Pavoni G, Corsini M, Ponchio F, et~al (2022) {TagLab: AI-}assisted annotation for the fast and accurate semantic segmentation of coral reef orthoimages. Journal of Field Robotics 39(3)

\bibitem[{Pedersen et~al(2022)Pedersen, Haurum, Moeslund, and Nyegaard}]{pedersen2022re}
Pedersen M, Haurum JB, Moeslund TB, et~al (2022) Re-identification of giant sunfish using keypoint matching. In: Proceedings of the Northern Lights Deep Learning Workshop

\bibitem[{Pham et~al(2019)Pham, Xia, Ha, Bui, Le, and Takeuchi}]{pham2019review}
Pham TD, Xia J, Ha NT, et~al (2019) A review of remote sensing approaches for monitoring blue carbon ecosystems: Mangroves, seagrasses and salt marshes during 2010--2018. Sensors 19(8)

\bibitem[{Phinn et~al(2008)Phinn, Roelfsema, Dekker, Brando, and Anstee}]{phinn2008mapping}
Phinn S, Roelfsema C, Dekker A, et~al (2008) Mapping seagrass species, cover and biomass in shallow waters: An assessment of satellite multi-spectral and airborne hyper-spectral imaging systems in {Moreton Bay (Australia}). Remote Sensing of Environment 112(8)

\bibitem[{Picek et~al(2020)Picek, {\v{R}}{\'\i}ha, and Zita}]{picek2020coral}
Picek L, {\v{R}}{\'\i}ha A, Zita A (2020) Coral reef annotation, localisation and pixel-wise classification using {Mask R-CNN} and bag of tricks. In: Proceedings of the CEUR Workshops

\bibitem[{Pierce et~al(2020)Pierce, Rzhanov, Lowell, and Dijkstra}]{pierce2020reducing}
Pierce JP, Rzhanov Y, Lowell K, et~al (2020) Reducing annotation times: Semantic segmentation of coral reef survey images. In: Proceedings of the OCEANS Conference

\bibitem[{Qin et~al(2016)Qin, Li, Liang, Peng, and Zhang}]{qin2016deepfish}
Qin H, Li X, Liang J, et~al (2016) {DeepFish: A}ccurate underwater live fish recognition with a deep architecture. Neurocomputing 187

\bibitem[{Qu et~al(2020)Qu, Yi, Huang, Wu, and Metaxas}]{qu2020nuclei}
Qu H, Yi J, Huang Q, et~al (2020) Nuclei segmentation using mixed points and masks selected from uncertainty. In: Proceedings of the IEEE 17th International Symposium on Biomedical Imaging

\bibitem[{Radford et~al(2021)Radford, Kim, Hallacy, Ramesh, Goh, Agarwal, Sastry, Askell, Mishkin, Clark et~al}]{radford2021learning}
Radford A, Kim JW, Hallacy C, et~al (2021) Learning transferable visual models from natural language supervision. In: International Conference on Machine Learning

\bibitem[{Raine et~al(2020)Raine, Marchant, Moghadam, Maire, Kettle, and Kusy}]{raine2020multi}
Raine S, Marchant R, Moghadam P, et~al (2020) Multi-species seagrass detection and classification from underwater images. In: Proceedings of the Digital Image Computing: Techniques and Applications

\bibitem[{Raine et~al(2021)Raine, Marchant, Moghadam, Maire, Kettle, and Kusy}]{raine2021deepseagrass}
Raine S, Marchant R, Moghadam P, et~al (2021) {DeepSeagrass} dataset. arXiv preprint arXiv:210305226

\bibitem[{Raine et~al(2022)Raine, Marchant, Kusy, Maire, and Fischer}]{raine2022point}
Raine S, Marchant R, Kusy B, et~al (2022) Point label aware superpixels for multi-species segmentation of underwater imagery. IEEE Robotics and Automation Letters 7(3):8291--8298

\bibitem[{Raine et~al(2024{\natexlab{a}})Raine, Marchant, Kusy, Maire, and Fischer}]{raine2024image}
Raine S, Marchant R, Kusy B, et~al (2024{\natexlab{a}}) Image labels are all you need for coarse seagrass segmentation. In: Proceedings of the IEEE/CVF Winter Conference on Applications of Computer Vision, pp 5943--5952

\bibitem[{Raine et~al(2024{\natexlab{b}})Raine, Marchant, Kusy, Maire, Sunderhauf, and Fischer}]{raine2024human}
Raine S, Marchant R, Kusy B, et~al (2024{\natexlab{b}}) Human-in-the-loop segmentation of multi-species coral imagery. In: Proceedings of the IEEE/CVF Conference on Computer Vision and Pattern Recognition Workshops, pp 2723--2732

\bibitem[{Rao et~al(2017)Rao, De~Deuge, Nourani-Vatani, Williams, and Pizarro}]{rao2017multimodal}
Rao D, De~Deuge M, Nourani-Vatani N, et~al (2017) Multimodal learning and inference from visual and remotely sensed data. International Journal of Robotics Research 36(1)

\bibitem[{Raphael et~al(2020{\natexlab{a}})Raphael, Dubinsky, Iluz, Benichou, and Netanyahu}]{raphael2020deep}
Raphael A, Dubinsky Z, Iluz D, et~al (2020{\natexlab{a}}) Deep neural network recognition of shallow water corals in the {Gulf of Eilat (Aqaba)}. Scientific Reports 10(1)

\bibitem[{Raphael et~al(2020{\natexlab{b}})Raphael, Dubinsky, Iluz, and Netanyahu}]{raphael2020neural}
Raphael A, Dubinsky Z, Iluz D, et~al (2020{\natexlab{b}}) Neural network recognition of marine benthos and corals. Diversity 12(1)

\bibitem[{Reguero et~al(2018)Reguero, Beck, Agostini, Kramer, and Hancock}]{reguero2018coral}
Reguero BG, Beck MW, Agostini VN, et~al (2018) Coral reefs for coastal protection: A new methodological approach and engineering case study in {Grenada}. Journal of Environmental Management 210

\bibitem[{Remmers et~al(2023)Remmers, Grech, Roelfsema, Gordon, Lechene, and Ferrari}]{remmers2023close}
Remmers T, Grech A, Roelfsema C, et~al (2023) Close-range underwater photogrammetry for coral reef ecology: A systematic literature review. Coral Reefs

\bibitem[{Rende et~al(2020)Rende, Bosman, Di~Mento, Bruno, Lagudi, Irving, Dattola, Giambattista, Lanera, Proietti et~al}]{rende2020ultra}
Rende SF, Bosman A, Di~Mento R, et~al (2020) Ultra-high-resolution mapping of {Posidonia oceanica (L.)} delile meadows through acoustic, optical data and object-based image classification. Journal of Marine Science and Engineering 8(9)

\bibitem[{Reus et~al(2018)Reus, M{\"o}ller, J{\"a}ger, Schultz, Kruschel, Hasenauer, Wolff, and Fricke-Neuderth}]{reus2018looking}
Reus G, M{\"o}ller T, J{\"a}ger J, et~al (2018) Looking for seagrass: Deep learning for visual coverage estimation. In: Proceedings of the OCEANS Conference

\bibitem[{Roelfsema and Phinn(2010)}]{roelfsema2010integrating}
Roelfsema CM, Phinn SR (2010) Integrating field data with high spatial resolution multispectral satellite imagery for calibration and validation of coral reef benthic community maps. Journal of Applied Remote Sensing 4(1)

\bibitem[{Roelfsema et~al(2015)Roelfsema, Kovacs, and Phinn}]{roelfsema}
Roelfsema CM, Kovacs E, Phinn SR (2015) Benthic and substrate cover data derived from a time series of photo-transect surveys for the {Eastern Banks, Moreton Bay Australia}, 2004-2014. \urlprefix\url{https://doi.org/10.1594/PANGAEA.846147}

\bibitem[{Rom{\'a}n et~al(2021)Rom{\'a}n, Tovar-S{\'a}nchez, Oliv{\'e}, and Navarro}]{roman2021using}
Rom{\'a}n A, Tovar-S{\'a}nchez A, Oliv{\'e} I, et~al (2021) Using a {UAV}-mounted multispectral camera for the monitoring of marine macrophytes. Frontiers in Marine Science

\bibitem[{Rong et~al(2023)Rong, Tu, Wang, and Li}]{rong2023boundary}
Rong S, Tu B, Wang Z, et~al (2023) Boundary-enhanced co-training for weakly supervised semantic segmentation. In: Proceedings of the IEEE/CVF Conference on Computer Vision and Pattern Recognition

\bibitem[{Roth et~al(2019)Roth, Zhang, Yang, Milletari, Xu, Wang, and Xu}]{roth2019weakly}
Roth H, Zhang L, Yang D, et~al (2019) Weakly supervised segmentation from extreme points. In: Proceedings of the International Conference on Medical Image Computing and Computer Assisted Intervention Workshops

\bibitem[{Runyan et~al(2022)Runyan, Petrovic, Edwards, Pedersen, Alcantar, Kuester, and Sandin}]{runyan2022automated}
Runyan H, Petrovic V, Edwards CB, et~al (2022) Automated {2D, 2.5 D, and 3D} segmentation of coral reef pointclouds and orthoprojections. Frontiers in Robotics and AI 9

\bibitem[{Ruscio et~al(2023)Ruscio, Costanzi, Gracias, Quintana, and Garcia}]{ruscio2023autonomous}
Ruscio F, Costanzi R, Gracias N, et~al (2023) Autonomous boundary inspection of {P}osidonia oceanica meadows using an underwater robot. Ocean Engineering 283

\bibitem[{Saleh(2023)}]{saleh2023novel}
Saleh A (2023) Novel deep learning architectures for marine and aquaculture applications. PhD thesis, James Cook University

\bibitem[{Sauder et~al(2023)Sauder, Banc-Prandi, Meibom, and Tuia}]{sauder2023scalable}
Sauder J, Banc-Prandi G, Meibom A, et~al (2023) Scalable semantic {3D} mapping of coral reefs with deep learning. Methods in Ecology and Evolution

\bibitem[{Seiler et~al(2012)Seiler, Friedman, Steinberg, Barrett, Williams, and Holbrook}]{seiler2012image}
Seiler J, Friedman A, Steinberg D, et~al (2012) Image-based continental shelf habitat mapping using novel automated data extraction techniques. Continental Shelf Research 45

\bibitem[{Sengupta et~al(2020)Sengupta, Ersb{\o}ll, and Stockmarr}]{sengupta2020seagrassdetect}
Sengupta S, Ersb{\o}ll BK, Stockmarr A (2020) Seagrassdetect: A novel method for the detection of seagrass from unlabelled underwater videos. Ecological Informatics 57

\bibitem[{Shakoor and Boostani(2018)}]{shakoor2018novel}
Shakoor MH, Boostani R (2018) A novel advanced local binary pattern for image-based coral reef classification. Multimedia Tools and Applications 77(2)

\bibitem[{Shaver et~al(2022)Shaver, McLeod, Hein, Palumbi, Quigley, Vardi, Mumby, Smith, Montoya-Maya, Muller et~al}]{shaver2022roadmap}
Shaver EC, McLeod E, Hein MY, et~al (2022) A roadmap to integrating resilience into the practice of coral reef restoration. Global Change Biology 28(16)

\bibitem[{Shen et~al(2021)Shen, Cao, Chen, Lian, Zhang, Su, Wu, Huang, and Ji}]{shen2021toward}
Shen Y, Cao L, Chen Z, et~al (2021) Toward joint thing-and-stuff mining for weakly supervised panoptic segmentation. In: Proceedings of the IEEE/CVF Conference on Computer Vision and Pattern Recognition

\bibitem[{Shields et~al(2020)Shields, Pizarro, and Williams}]{shields2020towards}
Shields J, Pizarro O, Williams SB (2020) Towards adaptive benthic habitat mapping. In: Proceedings of the IEEE International Conference on Robotics and Automation

\bibitem[{Shihavuddin(2017)}]{shihavuddin2017dataset}
Shihavuddin A (2017) Coral reef dataset, v2. \urlprefix\url{http://dx.doi.org/10.17632/86y667257h.2#file5a2847d2-4c9f-41a9-8d7c-cdc74a0195c2}

\bibitem[{Shihavuddin et~al(2013)Shihavuddin, Gracias, Garcia, Gleason, and Gintert}]{shihavuddin2013image}
Shihavuddin A, Gracias N, Garcia R, et~al (2013) Image-based coral reef classification and thematic mapping. Remote Sensing 5(4)

\bibitem[{Shin(2020)}]{shin2020semi}
Shin M (2020) Semi-supervised learning with a teacher-student network for generalized attribute prediction. In: Proceedings of the European Conference on Computer Vision, Springer

\bibitem[{{\v{S}}iaulys et~al(2024){\v{S}}iaulys, Vai{\v{c}}iukynas, Medelyt{\.e}, and Bu{\v{s}}kus}]{vsiaulys2024coverage}
{\v{S}}iaulys A, Vai{\v{c}}iukynas E, Medelyt{\.e} S, et~al (2024) Coverage estimation of benthic habitat features by semantic segmentation of underwater imagery from {South-eastern B}altic reefs using deep learning models. Oceanologia

\bibitem[{Singh et~al(2023)Singh, Rypkema, and Leonard}]{singh2023attention}
Singh K, Rypkema N, Leonard J (2023) Attention-based self-supervised hierarchical semantic segmentation for underwater imagery. In: Proceedings of the OCEANS Conference

\bibitem[{Song et~al(2021)Song, Mehdi, Zhang, Shentu, Wan, Wang, Raza, and Huang}]{song2021development}
Song H, Mehdi SR, Zhang Y, et~al (2021) Development of coral investigation system based on semantic segmentation of single-channel images. Sensors 21

\bibitem[{Song(2023)}]{song2023use}
Song S (2023) The use of color elements in graphic design based on convolutional neural network model. Applied Mathematics and Nonlinear Sciences

\bibitem[{Sorokin(2013)}]{sorokin2013coral}
Sorokin YI (2013) Coral reef ecology, vol 102. Springer Science and Business Media

\bibitem[{Sotoodeh et~al(2019)Sotoodeh, Moosavi, and Boostani}]{sotoodeh2019structural}
Sotoodeh M, Moosavi MR, Boostani R (2019) A structural based feature extraction for detecting the relation of hidden substructures in coral reef images. Multimedia Tools and Applications 78(24)

\bibitem[{Spampinato et~al(2010)Spampinato, Giordano, Di~Salvo, Chen-Burger, Fisher, and Nadarajan}]{spampinato2010automatic}
Spampinato C, Giordano D, Di~Salvo R, et~al (2010) Automatic fish classification for underwater species behavior understanding. In: Proceedings of the ACM International Workshop on Analysis and Retrieval of Tracked Events and Notion in Imagery Streams

\bibitem[{Sun et~al(2018)Sun, Shi, Liu, Dong, Plant, Wang, and Zhou}]{sun2018transferring}
Sun X, Shi J, Liu L, et~al (2018) Transferring deep knowledge for object recognition in low-quality underwater videos. Neurocomputing 275

\bibitem[{S{\"u}nderhauf et~al(2018)S{\"u}nderhauf, Brock, Scheirer, Hadsell, Fox, Leitner, Upcroft, Abbeel, Burgard, Milford et~al}]{sunderhauf2018limits}
S{\"u}nderhauf N, Brock O, Scheirer W, et~al (2018) The limits and potentials of deep learning for robotics. The International Journal of Robotics Research 37(4-5)

\bibitem[{Sward et~al(2019)Sward, Monk, and Barrett}]{sward2019systematic}
Sward D, Monk J, Barrett N (2019) A systematic review of remotely operated vehicle surveys for visually assessing fish assemblages. Frontiers in Marine Science 6

\bibitem[{Szegedy et~al(2016)Szegedy, Vanhoucke, Ioffe, Shlens, and Wojna}]{szegedy2016rethinking}
Szegedy C, Vanhoucke V, Ioffe S, et~al (2016) Rethinking the inception architecture for computer vision. In: Proceedings of the IEEE Conference on Computer Vision and Pattern Recognition

\bibitem[{Szymak et~al(2020)Szymak, Piskur, and Naus}]{szymak2020effectiveness}
Szymak P, Piskur P, Naus K (2020) The effectiveness of using a pretrained deep learning neural networks for object classification in underwater video. Remote Sensing 12(18)

\bibitem[{Tahara et~al(2022)Tahara, Sudo, Yamakita, and Nakaoka}]{tahara2022species}
Tahara S, Sudo K, Yamakita T, et~al (2022) Species level mapping of a seagrass bed using an unmanned aerial vehicle and deep learning technique. PeerJ 10

\bibitem[{Tallam et~al(2023)Tallam, Nguyen, Ventura, Fricker, Calhoun, O’Leary, Fitzgibbons, Robbins, and Walter}]{tallam2023application}
Tallam K, Nguyen N, Ventura J, et~al (2023) Application of deep learning for classification of intertidal eelgrass from drone-acquired imagery. Remote Sensing 15(9)

\bibitem[{Terayama et~al(2022)Terayama, Mizuno, Tabeta, Sakamoto, Sugimoto, Sugimoto, Fukami, Sakagami, and Jimenez}]{terayama2022cost}
Terayama K, Mizuno K, Tabeta S, et~al (2022) Cost-effective seafloor habitat mapping using a portable speedy sea scanner and deep-learning-based segmentation: A sea trial at {Pujada Bay, Philippines}. Methods in Ecology and Evolution 13(2)

\bibitem[{Tian et~al(2020)Tian, Zhang, Shen, Yan, Dong, Yao, Che, Luo, and Han}]{tian2020weakly}
Tian K, Zhang J, Shen H, et~al (2020) Weakly-supervised nucleus segmentation based on point annotations: A coarse-to-fine self-stimulated learning strategy. In: Proceedings of the International Conference on Medical Image Computing and Computer-Assisted Intervention

\bibitem[{Vaswani et~al(2017)Vaswani, Shazeer, Parmar, Uszkoreit, Jones, Gomez, Kaiser, and Polosukhin}]{vaswani2017attention}
Vaswani A, Shazeer N, Parmar N, et~al (2017) Attention is all you need. Advances in Neural Information Processing Systems 30

\bibitem[{Vernaza and Chandraker(2017)}]{vernaza2017learning}
Vernaza P, Chandraker M (2017) Learning random-walk label propagation for weakly-supervised semantic segmentation. In: Proceedings of the IEEE Conference on Computer Vision and Pattern Recognition

\bibitem[{Wang et~al(2020{\natexlab{a}})Wang, Li, Zhou, Meng, Rende, and Rocco}]{wang2020real}
Wang J, Li B, Zhou Y, et~al (2020{\natexlab{a}}) Real-time and embedded compact deep neural networks for seagrass monitoring. In: Proceedings of the IEEE International Conference on Systems, Man, and Cybernetics

\bibitem[{Wang et~al(2020{\natexlab{b}})Wang, Chen, Xie, Azzari, and Lobell}]{wang2020weakly}
Wang S, Chen W, Xie SM, et~al (2020{\natexlab{b}}) Weakly supervised deep learning for segmentation of remote sensing imagery. Remote Sensing 12(2)

\bibitem[{Wang et~al(2020{\natexlab{c}})Wang, Solovyev, Stempkovsky, Telpukhov, and Volkov}]{wang2020method}
Wang W, Solovyev R, Stempkovsky A, et~al (2020{\natexlab{c}}) Method for whale re-identification based on siamese nets and adversarial training. Optical Memory and Neural Networks 29

\bibitem[{Wang et~al(2019)Wang, Ouyang, Li, and Zhang}]{wang2019underwater}
Wang X, Ouyang J, Li D, et~al (2019) Underwater object recognition based on deep encoding-decoding network. Journal of Ocean University of China 18(2)

\bibitem[{Wang et~al(2020{\natexlab{d}})Wang, Zhang, Kan, Shan, and Chen}]{wang2020self}
Wang Y, Zhang J, Kan M, et~al (2020{\natexlab{d}}) Self-supervised equivariant attention mechanism for weakly supervised semantic segmentation. In: Proceedings of the IEEE/CVF Conference on Computer Vision and Pattern Recognition

\bibitem[{Weidmann et~al(2019)Weidmann, J{\"a}ger, Reus, Schultz, Kruschel, Wolff, and Fricke-Neuderth}]{weidmann2019closer}
Weidmann F, J{\"a}ger J, Reus G, et~al (2019) A closer look at seagrass meadows: Semantic segmentation for visual coverage estimation. In: Proceedings of the OCEANS Conference

\bibitem[{Williams et~al(2019)Williams, Couch, Beijbom, Oliver, Vargas-Angel, Schumacher, and Brainard}]{williams2019leveraging}
Williams ID, Couch CS, Beijbom O, et~al (2019) Leveraging automated image analysis tools to transform our capacity to assess status and trends of coral reefs. Frontiers in Marine Science 6

\bibitem[{Wyatt et~al(2022)Wyatt, Radford, Callow, Bennamoun, and Hickey}]{wyatt2022using}
Wyatt M, Radford B, Callow N, et~al (2022) Using ensemble methods to improve the robustness of deep learning for image classification in marine environments. Methods in Ecology and Evolution 13(6)

\bibitem[{Xu et~al(2019)Xu, Bennamoun, An, Sohel, and Boussaid}]{xu2019deep}
Xu L, Bennamoun M, An S, et~al (2019) Deep learning for marine species recognition. In: Handbook of Deep Learning Applications. Springer

\bibitem[{Yang and Gong(2024)}]{yang2024foundation}
Yang X, Gong X (2024) Foundation model assisted weakly supervised semantic segmentation. In: Proceedings of the IEEE/CVF Winter Conference on Applications of Computer Vision

\bibitem[{Yu et~al(2019{\natexlab{a}})Yu, Ma, Farrington, Reed, Ouyang, and Principe}]{yu2019fast}
Yu X, Ma Y, Farrington S, et~al (2019{\natexlab{a}}) Fast segmentation for large and sparsely labeled coral images. In: Proceedings of the International Joint Conference on Neural Networks

\bibitem[{Yu et~al(2019{\natexlab{b}})Yu, Ouyang, Principe, Farrington, Reed, and Li}]{yu2019weakly}
Yu X, Ouyang B, Principe JC, et~al (2019{\natexlab{b}}) Weakly supervised learning of point-level annotation for coral image segmentation. In: Proceedings of the OCEANS Conference

\bibitem[{Yuval et~al(2021)Yuval, Alonso, Eyal, Tchernov, Loya, Murillo, and Treibitz}]{yuval2021repeatable}
Yuval M, Alonso I, Eyal G, et~al (2021) Repeatable semantic reef-mapping through photogrammetry and label-augmentation. Remote Sensing 13(4)

\bibitem[{Zhang et~al(2022)Zhang, Gr{\"u}n, and Li}]{zhang2022deep}
Zhang H, Gr{\"u}n A, Li M (2022) Deep learning for semantic segmentation of coral images in underwater photogrammetry. ISPRS Annals of the Photogrammetry, Remote Sensing and Spatial Information Sciences 2

\bibitem[{Zhang et~al(2024)Zhang, Li, Zhong, and Qin}]{zhang2024cnet}
Zhang H, Li M, Zhong J, et~al (2024) {CNet: A} novel seabed coral reef image segmentation approach based on deep learning. In: Proceedings of the IEEE/CVF Winter Conference on Applications of Computer Vision

\bibitem[{Zhang et~al(2019)Zhang, Zhou, Zhao, Man, Liu, and Yao}]{zhang2019survey}
Zhang M, Zhou Y, Zhao J, et~al (2019) A survey of semi-and weakly supervised semantic segmentation of images. Artificial Intelligence Review

\bibitem[{Zhao and Yin(2020)}]{zhao2020weakly}
Zhao T, Yin Z (2020) Weakly supervised cell segmentation by point annotation. IEEE Transactions on Medical Imaging 40(10)

\bibitem[{Zheng et~al(2023)Zheng, Zhang, Vu, Diao, Tim, and Yeung}]{zheng2023marinegpt}
Zheng Z, Zhang J, Vu TA, et~al (2023) {MarineGPT: U}nlocking secrets of ocean to the public. arXiv preprint arXiv:231013596

\bibitem[{Zheng et~al(2024)Zheng, Liang, Hua, Wong, Ang, Chui, and Yeung}]{zheng2024coralscop}
Zheng Z, Liang H, Hua BS, et~al (2024) Coralscop: Segment any coral image on this planet. In: Proceedings of the IEEE/CVF Conference on Computer Vision and Pattern Recognition

\bibitem[{Zhong et~al(2023)Zhong, Li, Zhang, and Qin}]{zhong2023combining}
Zhong J, Li M, Zhang H, et~al (2023) Combining photogrammetric computer vision and semantic segmentation for fine-grained understanding of coral reef growth under climate change. In: Proceedings of the IEEE/CVF Winter Conference on Applications of Computer Vision

\bibitem[{Zhou et~al(2021)Zhou, Wei, Wang, Shen, Xie, Yuille, and Kong}]{zhou2021ibot}
Zhou J, Wei C, Wang H, et~al (2021) i{BOT}: Image bert pre-training with online tokenizer. arXiv preprint arXiv:211107832

\bibitem[{Zhu et~al(2023)Zhu, Xiong, and Lu}]{zhu2023survey}
Zhu K, Xiong NN, Lu M (2023) A survey of weakly-supervised semantic segmentation. In: Proceedings of the IEEE International Conference on Big Data Security on Cloud, IEEE International Conference on High Performance and Smart Computing, and IEEE International Conference on Intelligent Data and Security

\bibitem[{Ziqiang et~al(2023)Ziqiang, Yaofeng, Haixin, Zhibin, and Yeung}]{ziqiang2023coralvos}
Ziqiang Z, Yaofeng X, Haixin L, et~al (2023) {CoralVOS: D}ataset and benchmark for coral video segmentation. arXiv preprint arXiv:231001946

\end{thebibliography}

\end{document}